\documentclass[11pt]{article} %

\usepackage[preprint]{tmlr}

\usepackage{mathpazo}

\usepackage{microtype}

\usepackage{amsmath,amsfonts,bm}

\def\eqref#1{equation~\ref{#1}}

\def\1{\bm{1}}

\def\mA{{\bm{A}}}
\def\mB{{\bm{B}}}

\def\mM{{\bm{M}}}

\def\mP{{\bm{P}}}
\def\mQ{{\bm{Q}}}
\def\mR{{\bm{R}}}

\def\mX{{\bm{X}}}
\def\mY{{\bm{Y}}}

\DeclareMathAlphabet{\mathsfit}{\encodingdefault}{\sfdefault}{m}{sl}
\SetMathAlphabet{\mathsfit}{bold}{\encodingdefault}{\sfdefault}{bx}{n}

\def\tA{{\tens{A}}}

\newcommand{\E}{\mathbb{E}}

\DeclareMathOperator*{\argmax}{arg\,max}
\DeclareMathOperator*{\argmin}{arg\,min}

\usepackage{amsmath}
\usepackage{amsfonts} 
\usepackage{mathtools}
\usepackage{amsthm} %
\usepackage{amssymb}
\usepackage{mathabx}

\usepackage{graphicx}
\usepackage{subcaption}

\usepackage{graphicx}
\usepackage[font=small]{caption}
\usepackage[font=small]{subcaption}
\usepackage{amsthm} %
\usepackage[e]{esvect}
\usepackage{enumerate}
\usepackage{enumitem}
\usepackage{booktabs}

\usepackage[colorlinks=true, linkcolor=magenta, citecolor=magenta, urlcolor=magenta]{hyperref}
\usepackage{url}

\usepackage{xcolor}
\usepackage{mdframed}

\definecolor{mygray}{gray}{0.92}
\definecolor{mycolor2}{RGB}{255, 250, 176}

\newmdenv[backgroundcolor=mygray, linewidth=0pt]{sectionpurpose}

\usepackage{xspace}
\newcommand*{\eg}{e.g.\@\xspace}
\newcommand*{\ie}{i.e.\@\xspace}

\newcommand*{\versus}{vs.\@\xspace}

\newtheorem{theorem}{Theorem}[section]       %
\newtheorem{proposition}[theorem]{Proposition}
\newtheorem{definition}[theorem]{Definition}

\newtheorem{lemma}[theorem]{Lemma}
\newtheorem{remark}[theorem]{Remark}
\newtheorem{defremark}[theorem]{Remark/Def.}
\newtheorem{assumption}[theorem]{Assumption}
\newtheorem{example}[theorem]{Example}

\newtheorem{desideratum}[theorem]{Desideratum}
\newtheorem{definitionprop}[theorem]{Def./Observation}

\usepackage{xspace}

\newcommand{\RomanNumeralCaps}[1]
    {\MakeUppercase{\romannumeral #1}}

\renewcommand{\mX}{\mathcal{X}}
\renewcommand{\mY}{\mathcal{Y}}
\renewcommand{\mB}{\mathcal{B}}
\renewcommand{\mM}{\mathcal{M}}

\renewcommand{\mQ}{\mathcal{Q}}
\renewcommand{\mR}{\mathbb{R}}
\newcommand{\Pup}{\overline{P}}
\newcommand{\Plow}{\underline{P}}
\newcommand{\Rup}{\overline{R}}
\newcommand{\Rlow}{\underline{R}}

\newcommand{\gbr}{\operatorname{GBR}}

\newcommand{\linfty}{L^\infty}

\newcommand{\cobar}{\overline{\operatorname{co}}}

\newcommand{\iid}{\textit{i.i.d.}\@\xspace}
\newcommand{\etc}{etc.\@\xspace}
\renewcommand{\mP}{\mathcal{P}}
\renewcommand{\mQ}{\mathcal{Q}}

\makeatletter
\newcommand{\shorteq}{%
  \settowidth{\@tempdima}{-}%
  \resizebox{\@tempdima}{\height}{=}%
}
\makeatother

\newcommand{\setofips}{\mathcal{IP}}

\newcommand{\aopt}{a_{\mQ}^*}

\newcommand{\aoptx}{a_{\mQ(x)}^*}
\newcommand{\aoptXom}{a_{\mQ(X(\omega))}^*}

\newcommand{\picka}{\leftarrow}

\newcommand{\Lfuncs}{\mathcal{L}}

\newcommand{\samehull}{\simeq}
\newcommand{\notsamehull}{\not\simeq}

\newcommand{\mPx}{\mP_{|x}}

\newcommand{\actionpartition}{\mathcal{B}_{\mQ;\mathcal{A}}}
\newcommand{\actionpartelement}{B_{\mQ;a}}

\newcommand{\Sl}{\tilde{S}_\ell}

\newcommand{\baseip}{\mathcal{P}}

\newcommand{\ipscore}{\operatorname{IP-Score}^*}

\newcommand{\ta}{\tilde{a}}
\renewcommand{\tA}{\tilde{A}}

\renewcommand{\mA}{\mathcal{A}}

\newcommand{\ellwc}{\ell_{W;c}}

\newcommand{\ellB}{(\ell,\mathcal{B})}

\usepackage{stmaryrd}

\usepackage{tcolorbox}

\newtcolorbox{assumptionbox}{
    colback=white, %
    colframe=black, %
    sharp corners, %
    top=8pt, %
    bottom=8pt, %
    left=12pt, %
    right=12pt, %
}

\title{Scoring Rules and Calibration for Imprecise Probabilities}

\author{\name Christian Fröhlich \email christian.froehlich@uni-tuebingen.de \\
      \name Robert C. Williamson \email bob.williamson@uni-tuebingen.de \\
      \addr Department of Computer Science\\
      University of Tübingen\\ 
      and Tübingen AI Center
      }

\def\month{MM}  %
\def\year{YYYY} %
\def\openreview{\url{https://openreview.net/forum?id=XXXX}} %

\begin{document}

\maketitle

\begin{abstract}
What does it mean to say that, for example, the probability for rain tomorrow is between $20\%$ and $30\%$? The theory for the evaluation of precise probabilistic forecasts is well-developed and is grounded in the key concepts of \textit{proper scoring rules} and \textit{calibration}. For the case of imprecise probabilistic forecasts (sets of probabilities), such theory is still lacking. In this work, we therefore generalize proper scoring rules and calibration to the imprecise case. We develop these concepts as relative to data models and decision problems. 
As a consequence, the imprecision is embedded in a clear context.
We establish a close link to the paradigm of \textit{(group) distributional robustness} and in doing so provide new insights for it.
We argue that proper scoring rules and calibration serve two distinct goals, which are aligned in the precise case, but intriguingly are not necessarily aligned in the imprecise case.  
The concept of decision-theoretic entropy plays a key role for both goals. 
Finally, we demonstrate the theoretical insights in machine learning practice, in particular we illustrate subtle pitfalls relating to the choice of loss function in distributional robustness. 
\end{abstract}

\section{Introduction}
The field of machine learning is inextricably intertwined with probability theory. In particular, a range of machine learning methods output probabilistic predictions (or \textit{forecasts}). In the case of classification, given an input datum $X=x$, such a predictor $f(x)$ returns a probabilistic forecast for the associated label $Y$.
While it is often assumed that probability theory is \textit{the} comprehensive framework for mathematizing uncertainty, various challenges have been raised that focus on its shortcomings, both from subjectivist  \citep{levi1980enterprise,walley1991statistical,gilboa2009always,joyce2010defense} as well as generalized frequentist perspectives \citep{fine1988lower,frohlich2022risk,frohlich2024strictly} (see Section~\ref{sec:hierarchyofimprecision}).
In one cluster of responses, scholars have argued that relevant shortcomings may be addressed by moving from precise probabilities to \textit{imprecise} probabilities (IP), which are essentially equivalent to sets of probabilities. In the case of a binary event (``will it rain tomorrow?''), such an imprecise probability is simply an interval $[p,q] \subseteq [0,1]$.

Following this line, 
our goal is to construct machine learning methods which output \textit{imprecise} forecasts. However, this necessitates the development of a principled theory for the \textit{evaluation} of such forecasts which is non-existent so far: in light of training data, how ``good'' was a given IP forecast? Access to such an evaluation procedure would also provide guidance on how to obtain such forecasts,
 which allows the training of machine learning models.
In similar spirit, in recent years there have been attempts to develop a theory for the evaluation of imprecise \textit{beliefs} from a subjectivist perspective \citep{seidenfeld2012forecasting,konek2019epistemic,konek2019ip,konek2023evaluating}.\footnote{While \citet{seidenfeld2012forecasting} also use the term ``forecast'', their approach is subjectivist and thus aligned with \citet{konek2019epistemic}, who uses the term ``credences''. 
To us, the term ``forecast'' carries a (generalized) frequentist flavor and implies evaluation against a data sequence or a data model.} 
To the best of our knowledge, 
there is no proposal which claims to offer a satisfying positive development of a theory for the evaluation of imprecise forecasts, and indeed some authors provide impossibility theorems, which at first sight may seem to threaten the whole endeavour. We are not discouraged by the negative results (see the discussion in Section~\ref{sec:propriety}) and offer the following diagnosis for what has been missing in previous approaches so far: \textit{data models} and \textit{decision problems}. Our philosophy differs from the subjectivist paradigm in that we emphasize the primacy of data and data model in forecasting;  
we do not think that imprecise forecasts can be evaluated ``in the void'', but only relative to data (model) and decision problem. Consequently, attention centers on imprecision \textit{in the data model}, and indeed our findings support the position that this kind of imprecision is to some extent even more fundamental than imprecision \textit{in forecasts}.

We summarize our paper and its contributions as follows: we develop a viable account of proper IP scoring rules, which are relativized to data models and decision problems. We characterize the optimal IP forecast \textit{for a fixed loss function} based on the maximum entropy principle by \citet{grunwald2004game}. We demonstrate that imprecision in the forecasts might be required if the loss function is not known a priori. We shed light on the tight connection between our proposal of imprecise scoring rules and the paradigm of  \textit{(group) distributional robustness} \citep{hu2018does,sagawa2019distributionally}, where a \textit{coherent risk measure} of the loss is optimized.
Further, we develop an adequate concept of \textit{calibration} for imprecise probabilities. As for scoring rules, the concept of a generalized decision-theoretic \textit{entropy} plays a key role here. 
We discuss how proper scoring rules and calibration serve two distinct goals of forecasting quality, and show that they do not necessarily align anymore in the imprecise case.
In particular, the optimal forecast (in the sense of score) need not be calibrated. 
Finally,  in Section~\ref{sec:experiments}, we demonstrate the benefits of imprecise forecasts in machine learning practice.
In addition, we believe that even readers who are only interested in precise probability may profit from reading this paper, as the development we offer here provides new insight into the conceptual interplay between scoring rules and calibration in general.

To set the stage, 
by an \textit{imprecise probability} \citep{walley1991statistical,augustin2014introduction}, we here mean any non-empty set of probability measures $\baseip$. Associated to this set is a generalized non-linear expectation functional 
\[
\Rup_{\baseip}(X) \coloneqq \sup_{P \in \baseip} \E_P[X].
\] This functional is known as \textit{coherent upper prevision} \citep{walley1991statistical} in the IP community and equivalently as \textit{coherent risk measure} \citep{artzner1999coherent} in finance and insurance.\footnote{We refer the reader to \citep{frohlich2022risk} for an explanation of the subtle conceptual relations.} In this paper, we choose the perhaps more neutral term \textit{upper expectation}. Similarly, the \textit{upper probability} of an event $A$ is $\Pup_{\baseip}(A) \coloneqq \sup_{P \in \baseip} P(A)$. 
The conjugate lower quantities are $\Rlow(X) \coloneqq -\Rup(-X)$ and $\Plow(A)\coloneqq 1-\Pup(A^C)$. 
Hence for an event we obtain an interval $[\Plow(A),\Pup(A)] = [\inf_{P \in \baseip} P(A),\sup_{P \in \baseip} P(A)]$, where the width is the degree of imprecision, and similarly for $\Rlow$ and $\Rup$.

\subsection{Data Models and Imprecision}
\label{sec:hierarchyofimprecision}
In contrast to previous literature with the goal of developing scoring rules for imprecise forecasts, we put \textit{data models} \citep{frohlich2024data} at the center of the stage. Intuitively, we understand by data model a mathematical formalization and idealization of a data-generating process, therefore with an inevitable aspect of subjectivity as in any modelling process. As an example, consider the familiar \iid model, which, at the level of abstraction where we use the term, is one data model. Our logic is that we should first seek to understand what possible semantics imprecision can have with respect to data models. This enables us to subsequently embed imprecise forecasts in a meaningful context of data model and decision problem.
The situation is however complicated, since imprecision has various different semantics in the literature, which are moreover not conceptually clear-cut.
We therefore briefly survey these semantics now (non-comprehensively), noting that there is a rough hierarchical progression among them.

First, in the setting of objective \textit{risk}, the assumption is that the data model is parameterized by a single, precise probability, and that it is known to the decision maker. For example, consider flipping a fair coin. This process is well-described by an \iid model with a probability $p=0.5$ for the event ``heads''. In this setting there is still uncertainty about the outcome, often called \textit{aleatoric uncertainty}.
Following expected utility (EU) theory \citep{von1947theory}, a decision maker would use the expectation to assess the risk of a gamble (a random variable). However, a plenitude of criticism has been raised against EU theory \citep{allais1953comportement,buchak2013risk}. Importantly, it is unclear whether EU theory can adequately represent the phenomenon of risk aversion \citep{buchak2013risk}. As a response, \textit{rank-dependent} EU \citep{wang1995insurance,quiggin2012generalized} has been developed, closely related to particular upper expectations known as \textit{spectral risk measures} \citep{acerbi2002spectral,wang2000}.
In this way, imprecise probabilities can serve to express risk aversion. Note that here imprecision manifests not in the (precise) data model but in the non-linear \textit{aggregation} of consequences.

Risk is often contrasted with \textit{ambiguity} \citep{ellsberg1961risk}.\footnote{We remark that this distinction is not without controversy. Some authors define risk as the situation where probabilities are either known \textit{or} they can be reliably estimated from experience (\eg \citep{levy2010neural}). Risk in the context of insurance (\eg for workplace accidents) would fit this definition, for example. Other times, the term risk is used in a strong objective sense, where the probabilities are exactly known, as in the lotteries of \citeauthor{von1947theory}'s \citeyearpar{von1947theory} expected utility theory. To us, the line appears blurry.} Assume now that the hypothesized ``true'' data model is parameterized by a single, precise probability, but it is unknown to the decision maker. For instance, consider flipping a coin with unknown bias. Here, the model of an underlying precise probability still appears adequate. The problem now has a component of \textit{epistemic uncertainty}, although aleatoric uncertainty is additionally still present.\footnote{See \citep{hullermeier2021aleatoric} for the distinction of aleatoric versus epistemic uncertainty.} At this level, imprecise probabilities, appearing also under the banner of \textit{distributional robustness} \citep{kuhn2019wasserstein,rahimian2019distributionally} in machine learning, can be employed to capture the attitude of \textit{ambiguity aversion} towards epistemic uncertainty (see also \citep{fox2021ambiguity}). 
This setting describes a typical machine learning problem, where we only have access to the empirical train distribution (the training data), but not to the hypothesized ``true'' distribution. The goal of distributionally robust optimization (DRO) in this context is then to robustify against this lack of knowledge \citep{kuhn2019wasserstein}.

Until now, the data model was parameterized by a precise probability, and imprecision expressed an attitude towards either risk (related to aleatoric uncertainty) or ambiguity (related to epistemic uncertainty, plus aleatoric uncertainty). As another setting, intensifying further, we introduce the data model which will be the focus of this paper, where the data comes from a set of probabilities. Let $\Omega$ with $|\Omega|<\infty$ be a possibility set, which captures all relevant information. Intuitively, $\omega \in \Omega$ is a state of the world. For example, $\Omega = \mathcal{X} \times \mathcal{Y}$. Then $\omega^\infty = (\omega_1, \omega_2, ..)$ is a data sequence.
The \textit{non-stationary, locally precise (NSLP) data model} \citep{frohlich2024data} is defined using independent draws from a sequence of probabilities.
Fix a sequence $P^\infty=(P_1,P_2,..)$ of probability measures on $\Omega$, and demand that
\[
\lambda\left\{ \omega^\infty = (\omega_1, \omega_2, ..)\right\} = \prod_{i=1}^\infty P_i(\{\omega_i\}),
\]
where $\lambda$ is the Lebesgue measure.\footnote{We implicitly work on the measure space $([0,1),\mathcal{B}([0,1)),\lambda)$, but suppress this in the notation. See \citep{frohlich2024data} for the rigorous formulation involving maps of the signature $W_i : [0,1] \to \Omega$.}
This model generalizes the \iid model to a set of probabilities, an imprecise probability $\baseip \coloneqq \{P_i : i \in \mathbb{N}\}$, and models the data as independent with respect to a sequence of probabilities from this set. Intuitively, consider a sequence of coin flips with varying biases. This model, in our view, can be seen as a rather mild departure from an \iid model, as it keeps independence and the drawing of each datum is still precise probabilistic. Note, however, that under this model, imprecision is inherent to the data model, and not reflective of our epistemic uncertainty (the predictor only has access to information in $\omega$).\footnote{It should be highlighted that we think of the epistemic \versus aleatoric distinction as essentially contextual: adding additional information to $\omega$, for instance by collecting a new feature, can change the picture.} How to adequately model epistemic uncertainty in this picture in addition is an open problem to the best of our knowledge, which we put aside here. This model appears (implicitly) in many applications in machine learning, for example in the context of fairness \citep{buolamwini2018gender}, federated learning \citep{mcmahan2017communication} and more. The paradigm of \textit{group distributional robustness}\footnote{The distinction between \textit{distributional robustness} (\eg \citep{duchi2018learning}) and \textit{group distributional robustness} (\eg \citep{sagawa2019distributionally}) is, in our eyes, to some extent blurry, but can be related to how the set of probabilities is constructed. Distributional robustness would typically employ \textit{law-invariant coherent risk measures} \citep{follmer2016stochastic}, for example based on an $f$-divergence ball around the empirical distribution, whereas in group distributional robustness, the set of probabilities corresponds directly to a set of identified subpopulations in the data. Due to envelope representations \citep{pflug2007modeling}, both approaches can be interpreted in terms of subpopulations, however.} \citep{sagawa2019distributionally} fits well with this data model, where the assumption is that each group $S=s$ corresponds to a distinct probability $P_{S=s}$. 
Group DRO is relevant in practice, for example, in a fairness context when the ``sensitive attribute''\footnote{In fair machine learning, a sensitive attribute is a sensitive piece of information about a person, such as race or gender. Such an attribute  canonically induces a set of groups.} is only available at training time, but we aim to deploy a single model at test time, with the aim of robust performance with respect to the sensitive attribute; or, when we know of a spurious correlation in the training data, which can be used to define a set of groups \citep{sagawa2019distributionally}.
Throughout the paper we are guided by this data model for inspiration, as the imprecision here is fundamental to the data model, implying that the imprecision is not a matter of our lack of knowing the model. The mathematical development here is not tied to this data model though, but interpretations may change.

Finally, we mention in passing that there exists a whole zoo of stationary data models which are parameterized by an imprecise probability \citep{fine1988lower,frohlich2024data}, but we believe the clearest in terms of applications is the NSLP model presented above. For brevity, we refer to a data model parameterized by some imprecise probability $\baseip$ as a ``$\baseip$ data model''.

\section{IP Scoring Rules}
\label{sec:ipscoring}
In this section we present an account of proper scoring rules for imprecise probabilities.
Our primary goal is to enable decision making for an agent who is associated with a loss function. This means that our forecaster (the machine learning model) should recommend ``favorable'' actions to the agent, where a ``favorable'' action is one that leads to low loss values. That is, for us the evaluation of a forecast is inextricably intertwined with action recommendation, which shapes our perspective throughout the paper. The aggregation of these losses, which determines the overall performance, is grounded in the data model. 
We foreshadow that in Section~\ref{sec:ipcalibration} we will introduce another, orthogonal goal for the decision maker: intuitively, forecasts should also be trustworthy in the sense that their associated uncertainty should be ``appropriate'' (calibrated).

Our setup is as follows. Assume a possibility set $\Omega = \mathcal{X} \times \mathcal{Y}$, with features $X: \Omega \to \mathcal{X} \subset \mR$\footnote{We could also have $\mR^d$-valued features, since $\mR$ and $\mR^d$ have the same cardinality.} and labels $Y: \Omega \to \mathcal{Y} \subset \mR$. To reduce the burden of technicalities, we assume $2 \leq |\Omega|=k<\infty$ throughout the paper. The set of bounded functions is $\linfty \coloneqq \{Z: \Omega \to \mR\}$ in this case, and we call such functions $Z \in \linfty$ \textit{gambles}.
We assume that an agent needs to make a decision in the form of choosing an action $a^* \in \mathcal{A}$, where the action space $\mathcal{A}$ is assumed finite for simplicity and $|\mathcal{A}|\geq 2$. The consequence is then evaluated by a loss function $\ell: \mathcal{A} \times \Omega \to \mR$, where $\ell_a = \ell(a,\cdot) \in \linfty \; \forall a \in \mathcal{A}$. We denote the set of loss functions by $\mathcal{L}$. The quantity $\ell(a,\omega)$ represents the loss (disutility) that the agent incurs when following action $a \in \mA$ and the state of the world $\omega \in \Omega$ actually materializes. Note that we impose no further constraints on $\ell$. Throughout the paper, we often need to access the elementary event directly, and thus write expressions of the form $\E_P[\ell(a,\omega)]$, where $P$ is a probability measure on $\Omega$, as shorthand for $\E_P[\omega \mapsto \ell(a,\omega)]$.\footnote{In other words, in expressions such as $\E_P[\ell(a,\omega)]$ we implicitly treat $\omega$ as the identity random variable.}

\begin{example}\normalfont
    As a standard example, consider predicting whether it will rain ($Y=1$) or not ($Y=0$) on a given day, which is represented by some  features $X$ (temperature \etc). The action space $\mathcal{A}$ could consist of the actions ``bringing an umbrella'' ($a=1$) and ``not bringing an umbrella'' ($a=0$). A choice of a loss function $\ell$ is then equivalent to a real-valued $2 \times 2$ matrix, assessing the possible consequences. Negative values would correspond to utility (gain), whereas positive values correspond to disutility (loss). We are then interested in evaluating an imprecise forecaster in light of an actual outcome sequence, in this case a binary sequence indicating ``rain'' or ``no rain''. We extend this example in Appendix~\ref{app:exampleofmaxentfailure}.
\end{example}

The main goal is to recommend a supposedly ``optimal'' action to the agent in terms of the consequences. There are two main open questions:
\begin{enumerate}[nolistsep,start=1,label=\textbf{Q\arabic*.}, ref=Q\arabic*]
    \item \label{question:actionchoice} Given an imprecise forecast, how does the agent choose an action?
    \item \label{question:aggregation} How do we aggregate losses, that is, how do we summarize the performance in the long run over many $\omega$s?
\end{enumerate}
To answer~\ref{question:actionchoice}, we first define the notion of an imprecise forecast.
Since $|\Omega|<\infty$, a probability measure $P$ on $\Omega$ can be identified with a point in the simplex $\Delta^k \coloneqq \{(p_1,p_2,..,p_k) : \sum_{j=1}^k p_j = 1\}$.\footnote{On $\Delta^k$ we can assume the Euclidean topology without loss of generality \citep{frohlich2024data}.} We write $\setofips$ for the set of nonempty subsets of $\Delta^k$, the imprecise probabilities. A precise probability $P \in \Delta^k$ can be seen as an imprecise probability $\{P\} \in \setofips$, a viewpoint which we adopt implicitly throughout the paper.
An imprecise forecast is then a function $\mathcal{Q} : \mathcal{X} \to \setofips$, which takes as input the features $X=x$ and outputs a set of probabilities $\mathcal{Q}(x) \coloneqq \mQ(X(\omega)) \in \setofips$ on $\Omega$. Clearly, a reasonable forecast $\mathcal{Q}(x)$ would assign zero mass to those $\omega$ where $X(\omega) \neq x$, at least if the forecaster considers the $X=x$ information as perfectly trustworthy.
However, we do not impose this restriction in this paper for additional flexibility, although this introduces some new translation effort. 
We slightly abuse notation and also write $\mathcal{Q} \in \setofips$ for a constant forecast (be it precise or actually imprecise).

Having access to a forecast $\mQ(x)$, the agent needs to choose an action. There are multiple, non-equivalent decision-making principles for imprecise probabilities \citep{troffaes2007decision}.  %
In this paper we consider a \textit{worst-case} (pessimistic) attitude for multiple reasons, primarily because it can interplay nicely with aggregation in the data model. This principle is also known as ($\Gamma$-)MinMax \citep{gilboa1989maxmin}.\footnote{Or ``$\Gamma$-MaxMin'' if one works with a utility-based orientation, equivalent up to a sign flip.}
\begin{assumption}
\label{minmaxassumption1}
    The agent chooses the action by the \textit{MinMax} principle:
    \begin{equation}
    \label{eq:minmaxchoice}
        \aoptx \picka \argmin_{a \in \mathcal{A}} \sup_{Q \in \mQ(x)} \E_Q[\ell(a,\omega)],
    \end{equation}
    where $\mathcal{Q}(x) \in \setofips$ is the imprecise forecast. 
    The ``$\picka$'' operator picks a single element from the potentially set-valued $\argmin$ in a consistent way, see Appendix~\ref{app:tiebreaking}.
\end{assumption}
For brevity, we usually write $\mQ(X)$ or $\mQ(\omega)$ instead of $\mQ(X(\omega))$, but sometimes choose the latter for emphasis.
After having committed to an action $\aoptXom$, the agent incurs the loss $\ell(\aoptXom,\omega)$. This key quantity of interest defines our IP scoring rule.
\begin{definition}
\label{def:ipscoringrule}
    For a loss function $\ell \in \Lfuncs$, we define the IP scoring rule as:
    \begin{equation}
        \Sl(\omega) \coloneqq S_\ell\left(\mQ(X(\omega)),\omega\right) \coloneqq \ell(\aoptXom,\omega).
    \end{equation}
\end{definition}
Note that the definition of this function might implicitly depend on how tie-breaking in the $\argmin$ (\eqref{eq:minmaxchoice}) is resolved.
The IP scoring rules quantifies the adverse consequences resulting from following the forecast $\mQ$ when $\omega$ materializes. This way of obtaining a scoring rule essentially generalizes the concept of a \textit{tailored scoring rule} \citep{dawid1999coherent,johnstone2011tailored} by incorporating a MinMax assumption. Here, \textit{tailored} means that the scoring rule $S_\ell$ is constructed in such a way so as to enable decision making for an agent with underlying loss function $\ell$.

There is still one crucial ingredient missing here, which is the data model. Introducing a data model then allows answering~\ref{question:aggregation}, that is, how to aggregate the scores over repeated outcomes $\omega_1,\omega_2,..$ and so on. For the experiments in this paper, we will focus on the NSLP data model presented in Section~\ref{sec:hierarchyofimprecision}, which is parameterized by a set of probabilities $\baseip$ on $\Omega$; the theoretical developments however can also be instantiated with other models, as long as they are parameterized by an imprecise probability. For such decision problems, we propose a MinMax assumption, which will establish the link to DRO (Section~\ref{sec:linktodro}):
\begin{enumerate}[nolistsep,start=1,label=\textbf{D\arabic*.}, ref=D\arabic*]
    \item \label{item:D2} MinMax-Assumption 2: we aim for worst-case (``robust'') performance over the underlying set of probabilities.
\end{enumerate}
Combining this with our IP scoring rule, we express \ref{item:D2} as a MinMax objective:%
\begin{assumption}
\label{minmaxassumtion2}
    When the data model is parameterized by $\baseip \in \setofips$, the overall problem (minimizing IP-Score) from the perspective of the forecaster is:
    \begin{equation}
        \label{eq:overallobjective}
        \ipscore(\ell,\baseip) \coloneqq \min_{\mQ : \mathcal{X} \rightarrow \setofips} \sup_{P \in \baseip} \E_{P}[S_\ell(\mQ(X(\omega)),\omega)],
    \end{equation}
    where the minimization is over all imprecise forecasts.\footnote{The minimum is indeed attained since with a finite action space, $S_\ell(\mQ(X(\omega)),\omega)$ can only take on finitely many distinct values.} %
\end{assumption}
Using the upper expectation, $\ipscore(\ell,\baseip) = \min_{\mQ : \mathcal{X} \rightarrow \setofips} \Rup_{\mP}(S_\ell\left(\mQ(X(\omega)),\omega\right))$. 
We refer to the inner quantity $\Rup_{\mP}(S_\ell\left(\mQ(X(\omega)),\omega\right))$ as the IP score of $\mQ$ under a $\baseip$ data model.
Even though in practice, we of course work with constrained hypothesis classes, we consider the unconstrained objective in theory.
Observe that the MinMax assumptions~\ref{minmaxassumption1} and~\ref{minmaxassumtion2} appear similar, but are conceptually distinct. Whereas~\ref{minmaxassumption1} refers to how the agent subjectively picks an action, \ref{minmaxassumtion2} is about aggregation of losses in the data model. However, it is the shared MinMax form that enables their constructive interplay.
What we believe is unique about the MinMax criterion, in contrast to other criteria \citep{troffaes2007decision}, is that it appears sensible both from an agent's as well as a data model's perspective.

\begin{defremark}\normalfont
    We note that if $\cobar(\baseip)=\cobar(\baseip')$, then $\Rup_{\baseip}=\Rup_{\baseip'}$, where $\cobar(\cdot)$ denotes the closed convex hull, that is, upper expectations are unique only up to closed convex hull \citep[Section 3.6]{walley1991statistical}; see also \citep[Proposition 2.2.1]{hiriart2004fundamentals}. If $\cobar(\baseip)=\cobar(\baseip')$, we write $\baseip \samehull \baseip'$, and otherwise $\baseip \notsamehull \baseip'$. 
\end{defremark}

\subsection{Weak and Strong Propriety}
\label{sec:propriety}
A first question to ask, and also a sanity check, is whether our IP scoring rules satisfy a notion of \textit{propriety}. From a data model perspective, we suggest, a scoring rule may be called (strictly) proper if the optimal forecast coincides with the parameterization of the data model. This is the key for estimation procedures to work. In the precise case of an \iid model parameterized by $P$, this means that $P$ is an optimal constant forecast under a proper scoring rule.

To state the definition of propriety, we need to ignore the features for a moment and focus on constant forecasts. In the precise case, a scoring rule $s : \Delta^k \times \Omega \to \mR$ is called weakly proper if $\mathbb{E}_P[s(P,\omega)] \leq \mathbb{E}_P[s(Q,\omega)] \quad \forall P, Q \in \Delta^k$.
That is, no other forecast $Q$ can do better than the underlying $P$. This can straightforwardly be generalized to our IP scoring rules.

\begin{proposition}[Weak propriety] 
\label{prop:weakpropriety}
Let $\ell \in \Lfuncs$ and $\baseip,\mQ \in \setofips$. Then it holds that
    \begin{equation}
        \label{eq:weakpropriety}
        \Rup_{\baseip}(S_\ell(\baseip,\omega)) \leq \Rup_{\baseip}(S_\ell(\mQ,\omega)).
    \end{equation}
\end{proposition}
The proof is in Appendix~\ref{app:prop:weakpropriety}. 
Note that the existence of weakly proper IP scoring rules is not in contradiction to previous impossibility theorems in the literature \citep{seidenfeld2012forecasting,mayo2016scoring,schoenfield2017accuracy}, which preclude \textit{strictly} proper IP scoring rules. Strict propriety would mean that equality in \eqref{eq:weakpropriety} only holds if $\baseip=\mQ$. Clearly, if $\mP \samehull \mQ$, then $S_\ell(\baseip,\omega) = S_\ell(\mQ,\omega)$, meaning the set of extreme points is an equivalent forecast since $\operatorname{ext}(\baseip) \samehull \baseip$. This could be seen as a failure of strict propriety, but we only really differentiate between sets of probabilities up to closed convex hull. Another question is whether this can happen if $\baseip \notsamehull \mQ$. Thus a further question is whether strict propriety can fail with an actually different forecast. 
First, we note that there are $\ell \in \Lfuncs$ which lead to complete failure of strict propriety (consider for example the constant zero loss). Thus the question of interest is whether there exist any $\ell$ such that the induced IP scoring rule is strictly proper (when accounting for the ``$\samehull$'' equivalence).
Indeed our IP scoring rules have the following defect when looking at precise forecasts for comparison.

\begin{proposition}[Partial failure of strict propriety]
\label{prop:failureofstrict}
 The following statements hold:
\begin{enumerate}[label=(\theproposition.\arabic*)]
    \item \label{failurestrict2} $\forall \ell \in \Lfuncs : \exists \baseip \in \setofips : \exists P \in \cobar(\baseip)$ so that $\Rup_{\baseip}(S_\ell(\baseip,\omega)) = \Rup_{\baseip}(S_\ell(P,\omega))$.
    \item \label{failurestrict3} Let $|\mathcal{A}|\geq 3$. Then $\exists \ell : \exists \baseip$ so that $\forall P \in \Delta^k : \Rup_{\baseip}(S_\ell(\baseip,\omega)) < \Rup_{\baseip}(S_\ell(P,\omega))$.
\end{enumerate}
\end{proposition}
The proof is in Appendix~\ref{app:prop:failureofstrict}.
In summary, if we we mean by strict propriety that the inequality in Definition~\ref{eq:weakpropriety}, for a fixed $\ell \in \Lfuncs$, holds \textit{strictly} for all pairs of $\baseip,\mQ$, even if restricting us to $\baseip \notsamehull \mQ$, then our IP scoring rules are not strictly proper. 
Statement~\ref{failurestrict2} asserts that for any $\ell$, strict propriety must fail in combination with some $\baseip$ due to the existence of an equivalent precise forecast $P \in \cobar(\baseip)$; although statement~\ref{failurestrict3} assures us that for some $\ell$ this need not happen \textit{for all} $\baseip$, meaning there exist pairs $(\ell,\baseip)$ so that no precise forecast can be optimal. The take-away is that the situation is intricate: whether we can reduce the problem of finding an optimal constant IP forecast to finding an constant optimal precise forecast may depend on both $\ell$ and $\baseip$. This will generalize to the non-constant case, see Section~\ref{sec:reductiontoprecise}.

Moreover, observe that in statement~\ref{failurestrict2} the precise forecast $P \in \cobar(\baseip)$ which is equivalent to $\baseip$ may depend on $\ell$.
In practice, the loss function relevant at test time may not always be known at training time, where we typically optimize a surrogate loss instead of the loss we actually care about. Recently, the paradigm of \textit{omniprediction} emerged in the machine learning community \citep{gopalan2021omnipredictors}, where the goal is to perform well with respect to many loss functions simultaneously. Related to this framework, we make the following observation about our IP scoring rules: for any other candidate forecast $\mQ \in \setofips$, there exists a loss function $\ell$ so that the IP score of the forecast $\mP$ under a $\baseip$ data model will be lower than that of  $\mQ$.

\begin{proposition}[``Strong'' propriety]
\label{prop:strongpropriety}
    Let $\baseip,\mQ \in \setofips$ be any constant forecast so that $\mQ \notsamehull \baseip$. Then there must exist some loss function $\ell \in \Lfuncs$ such that
    \[
    \Rup_{\baseip}(S_\ell(\baseip,\omega)) < \Rup_{\baseip}(S_\ell(\mQ,\omega)).
    \]
\end{proposition}
The proof is in Appendix~\ref{app:prop:strongpropriety}. Note the order of the quantifiers, constrasting with that of Proposition~\ref{prop:failureofstrict}. In words, if our goal is ``omniprediction'', then the underlying $\baseip$ will win over all constant competitor forecasts in the above sense, whether these competitors are precise or imprecise.

\subsection{Reduction to Precise Forecasts via Maximum Entropy}
\label{sec:reductiontoprecise}
We have so far investigated propriety for the case of constant forecasts, which is also how propriety is formulated in the precise case. Yet our interest in practice lies with forecasts that take the features $X$ into account. In the precise case, the unconditional and the conditional viewpoints interact nicely due to additivity of the expectation, and it is possible to derive that for a weakly proper scoring rule (with propriety defined in the constant case) an optimal forecast is $f^*(x) \coloneqq P(\cdot|X=x)$ when $P(X=x)>0$, under a $P$ data model. In a sense, this forecast is exactly tuned to the underlying data model. In the case of an imprecise probability $\baseip$, it is not immediately clear what the corresponding ``conditional'' forecast is. Thus we now focus on the IP score objective in~\eqref{eq:overallobjective}; we recall:
\[
\ipscore(\ell,\baseip) \coloneqq \min_{\mQ : \mathcal{X} \rightarrow \setofips} \sup_{P \in \baseip} \E_{P}[S_\ell(\mQ(X(\omega)),\omega)].
\]
This is our notion of a ``Bayes risk'' for imprecise forecasts, as it quantifies the inherent difficulty of the task. We now show that under a suitable assumption on $\ell$ and $\baseip$ \textit{jointly}, we can restrict ourselves to optimizing over only precise forecasts and indeed then characterize this optimum. For trivial (constant) $X$, this yields a reduction of imprecise to precise forecasts for the unconditional case of Section~\ref{sec:propriety}.

The key will be to leverage the \textit{maximum entropy} bridge of \citet{grunwald2004game}, which we now work towards.
\begin{definition}
\label{def:condentropy}
    For an arbitrary loss $\ell \in \Lfuncs$, we can define a generalized conditional entropy of a probability $P \in \Delta^k$ as
\[
H_{\ell}(P|X) \coloneqq \E_{P}[S_\ell(P(\cdot|X),\omega)], \quad P \in \Delta^k.
\]
(Compare also \citep{xu2020continuity}).
\end{definition}
Intuitively, this quantity measures the variability in $P$ (as measured by the difficulty of forecasting when loss is measured by $\ell$) when allowing the forecaster to see the value of $X$. When $\ell$ is the log-loss, this quantity becomes the familiar conditional Shannon entropy.\footnote{\label{footnotefiniteness} Due to our finiteness assumption on $\mathcal{A}$ we cannot exactly recover the log-loss in our framework, but approximately. Note that usually conditional (Shannon) entropy is defined for random variables, \ie $H(Y|X)$, and not with a probability measure as $H(P|X)$, but the relation is straightforward.}
In the decision-theoretic view, the entropy concept expresses the inherent difficulty of a task: \textit{entropy is Bayes risk} \citep{osterreicher1993statistical,williamson2024information}. In this way, $H_\ell(P|X)$ is a conditional version of the decision-theoretic entropy of \citet{dawid1999coherent}. It turns out that, under a subtle condition, an optimal forecast in the imprecise case is simply an optimal precise forecast for the ``most difficult'' $P \in \cobar(\baseip)$.
\begin{proposition}
\label{prop:optimalforecast}
Let $\ell \in \Lfuncs$ and $\baseip \in \setofips$.
    Compute a maximum entropy probability over $\cobar(\baseip)$:\footnote{Any $P^*$ in the $\argmax$ will do, but there is no general uniqueness guarantee.}
     \begin{equation}
        \label{eq:maxent}
        P^* \in \operatorname{argmax}_{P \in \cobar(\baseip)} H_\ell(P|X).
     \end{equation}
If $P^*(X=x)>0 \; \forall x \in \mathcal{X}$ and $\argmin_{a \in \mathcal{A}} \E_{P^*(\cdot|X=x)}[\ell(a,\omega)]$ is a singleton for each $X=x$ (the action recommendation is unique), then the forecast $f^*(x) \coloneqq P^*(\cdot|X=x)$ is optimal up to weak inequality:
    \[
    \label{eq:ipscoreofoptimal}
        \ipscore(\ell,\baseip) 
= \Rup_{\baseip}\left(S_\ell(f^*(X),\omega)\right) = \E_{X,\omega \sim P^*}[S_\ell(f^*(X),\omega)].
\]
\end{proposition}
The proof of this result, which draws on the maximum entropy principle of \citet{grunwald2004game}, is in Appendix~\ref{app:prop:optimalforecast}. 
When $\Plow_{\baseip}(X=x)>0 \; \forall x \in \mathcal{X}$, then also $P^*(X=x)>0 \; \forall x \in \mX$. 
We emphasize that when computing the maximum entropy probability, it is crucial to consider the closed convex hull of $\baseip$; indeed $P^*$ can lie in the interior of $\cobar(\baseip)$. If the maximum entropy probability does not provide unique action recommendations, it can be that 
\[
\E_{P^*}[S_\ell(f^*(X),\omega)]<\ipscore(\ell,\baseip) < \Rup_{\baseip}\left(S_\ell(f^*(X),\omega)\right),
\] (for an example see the proof of~\ref{failurestrict3}). 

To compare to the precise case: under an \iid data model with probability $P$, an optimal forecast is $f^*(x)=P(\cdot|X=x)$ \textit{under any $\ell$} (due to the ``automatic'' weak propriety of $S_\ell$), which would amount to the action recommendation $a^*(x) = f^*(x)$ if we had $\mathcal{A}=\Delta^k$ and $\ell$ were already a strictly proper scoring rule.\footnote{In this paper, we cannot have $\ell$ be a proper precise scoring rule due to $|\mathcal{A}|<\infty$. See~\ref{footnotefiniteness} and Remark~\ref{remark:standardpropriety}.} Similarly, in the imprecise case, the optimal forecast of the maximum entropy principle corresponds to one of the probabilities $P \in \baseip$ in the same way, but $\ell$ determines which $P \in \baseip$ we pick. The condition that the action recommendations are unique is subtle, however. The proof of~\ref{failurestrict3} already includes such an example. We provide another example to illustrate this failure of uniqueness in Appendix~\ref{app:exampleofmaxentfailure}, meaning we cannot characterize the optimal forecast using Proposition~\ref{prop:optimalforecast}.

We remark that when taking $X$ to be trivial, $X(\omega)=c \; \forall \omega \in \Omega$, the findings here apply also to the unconditional case of Section~\ref{sec:propriety}. In this case, the generalized conditional entropy of Definition~\ref{def:condentropy} becomes a generalized unconditional entropy $H_\ell(P) \coloneqq \min_{a \in \mathcal{A}} \E_P[\ell(a,\omega)]$. 

\begin{remark}[Connection to standard setup and intuition for uniqueness condition] \normalfont
\label{remark:standardpropriety}
    Usually a loss function would depend on $\omega$ only through $Y(\omega)$, and a forecast would correspond to a probability on $\mathcal{Y}$, not on $\Omega = \mathcal{X} \times \mathcal{Y}$ as in our case. For example, for binary $\mathcal{Y} = \{0,1\}$ the log-loss is 
    \[
    \ell_{\text{log}} : \Delta^2 \times \mathcal{Y} \to \mR, \quad \ell_{\text{log}}(q,y) \coloneqq -y \log(q) - (1-y)\log(1-q). 
    \]
    The log-loss is strictly proper in the sense of a precise scoring rule $s : \Delta^{|\mathcal{Y}|} \times \mathcal{Y} \to \mR$, meaning 
    \[
    \E_{P}[s(P,Y)] \leq \E_{P}[s(Q,Y)],
    \]
    for $P,Q \in \Delta^{|\mathcal{Y}|}$, with equality only if $P=Q$. 
    Note here the possibly confusing terminology: a tailored \textit{scoring rule} $S_\ell$ depends on an underlying \textit{loss function} $\ell$, but when the action space is the simplex itself and $\ell$ formally fulfills strict propriety, then $S_\ell = \ell$ because $a_Q^* = Q$ \citep{dawid2014theory}. 
In this paper, we stick to finite action spaces since it simplifies some aspects of the mathematics and we believe that from a conceptual viewpoint there is no loss of generality. But if $\ell$ were a strictly proper scoring rule like the log-loss, then the uniqueness condition ``$\argmin_{a \in \mathcal{A}} \E_{P^*(\cdot|X=x)}[\ell(a,\omega)]$ is a singleton for each $X=x$'' in Proposition~\ref{prop:optimalforecast} would be satisfied, since then only $P^*(\cdot|X=x)$ would be in the $\argmin$.
Since strict propriety of $\ell$ implies that $a \mapsto \ell(a,y)$ is continuous on the interior of $\Delta^{|\mathcal{Y}|}$ for any fixed $y \in \mathcal{Y}$ \citep[Lemma 26]{mhammedi2018constantregretgeneralizedmixability}, we can then safely approximate $\Delta^{|\mathcal{Y}|}$ by a finite discretization in practice (\eg the set of floats in a computer) and apply Proposition~\ref{prop:optimalforecast}.

\end{remark}

\subsubsection{The Link to Distributional Robustness}
\label{sec:linktodro}
Observe that under the condition in Proposition~\ref{prop:optimalforecast}, the problem of finding an optimal forecast is reduced to optimizing over only precise forecasts:
\[ \min_{\mQ : \mathcal{X} \rightarrow \setofips} \Rup_{\baseip}\left(S_\ell(\mQ(X),\omega)\right)
     = \min_{Q : \mathcal{X} \rightarrow \Delta^k} \Rup_{\baseip}\left(S_\ell(Q(X),\omega)\right).
\]
This means that, under the uniqueness condition, our IP scoring rule objective turns out to be approximately equivalent to the objective considered in distributionally robust optimization (DRO), well-known in the machine learning community \citep{sagawa2019distributionally,duchi2018learning,frohlich2022risk}. 
To our knowledge, the connection between DRO and the maximum entropy principle has not yet been laid out in the machine learning literature. 

What still distinguishes the objective here from standard DRO is that we use the tailored scoring rule $S_\ell$. We might call this \textit{tailored DRO}, and our results in Section~\ref{sec:experiments} highlight the importance of the choice of loss function in DRO. 
Remark~\ref{remark:standardpropriety} tells us that if we use a strictly proper $\ell$ (in the precise sense) and a finite discretization of the simplex as our action space, then we can effectively apply Proposition~\ref{prop:optimalforecast} to obtain the optimal forecast. 
\textit{Prima facie} it seems unsettling that for optimal performance we do not need imprecision in the forecasts --- what is more fundamental is the imprecision in the data model, expressed by $\Rup_{\baseip}$. Observe, however, that the above statement holds \textit{for a fixed $\ell$}. We do not always know at training time the loss function of the agent, or we may want to enable decision making for multiple agents with different loss functions. Yet the minimizer in the DRO objective is tailored to the exact loss function, which at training time is usually a convenient surrogate loss function. In summary, we highlight that the choice of loss function matters in DRO, since this determines the maximum entropy probability.

\subsection{A Variety of Entropies}
\label{sec:entropies}
Recall the idea that entropy is Bayes risk (\eg \citep{williamson2024information}).
The IP score, our generalized Bayes risk, can be considered as a generalized conditional entropy notion for imprecise probabilities, which we might write as $\overline{H}_\ell(\baseip|X) \coloneqq \ipscore(\ell,\baseip)$, where $X$ is implicit on the right side. We may call it the \textit{imprecise conditional $\ell$-entropy} of $\baseip$. In contrast, the quantity $\operatorname{MaxEnt} \coloneqq \max_{P \in \cobar{P}} H_\ell(P|X)$, featuring in \eqref{eq:maxent}, could be called the \textit{maximum conditional $\ell$-entropy}. In the unconditional case, this has been indeed been proposed as a generalized entropy for imprecise probabilities \citep{abellan2003maximum,abellan2005upper,shaker2021ensemble}. Importantly, we have seen that possibly $\operatorname{MaxEnt} < \overline{H}_\ell(\baseip|X)$, although they sometimes coincide. Our interpretation of the situation is that the IP score $\overline{H}_\ell(\baseip|X)$ is more suited as a generalization of the entropy concept, as it quantifies the inherent difficulty of the prediction task.
For clarification of these concepts, we specialize the insight of Proposition~\ref{prop:optimalforecast} to the unconditional case of constant forecasts.
\begin{definitionprop}
    Let $\baseip \in \setofips$. In analogy to $\overline{H}_\ell(\baseip|X)$,  we call $\overline{H}_\ell(\baseip) \coloneqq \Rup_{\baseip}(S_\ell(\baseip,\omega))$ the imprecise $\ell$-entropy of $\baseip$. Observe that it corresponds to the IP score of the optimal constant forecast under a $\baseip$ data model due to weak propriety (Proposition~\ref{eq:weakpropriety}). Let $P^* \in \argmax_{P \in \cobar(\baseip)} \E_P[\ell(a_P^*,\omega)]$ be a maximum entropy probability. If it recommends a unique action, meaning $\{a^*\} = \argmin_{a\in \mathcal{A}} \E_{P^*}[\ell(a,\omega)]$ is a singleton, then $\overline{H}_\ell(\baseip) = \Rup_{\baseip}(S_\ell(P^*,\omega)) = \E_{P^*}[\ell(a^*,\omega)]$. 
\end{definitionprop}
We view the unconditional quantity as a generalization of the decision-theoretic entropy of \citet{dawid1999coherent} to the imprecise case.

\subsection{The Generalized Bayes Rule}
\label{sec:gbrforecast}
The optimal forecast $f^*$ of Proposition~\ref{prop:optimalforecast}, if the condition holds, picks exactly one $P \in \baseip$ and conditions it. A more conservative, intuitive approach would be to simply forecast the whole set of conditionals for a given $X=x$. This is effectively a generalization of Bayes rule to a set of probabilities, and can be seen as conditioning the data model. As a prerequisite for Section~\ref{sec:ipcalibration},
we first define it for the general case of an arbitrary partition $\mB$ of $\Omega$ and then specialize. For a set $B \subseteq \Omega$, we denote its indicator function by $\chi_{B}(\omega) \coloneqq 1$ if $\omega \in B$, and $0$ otherwise.
\begin{definition}[{\citep[Chapter 6]{walley1991statistical}}]
\label{def:generalgbr}
Let $\baseip \in \setofips$. For a partition $\mathcal{B}$ of $\Omega$, where $\Plow_{\baseip}(B)>0 \; \forall B \in \mathcal{B}$, define 
\[
\mP_{|B} \coloneqq \{P(\cdot|B) : P \in \mP\}, \quad \gbr_{\mP}(Z|B) \coloneqq \Rup_{\mP_{|B}}(Z), \quad B \in \mB, \; Z \in \linfty.
\]
It follows that 
$\gbr_{\mM}(Z|B) = \min\{\alpha \in \mR: \Rup_{\baseip}\left(\chi_B(Z-\alpha)\right) = 0\}$. 
Due to the 1:1 relation, we refer to either $\mP_{|B}$ or $\gbr_{\baseip}(\cdot|B)$ as the \textit{generalized Bayes rule} (GBR).
\end{definition}
We will be mostly interested in the special case where the partition is induced by the features $X$. In this case we obtain:
\begin{definition}
    Let $\baseip \in \setofips$. Define, if $\Plow(X=x)>0 \; \forall x \in \mathcal{X}$:
    \[
    \mPx \coloneqq \{P(\cdot|X=x) : P \in \baseip\}, \quad \gbr_{\baseip}(Z|x)=\Rup_{\mPx}(Z), \quad Z \in \linfty.
    \]
    The associated forecast is
\[
\mP_{|X}(\omega) \coloneqq \mP_{|x} \text{ if } X(\omega)=x.
\]
    We refer to $\mP_{|X}$ as the \textit{generalized Bayes rule} (GBR) forecast.
\end{definition}
While this forecast has an intuitive appeal, one can observe by example that it need not be optimal in terms of IP score (see Appendix~\ref{app:exampleofmaxentfailure}). Since $f^*(x) \in \mPx$ ($f^*$ of Proposition~\ref{prop:optimalforecast}, the GBR is a more conservative (pessimistic) forecast than $f^*$. However, as we now shift the focus to a distinct goal in decision making, we will see what the strength of the GBR is, making it attractive again. In the next section, we present a general account of \textit{calibration} in terms of entropy, which applies to both the precise and the imprecise case, and then relate it to familiar notions of calibration.

\section{IP Calibration}
\label{sec:ipcalibration}
In a recent paper, \citet{frohlich2024insights} put forward the thesis that there are strong conceptual parallels between \textit{insurance} and fair machine learning. While the authors focused on fairness and its interplay with uncertainty, their arguments suggest that the relevance of insurance as an analogy extends beyond fairness to issues of uncertainty in general. 
This heuristic motivates us to consider decision making with imprecise forecasts in an abstract insurance context, in order to arrive at a notion of calibration for imprecise forecasts. More generally, we are motivated here by the game-theoretic account of probability (see Section~\ref{sec:relatedworkscoringrules}).

\subsection{Calibration as Actuarially Fair Insurance}
What was lacking in the previous section was a justification for why the agent might \textit{trust} the forecasts. So far, we have assumed that the agent must choose an action based on the forecasts without any guarantees for the quality of the forecasts. As such, the burden of uncertainty lies fully on the agent. Insurance manages uncertainty by pooling losses, which in our case means shifting this burden to the forecaster (the insurer), whereas the agent who repeatedly makes decisions (or a group of distinct agents) corresponds to the insurance pool of policyholders. To instantiate the analogy, when the agent has chosen the action $\aopt$ as recommended by the forecaster, the loss to be incurred is $\ell(\aopt,\omega)$, and this is the uncertain claim that the forecaster insures. Putting aside normative questions of solidarity (risk subsidizing), a standard desideratum for insurance is \textit{actuarial fairness} \citep{arrow1963uncertainty}.
\begin{desideratum}
    The forecaster should offer \textit{actuarially fair} insurance for the uncertain loss $\tilde{S}_{\ell}(\omega)=\ell(\aoptXom,\omega)$. Actuarial fairness here means that in the long run, the forecaster neither loses nor profits from offering insurance \textit{under the data model}. %
\end{desideratum}
In ML practice, the burden of uncertainty will still usually lie on the agent; but if we know that forecasts are actuarially fair, there is a certain adequateness to their associated uncertainty. Within the ML research community, this philosophy aligns with \citet{yoo2019learning} and \citet{kirchhof2024pretrained}, who ground the quality of uncertainty estimates in accurate loss prediction.
For fixed $\omega \in \Omega$, consider the decomposition 
\[
\Sl(\omega) = \left(\Sl(\omega) - \Rup_{\mQ(\omega)}(\omega' \mapsto \Sl(\omega'))\right) + \Rup_{\mQ(\omega)}(\omega' \mapsto \Sl(\omega')),
\]
where we highlight the dependence on $\omega$ \versus $\omega'$. 
A gamble of the form $Z-\Rup(Z)$, $Z \in \linfty$, is known as a \textit{marginally desirable gamble} in the IP literature \citep{augustin2014introduction}, since a forecaster with subjective upper expectation $\Rup$ should be indifferent to it. 
Such gambles can be used to evaluate forecasts, an idea discussed in depth by \citet{derr2024four}. 
In our case, the forecaster will shoulder the uncertain gamble $\Sl(\omega) - \Rup_{\mQ(\omega)}(\Sl)$ and the agent will shoulder the constant\footnote{By \textit{constant} loss, we here mean that the price of insurance can be specified after learning $X=x$, which the agent then shoulders, but before all information in $\omega$ is revealed.} loss $\Rup_{\mQ(\omega)}(\Sl)$, which is the price for the insurance as specified by the forecaster. The quantity $\Rup_{\mQ(\omega)}(\Sl)$ represents a generalization of the entropy concept to the imprecise case (recall Section~\ref{sec:entropies}), quantifying the uncertainty inherent in the task, with evaluation by $\ell$, from the subjective viewpoint of the forecaster.
If the forecaster believes their own forecast in the sense that if $X(\omega)=x$, then $\Plow_{\mQ(\omega)}(X=x)=1$, then indeed $\Rup_{\mQ(\omega)}(\Sl)=\overline{H}_\ell(\mQ(\omega))$ is the imprecise $\ell$-entropy of $\mQ(\omega)$.\footnote{The reason is that under this condition, the forecast is constant according to $\Rup_{\mQ(\omega)}$, and then the quantity $\Rup_{\mQ(\omega)}(\Sl)$ is of the form $\Rup_{\mQ}(S_\ell(\mQ,\omega))$ for $\mQ \in \setofips$.}

In the transaction above, the agent desires to have as little constant loss as possible (cheap insurance), whereas the forecaster wants to avoid bankruptcy. The tension is resolved by the principle of actuarial fairness. \citet{frohlich2024insights} have demonstrated a tight relationship between actuarial fairness and calibration. We now use the term \textit{(IP) calibration}, but it might be equivalently called \textit{(IP) actuarial fairness}.

\subsection{Defining IP Calibration}
\begin{definition}[IP calibration without groups]
\label{def:actuarialfairnesswithoutgroups}
Let $\ell \in \Lfuncs, \baseip \in \setofips$. The forecast $\mQ$ is $\ell$-calibrated (without groups) under a $\baseip$ data model if, with $\Sl(\omega)=S(\mQ(\omega),\omega)$:
    \begin{equation}
        \label{eq:actuarialfairness}
        \Rup_{\baseip}\left(\Sl(\omega) - \Rup_{\mQ(\omega)}(\Sl)\right) = 0.
    \end{equation}
    We say that $\mQ$ is sub-calibrated if \eqref{eq:actuarialfairness} holds with a ``$\leq$'' inequality.
\end{definition}
Throughout this section, we continue to assume that the data model is parameterized by $\baseip$. 
Intuitively, calibration as a criterion is about a matching of the amount of loss to be expected \textit{under the data model} and the entropy \textit{with respect to the forecaster}. 
A calibrated forecaster makes adequate entropy assessments, hence accurately predicts the loss to be incurred.
This criterion in itself is rather weak. For instance, the constant forecast $\baseip$ clearly achieves this. 
Central in insurance, particularly in debates revolving around fairness, is the question of forming groups. An individual is then treated as equivalent to the ``average human'' of their group \citep{krippner2022person}.
In the above definition, there is only a single group. The condition can be strengthened by introducing a partition into groups.

We first illustrate this logic with a highly relevant example, the partition that is induced by the action recommendations. Assume the forecaster announces $\mQ(x)$, based on which the agent chooses the action $a_{\mQ(x)}^*$. 
The agent would now like to know which loss they get in terms of worst-case expected loss under the data model, conditional on this action recommendation; then forecasts are trustworthy when their uncertainty estimates are actuarially fair.
Formally, the action-induced partition is 
\[
\actionpartition \coloneqq \bigcup_{a \in \mathcal{A}} \actionpartelement; \quad \actionpartelement \coloneqq \{ \omega \in \Omega:  a_{\mQ(\omega)}^* = a\}. 
\]
This leads us to the following definition of IP calibration with respect to a partition. 
\begin{definition}[IP calibration with groups]
\label{def:calibrationwithpartition}
Let $\ell \in \Lfuncs, \baseip \in \setofips$ and $\mathcal{B}$ be a partition of $\Omega$. 
    We say that a forecast $\mathcal{Q}$ is $\ellB$-calibrated under a $\baseip$ data model if, with $\Sl(\omega)=S_\ell(\mQ(\omega),\omega)$:
 \[
    \forall B \in \mathcal{B}: \Rup_\baseip\left(\chi_B(\omega) \left(\Sl(\omega) - \Rup_{\mQ(\omega)}\left( \Sl \right) \right) \right) = 0.
    \]
    We say that $\mQ$ is $\ellB$-sub-calibrated if the above holds with a ``$\leq$'' inequality.
\end{definition}
In comparison to the formulation without groups, we now demand adequate entropy assessments \textit{conditionally}. 
IP calibration has a useful structural property. We say that a partition $\mathcal{B}_1$ is coarser than $\mathcal{B}_2$ if every set in $\mathcal{B}_2$ is a subset of some set in $\mathcal{B}_1$ (\eg \citep{holtgen2023richness}). We then have the following implication.
\begin{proposition}
\label{prop:partitionsubcal}
Let $\ell \in \Lfuncs$ and $\mathcal{B}_1,\mathcal{B}_2$ be partitions of $\Omega$, with $\mathcal{B}_1$ coarser than $\mathcal{B}_2$.
Then, if $\mQ$ is $(\ell,\mathcal{B}_2)$-sub-calibrated, it is $(\ell,\mathcal{B}_1)$-sub-calibrated. In particular, $(\ell,\mB_2)$-calibration implies $(\ell,\mB_1)$-sub-calibration.
\end{proposition}
The proof is in Appendix~\ref{app:prop:partitionsubcal}.
Why care about sub-calibration? Due to the loss orientation of the scoring rule $S_\ell$, our situation is an asymmetric one; we choose to prioritize \textit{not underestimating} the loss, which aligns with the goal of deploying robust and safe machine learning systems. In insurantial terms this would mean ensuring that the price for insurance is at least sufficient to cover the losses, but potentially more pessimistic.

Since calibration embodies trustworthiness of uncertainty estimates, it appears as an important goal. Yet we observe that it can diverge from the goal of Section~\ref{sec:ipscoring}, \ie recommending actions with favorable consequences.
\begin{proposition}
\label{prop:optimalforecastnotcal}
    There exist $\ell \in \Lfuncs$ and $\baseip \in \setofips$ so that the condition of Proposition~\ref{prop:optimalforecast} applies, and the optimal forecast $f^*$ of Proposition~\ref{prop:optimalforecast} fails to be $\ell$-sub-calibrated and therefore also fails to be $(\ell,\mB_{f^*;\mA})$-sub-calibrated.
\end{proposition}
As a proof, see the example in Appendix~\ref{app:exampleofmaxentfailure}, where this failure is illustrated both in an unconditional (without features) fashion, as well as in a conditional (with features) fashion. 
Note that a failure of sub-calibration implies a failure of calibration.
We remark that it can be that $\mQ \notsamehull \mQ'$ are both optimal in \eqref{eq:overallobjective} (\eg if they recommend the same actions), but one of them is calibrated and the other is not (see Appendix~\ref{app:exampleofmaxentfailure}). 
This negative result means that the optimal forecast can be overly optimistic, as it underestimates the loss to be incurred on some $P \in \setofips$, which can happen on both the unconditional level (without groups) and the conditional (group) level. In other words, the price for insurance is too low, leading to long-run bankruptcy.

\subsection{Calibration and the GBR}
On the other hand, the GBR has an attractive calibration guarantee, as it is more conservative. First, we write the calibration condition in terms of the GBR itself.

\begin{remark}
\label{remark:rewritingcalgbr}
    With $\Sl(\omega)=S_\ell(\mQ(\omega),\omega)$, the IP calibration condition in Definition~\ref{def:calibrationwithpartition} can be equivalently rewritten as
    \[
    \forall B \in \mathcal{B}: \gbr_{\baseip}( \Sl(\omega) - \Rup_{\mQ(\omega)}\left( \Sl \right) |B) = 0.
    \]
\end{remark}
\begin{proof}
    From $\gbr_{\baseip}(Z|B) = \min\{\alpha \in \mR: \Rup_{\baseip}\left(\chi_B(Z-\alpha)\right) = 0\}$, it follows that, with $Y(\omega) \coloneqq \Sl(\omega) - \Rup_{\mQ(\omega)}\left( \Sl \right)$, $\gbr(Y|B)=0 \Leftrightarrow \Rup_{\baseip}(\chi_B(\omega) Y(\omega)) = 0$, which is the definition of calibration with groups.
\end{proof}
Rewriting the calibration criterion in terms of the GBR suggests a tight conceptual connection. Comparing two forecasts is particularly of interest when they ``get to see the same information'', intuitively; in our context, this means usually the features $X$ but we generalize.

\begin{definition}
    Let $\mB$ be a partition of $\Omega$ and $\mQ_{\mB} : \mathcal{X} \to \setofips$ be a forecast. We say that $\mQ_{\mB}$ is $\mB$-measurable if it is constant on each $B \in \mB$ (following terminology of \citet[Section 6.2.5]{walley1991statistical}).
    We then write $\mQ_B \in \setofips$ for the value it takes on $B \in \mB$.
\end{definition}

Using this definition, we find that, for any loss $\ell$, the GBR is $\ell$-calibrated and its entropy provides a lower bound to that of other calibrated forecasts. We write $\mathfrak{B}$ for the set of partitions of $\Omega$ with the property that each element of the partition has positive lower probability under the data model, meaning $\mathcal{B} \in \mathfrak{B}$ if $\Plow_{\baseip}(B)>0 \; \forall B \in \mathcal{B}$.
\begin{proposition}
\label{prop:gbrcalandentropybound}
Let $\ell \in \Lfuncs, \baseip \in \setofips$ and $\mB \in \mathfrak{B}$ a partition of $\Omega$. 
Then the GBR forecast is $\ellB$-calibrated: 
    \[
    \forall B \in \mathcal{B}: \gbr_{\baseip}(S_\ell(\mP_{|B},\omega) - \gbr_\baseip\left(S_\ell(\mP_{|B},\omega) | B\right) |B) = 0.
    \]
    Moreover, if $\mQ_{\mB}$ is any other $\mB$-measurable forecast which is $\ellB$-calibrated, then
    \[
    \label{eq:gbrentropylowerbound}
    \forall B \in \mB: \overline{H}_\ell(\baseip_{|B}) = \gbr_\baseip\left(S_\ell(\mP_{|B},\omega) | B\right) \leq \Rup_{\mQ_{B}}(S(\mQ_B,\omega)) = \overline{H}_\ell(\mQ_B). %
    \]
    In words, for each $B \in \mB$ the imprecise $\ell$-entropy of the ``\textit{conditional}'' data model parameterized by $\baseip_{|B}$ is a lower bound for the imprecise $\ell$-entropy of any calibrated competitor forecast $\mQ_B$.
    In addition, there exist $(\ell,\baseip,\mB,\mQ_\mB)$ satisfying the assumptions so that the inequality in \eqref{eq:gbrentropylowerbound} is strict.\footnote{Meaning that there exists a pair $(\ell,\baseip,\mB,\mQ_\mB)$, so that the forecast $\mQ_\mB$ is also $\ellB$-calibrated but $\overline{H}_\ell(\baseip_{|B}) < \overline{H}_\ell(\mQ_B) \; \forall B \in \mB$.}
\end{proposition}
The proof is in Appendix~\ref{app:prop:gbrcalandentropybound}. Elaborating on this result, if a forecast is calibrated, we can use its entropy to get a bound on the IP score.

\begin{proposition}
\label{prop:scoreentropybound}
Let $\ell \in \Lfuncs, \baseip \in \setofips$ and $\mB \in \mathfrak{B}$ a partition of $\Omega$. 
    Let $\mQ_\mB$ be a $\mB$-measurable and $\mB$-calibrated forecast. Then
    \[
    \Rup_{\baseip}(S_\ell(\mQ_\mB(X),\omega)) \leq \sum_{B \in \mB} H_\ell(\mQ_{B}) \cdot \Rup_{\baseip}(\chi_B).
    \]
\end{proposition}
The proof is in Appendix~\ref{app:prop:scoreentropybound}.

For a fixed loss $\ell$, a distinct ($\notsamehull$) forecast could yield the same entropy assessments as the GBR. 
When considering all losses, however, the GBR has a nice uniqueness guarantee.
\begin{proposition}
\label{prop:gbruniqueness}
Let $\ell \in \Lfuncs, \baseip \in \setofips$ and $\mB \in \mathfrak{B}$ a partition of $\Omega$. 
Let $\mQ_\mB$ be any $\mB$-measurable forecast. Assume without loss of generality that $\mQ_B$ and $\baseip_{|B}$ are closed convex $\forall B \in \mB$. Then
    \begin{enumerate}
        \item \label{item:gbruniqueness1} (More optimistic than GBR): If $\mQ_B \subseteq \baseip_{|B} \; \forall B$ and $\mQ_B \subsetneq \baseip_{|B}$ for at least one $B$, then $\mQ_\mB$ cannot be $\ellB$-sub-calibrated for all loss functions.
        \item \label{item:gbruniqueness2}  (More pessimistic than GBR): If $\baseip_{|B} \subseteq \mQ_B \; \forall B$, then $\mQ_\mB$ is $\ellB$-sub-calibrated.
        \item \label{item:gbruniqueness3}  (Different to GBR): If $\exists B\in \mB: \baseip_{|B} \neq \mQ_B$, then
        $\mQ$ cannot be $\ellB$-calibrated for all loss functions.
    \end{enumerate}
\end{proposition}
The proof, based on the general monotonicity property that if $\mQ \subseteq \mQ'$, then $\Rup_{\mQ}(Z) \leq \Rup_{\mQ'}(Z)$ for any $Z \in \linfty$, is in Appendix~\ref{app:prop:gbruniqueness}.

The take-away of this section is that calibration, which expresses trustworthiness of uncertainty estimates in terms of entropy, is a goal which is distinct to the goal of recommending actions with favorable consequences. In the classical, precise case, this distinction has been observed on a formal level in the decomposition of proper scoring rules \citep{degroot1983comparison,brocker2009reliability}, but the key difference is that these goals are aligned in the precise case (the Bayes classifier is the optimal forecast and it is calibrated), but they can diverge in the imprecise case. In particular, the optimal forecast need not be calibrated.

\section{Experiments}
\label{sec:experiments}
In the present paper we have proposed a general conceptual framework for the evaluation of imprecise probabilistic forecasts, which opens up ample room for practical investigations and algorithmic developments in the future. Many questions arise (see the discussion in Section~\ref{sec:discussion}) which require more detailed attention than we can give here. Our focus lies therefore on illustrative experiments, with deliberately simple model architecture (\ie logistic regression) to avoid confounding due to contingencies of the training process. Our focus lies on demonstrating theoretical insights in practice. Others works have already shown the usefulness of group DRO in deep learning practice (\citep{sagawa2019distributionally}). We here also stick to the group DRO setting as our imprecise data model due to its high relevance. Thus, to connect back to the presented hierarchy in Section~\ref{sec:hierarchyofimprecision}, in this setting the imprecision is assumed to be fundamental to the data model, and thus does not express our epistemic uncertainty or risk aversion. Recall (Section~\ref{sec:hierarchyofimprecision}) that the evaluation of imprecise forecasts may also be studied under other (possibly even a precise) data model. 
Our aim here is to showcase subtle potential pitfalls with (im)precise forecasts under group DRO.
In particular, we illustrate how the choice of the loss function \textit{can}  matter in group DRO, and how the two key qualitities of good forecasts, embodied by scoring rules and calibration, \textit{can} diverge in the group DRO setting; our experiments should be interpreted with an existential quanitifier instead of a universal quantifier in mind. Code for reproducing our experiments can be found at \url{https://github.com/froec/ip_scoring_rules_experiments}.

\subsection{Data and Decision Problems}
\label{sec:dataanddecisions}
Inspired by \citet{ding2021retiring}, the first dataset that we use, \textsc{acs pums}, stems from the american community survey. In our variant, the binary classification task is to predict whether a person has a job ($Y=1$), or is unemployed/not on the labor market ($Y=0$), based on demographic information about the person (their age, marital status, gender \etc).
We form the groups based on membership to the $50$ U.S. states. The sample size across groups is highly uneven, with the largest group California having $302{,}640$ examples\footnote{We remark that a datum of the \textsc{acs pums} dataset does not correspond to a single person, but rather to a sort of \textit{statistical individual.}}, contrasting with the smallest, Wyoming, having only $4{,}552$ examples. For each group, we have a training as well as a test set.
In the group DRO setting, we keep the data from these groups (states) separate; in the empirical risk minimization (ERM) setting, we stack the data of the groups.

The other datasets share the same structure, each paired with a binary classification task ($Y=0$ or $Y=1$).
The next dataset we use is \textsc{framingham}, where the goal is to predict 10-year heart disease risk based on information such as age, blood pressure, having diabetes \etc; We form two groups based on thresholding age at $60$.
Finally, loosely inspired by \citep{sagawa2019distributionally}, on \textsc{celeba}, we use the embeddings of a pre-trained (on ImageNet) Resnet50 to detect whether a person wears an earring or not. We form two groups based on gender (``male'' or ``not male'', where the latter includes non-binary identities).
For detailed information about the datasets see Appendix~\ref{app:data}. In the main paper, we report results for \textsc{acs pums} and shift results for the other datasets to Appendix~\ref{app:moreresults}. 

Besides the data model, the second key ingredient in our framework is the emphasis on the decision-making context. In machine learning practice, it is common to choose a proper scoring rule as a \textit{surrogate loss}, such as the log-loss (also known as cross-entropy loss) or Brier loss (squared loss) at training time. Crucially, these are continuous and differentiable which enables training, and they are moreover symmetric. Test performance is often summarized using accuracy on the test set. Accuracy is, up to a constant factor, a special case ($c=0.5$) of the following family of loss functions:
\begin{definition}[{\eg \citep{buja2005loss}\footnote{What \citet{buja2005loss} calls ``cost-weighted
misclassification losses'' is the tailored scoring rule $S_{\ell_c}$ induced by $\ell_c$.}}]
    The cost-sensitive loss $\ell_c$ with parameter $c \in (0,1)$ and action space $\mathcal{A} \coloneqq \{0,1\}$ is defined by $\ell_c(0,Y(\omega)=1) \coloneqq 1-c$, $\ell_c(1,Y(\omega)=0) \coloneqq c$, and $0$ otherwise.
\end{definition}
Importantly, these cost-sensitive losses form a ``basis'' of all binary proper scoring rules \citep{schervish1989general}. 
The parameter $c$ controls the asymmetry in the decision problem: often, a false positive is much more consequential than a false negative, or conversely. For example, in heart disease prediction,  accuracy might be a poor measure of performance, and we might be inclined to choose a small $c$, which means we get high loss when falsely predicting ``has no risk of heart disease''. While cost-sensitive losses are flexible, they are unfortunately hardly usable for training due to discontinuity: with only the binary action space $\mathcal{A}=\{0,1\}$, a small change in the forecast can imply a big change in the resulting loss.
We therefore use a similarly parameterized family of continuous surrogate losses at training time due to \citet{winkler1994evaluating}, which we here specialize to use the Brier loss as the base.\footnote{We also remark that \citet{buja2005loss} have proposed an asymmetric family of proper scoring rules based on the Beta distribution, with analytical form for some values of the asymmetry parameters.}
\begin{definition}[{\citep{winkler1994evaluating}}]
    With $\mathcal{A} \coloneqq [0,1]$ and $c \in (0,1)$, an asymmetric, strictly proper scoring rule\footnote{Recall from Section~\ref{sec:linktodro} \citep{dawid2014theory} that if $\ell$ is a precise, strictly proper scoring rule with $\mathcal{A}=\Delta^{|\mathcal{Y}|}$ (here $\mathcal{A} = [0,1]$ due to the binary setting), then $S_\ell=\ell$. That is, we could call $\ell$ either (proper) ``loss'' or (proper) ``scoring rule'' \citep{williamson2023geometry}. In our view, it is beneficial to generally distinguish the roles of $\ell$ and $S_\ell$ though.} can be defined as:
    \[
    \ellwc(a,Y(\omega)=y) = 1 - \frac{\ell^2(a,y) - \ell^2(c,y)}{T(c,a)}; \quad  T(c,a) \coloneqq \begin{cases}
        -\ell^2(c,1) & a \geq c \\
        -\ell^2(c,0) & a < c
    \end{cases}; \quad \ell^2(a,y) \coloneqq (a-y)^2.
    \]
\end{definition}
Note that for any $c \in (0,1)$, $a \mapsto \ellwc(a,y)$ is continuous for fixed $y\in \{0,1\}$, even at $a=c$. 
To improve optimizer behaviour near the point $c$ of non-differentiability, in practice we replace the step function $T$ by a very close sigmoid approximation $\tilde{T}$ (see Appendix~\ref{app:sigmoidapprox}).  
\begin{remark}\normalfont
In the theoretical framework we have presented, the focus was on the idealized case where we minimize over \textit{all} forecasts $\mQ: \mX \to \Delta^k$. This is of course infeasible in practice, and we thus use a constrained model class throughout. Concretely, in our experiments, we train simple models consisting of a linear layer followed by a sigmoid (the output of which is considered as a precise probabilistic forecast) and then a loss function. We denote the respective constrained entropy quantities as $\hat{H}_\ell$, and the (precise) model class as $\operatorname{Lin} \coloneqq \{Q : \mX \to \Delta^2 : Q \text{ is linear followed by sigmoid}\}$. We restrict ourselves to $\Delta^2$ since our loss functions will depend on $\omega$ only through $Y$ and we now assume that the forecaster trusts the $X=x$ information.
\end{remark}

To compare the associated decision problems, we plot the generalized unconditional entropies $\hat{H}_{\ell_c}$ and $\hat{H}_{\ellwc}$ (obtained by taking constant $X$ in Definition~\ref{def:condentropy}).  
Both the entropies of $\ell_c$ and $\ellwc$ have their maximum at $c$; in this way, $\ellwc$ can be viewed as a surrogate loss for $\ell_c$ (see Figure~\ref{fig:entropycomparison}). We will therefore use $\ellwc$ at training time when we are actually interested in the test loss $\ell_c$. We caution that $a \mapsto \ell_{W;c}(a,y)$ need not be convex.
Since $\ell_{W;c}$ is strictly proper, Proposition~\ref{prop:optimalforecast} can be applied approximately (see Remark~\ref{remark:standardpropriety}), and we can therefore follow the maximum entropy principle.

\subsection{Models and Training}
\label{sec:modelsandtraining}
We now train different models on the \textsc{acs pums} as well as the other datasets (Appendix~\ref{app:moreresults}). 
Again and throughout the paper, as the base of all our models we use a linear regression followed by a sigmoid which outputs a precise probabilistic forecast, followed by a loss function. This means that for the log-loss, we recover the familiar logistic regression, and otherwise we get asymmetric variants of it.
First, we train a standard logistic regression model (\textbf{ERM(log)}), using the log-loss, on the full data, meaning the stacked data of all states. We also train group DRO models according to Proposition~\ref{prop:optimalforecast} for different loss functions.  Our training follows the maximum entropy principle:
\[
 \max_{\lambda \in \Delta^G} \min_{Q \in \operatorname{Lin}} \E_{\hat{P}_\lambda}\left[S_\ell(Q(X),Y)\right], \quad \hat{P}_\lambda \coloneqq \sum_{i=1}^G \lambda_i \hat{P}_i.
\]
where $\hat{P}_i$, $i=1..G$, is the empirical distribution (training data) for the $i$-th group. Here, we express the closed convex hull as a weighting (by $\lambda$) of the extreme points. Our loss functions depend on $\omega$ only through $Y$, hence a forecast can be identified with a mapping $\mQ: \mX \to \Delta^2$ for binary $Y$.

Our approach to training is naïve: at iteration $t=1..N_{\text{outer}}$, our forecast $Q_{\theta_t}$ is parameterized by $\theta_t$ and the weighting is $\lambda_t$. We perform an inner minimization loop for $N_{\text{inner}}$ number of iterations which updates $\theta_t$. At the end of this, the gradient for $\lambda_t$ is simply the vector of mean group losses, \ie $\nabla \lambda_t = [\E_{\hat{P}_1}[S_\ell(Q_{\theta_t}(X),Y)], ..]^\intercal$.
To update $\lambda_t$ so as to enforce $\lambda_t \in \Delta^k$ (unit L1 norm and positivity), we use exponentiated gradient descent \citep{kivinen1997exponentiated,wei2023distributionally}:
\begin{align}
    \lambda_{t+1} \coloneqq \lambda_t \cdot \exp\left(- \eta \cdot \nabla \lambda_t \right),\\
    \lambda_{t+1} \coloneqq \lambda_{t+1} / \sum \lambda_{t+1},
\end{align}
with some learning rate $\eta>0$. In a DRO context, exponentiated gradient descent to optimize the group weights has already been employed by \citet{wei2023distributionally}. Additional details on our procedure using batches can be found in Appendix~\ref{app:training}.
We train a DRO model for each of the losses $\ell \in \{\ell_{\text{log}}, \ell_{W;0.1},\ell_{W;0.3},\ell_{W;0.7},\ell_{W;0.9}\}$. We refer to the models as \textbf{DRO(log)} and \textbf{DRO($c$)}, respectively.

For the \textbf{GBR}, we choose a naïve estimation method: we fit a logistic regressor, using the log-loss, for each of the groups, and then take the set of these precise forecasts as our imprecise forecast. Setting aside the constraint of the linear model class, this would recover in the limit of infinite data the group-wise conditional probabilities due to propriety of the log-loss.

\subsection{Evaluation via IP Score}
\begin{table}[]
\footnotesize
    \centering
\begin{tabular}{lllll}
\toprule
 & $\ell_{0.1}$ & $\ell_{0.3}$ & $\ell_{0.7}$ & $\ell_{0.9}$ \\
Predictor &  &  &  &  \\
\midrule
GBR & 0.0492 & 0.1259 & 0.1475 & 0.0614 \\
DRO($0.1$) & \textbf{0.0447} & 0.1044 & 0.1343 & 0.0769 \\
DRO($0.3$) & 0.0456 & \textbf{0.0998} & 0.1350 & 0.0748 \\
DRO($0.7$) & 0.0473 & 0.1144 & \textbf{0.1242} & 0.0589 \\
DRO($0.9$) & 0.0474 & 0.1139 & 0.1315 & \textbf{0.0574} \\
ERM(log) & 0.0466 & 0.1114 & 0.1329 & 0.0715 \\
DRO(log) & 0.0463 & 0.1071 & 0.1299 & 0.0637 \\
\bottomrule
\end{tabular}
\quad \quad 
\begin{tabular}{lllll}
\toprule
 & $\ell_{0.1}$ & $\ell_{0.3}$ & $\ell_{0.7}$ & $\ell_{0.9}$ \\
Predictor &  &  &  &  \\
\midrule
GBR & 0.0504 & 0.1267 & 0.1440 & 0.0609 \\
DRO($0.1$) & \textbf{0.0463} & 0.1054 & 0.1364 & 0.0779 \\
DRO($0.3$) & 0.0474 & \textbf{0.1050} & 0.1349 & 0.0743 \\
DRO($0.7$) & 0.0493 & 0.1146 & \textbf{0.1294} & 0.0593 \\
DRO($0.9$) & 0.0493 & 0.1146 & 0.1310 & \textbf{0.0579} \\
ERM(log) & 0.0487 & 0.1120 & 0.1337 & 0.0709 \\
DRO(log) & 0.0476 & 0.1104 & 0.1331 & 0.0632 \\
\bottomrule
\end{tabular}

    \caption{Train (left) and test (right) IP Scores of different forecasts (rows), evaluated under different loss functions (columns) on \textsc{acs pums} data.}
    \label{tab:ipscoresacs}
\end{table}
We now subject these trained models to an evaluation by the losses $\{\ell_{0.1},\ell_{0.3},\ell_{0.7},\ell_{0.9}\}$, where the action space is now $\mathcal{A}=\{0,1\}$. Note the difference between \textit{training} a model using some $\ell_{W;c}$ and \textit{evaluating} by some $\ell_{c'}$, with possibly $c \neq c'$.
Evaluation, in any case, means selecting the optimal action $a_{\mQ(X)}^*$ based on the evaluation loss function, which may not coincide with the loss function used at training time. 
However, Table~\ref{tab:ipscoresacs} shows that indeed, $\ell_{W;c}$ is a reasonable surrogate loss to use in place of $\ell_{c}$. Recall that the IP score of the forecast $\mQ$ under a $\baseip$ data model is $\Rup_{\mP}(S_\ell\left(\mQ(X(\omega)),\omega\right))$. The \textit{train data model} is $\hat{\baseip} \coloneqq \{\hat{P}_i : i=1..G\}$, and similarly, the \textit{test data model} $\hat{P}_{\text{test}}$ consists of the test sets (their empirical distributions) of the groups.
Typically (this happens in most of our cases), the model with the best IP score under the training data model with $\ell_c$ is the \textbf{DRO(c)} model. Due to generalization gaps, under the test data model and $\ell_c$, it sometimes happens that the best performing model is not the \textbf{DRO(c)} model (see Appendix~\ref{app:moreresults}). In some cases, the \textbf{GBR} in fact performs best in terms of IP score under the test data model.

We reiterate the key insight: for DRO, the choice of loss function can indeed matter in practice, in particular when we face highly asymmetric decision problems. The reason is that the shape of the entropy may depend on the loss function. As a consequence, we cannot hope to achieve optimal downstream performance by simply training using \eg log-loss and then later using these predictions in an asymmetric downstream task.

\subsection{Evaluation via IP Calibration}
\begin{figure}
\centering
 \includegraphics[width=1.0\linewidth]{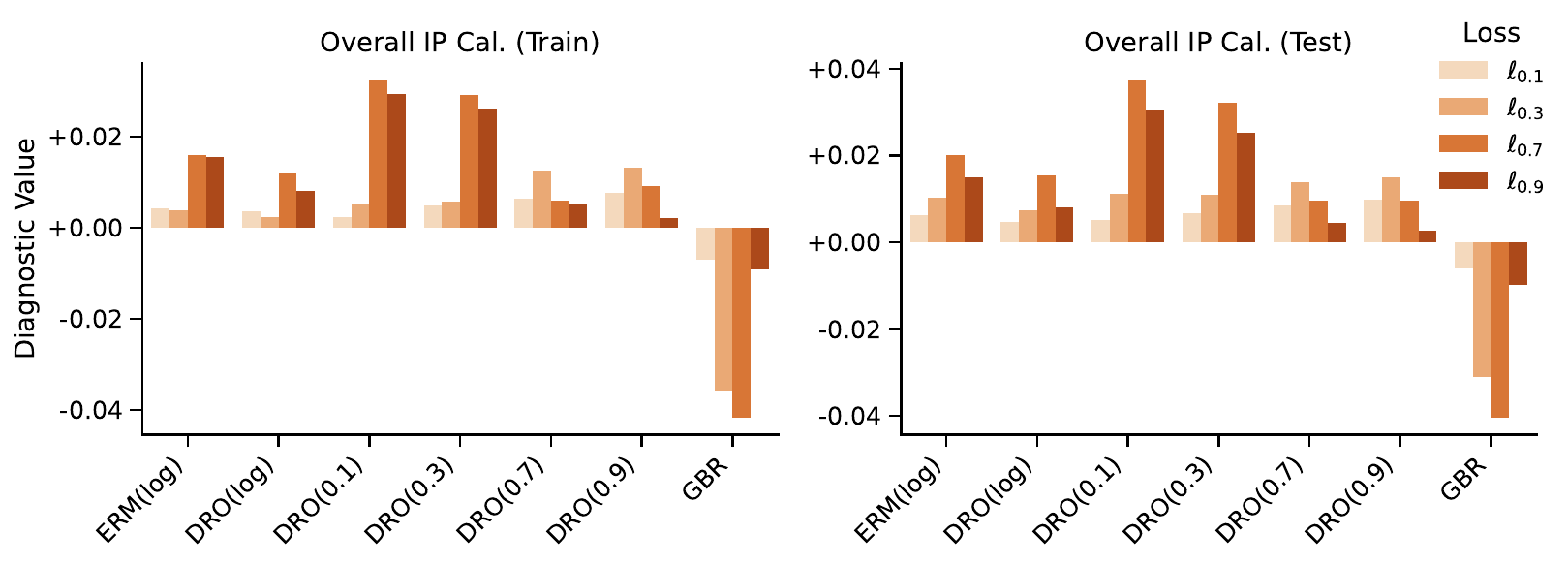}
  \includegraphics[width=1.0\linewidth]{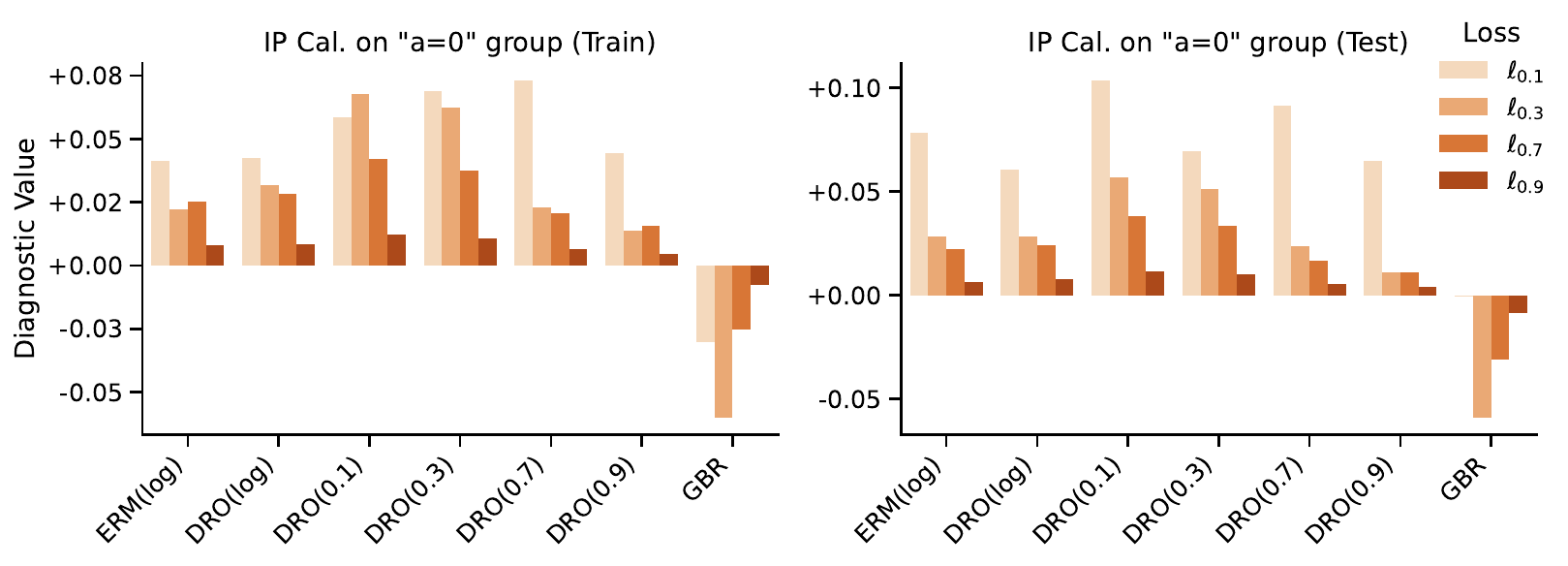}
   \includegraphics[width=1.0\linewidth]{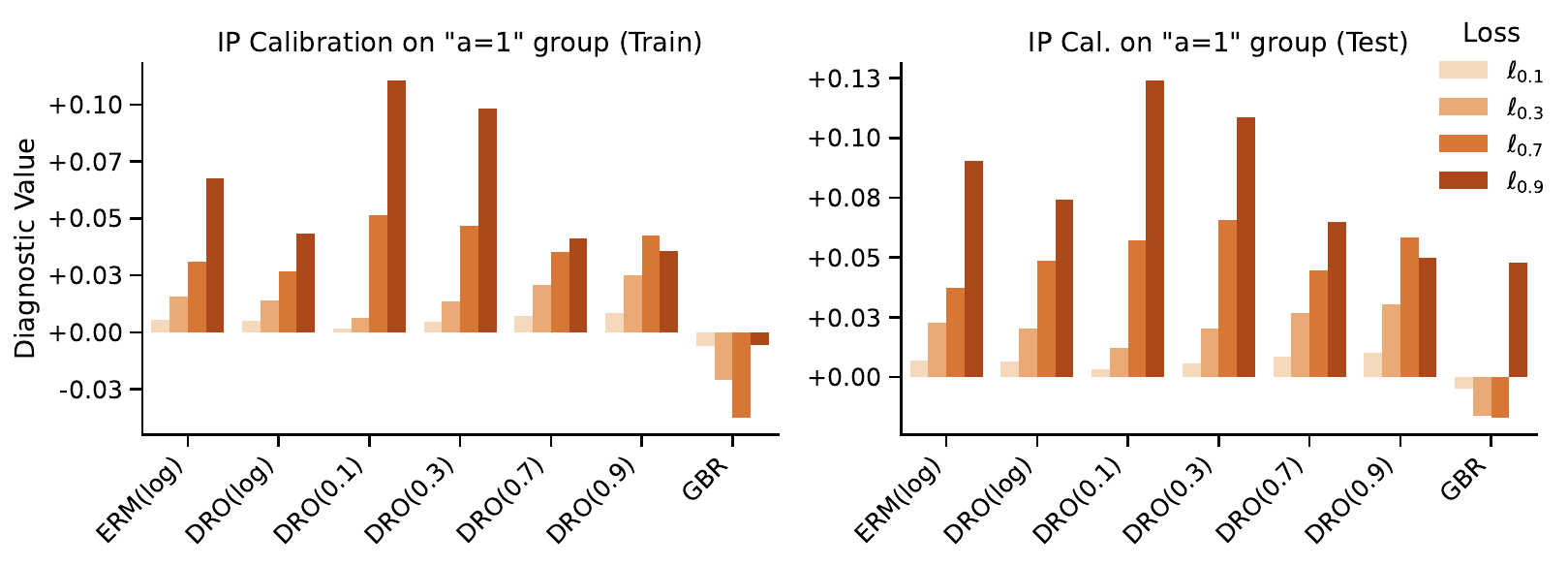}
    \caption{Evaluation of IP calibration under the train (left column) and test (right column) data models on \textsc{acs pums}. In each subplot, the forecasts $\mQ$ vary along the $x$-axis and the loss function used for evaluation corresponds to a hue. The top row shows diagnostic values for IP calibration without groups, that is, the $y$-axis shows $\Rup_{\hat{\baseip}}\left(\Sl(\omega) - \Rup_{\mQ(\omega)}\left( \Sl \right) \right)$ (top left) and $\Rup_{\hat{\baseip}_\text{test}}\left(\Sl(\omega) - \Rup_{\mQ(\omega)}\left( \Sl \right) \right)$ (top right).
    In the other subplots, the $y$-axis shows the diagnostics for IP calibration with respect to the action-induced partition (of Equation~\ref{eq:deccaldiagnostic}) under $\hat{P}$ (left) and $\hat{P}_{\text{test}}$ (right), respectively.
    We observe that the $\textbf{GBR}$ forecast, in contrast to the other forecasts, is sub-calibrated in almost all cases (recall that we consider negative values on the $y$-axis as more desirable than positive values).
    }
    \label{fig:acs_deccal}
\end{figure}

In Section~\ref{sec:ipcalibration} we have discussed IP calibration as a quality of forecasts in the sense of trustworthiness, conceptually separate from achieving optimal IP scores. While the numerical differences in Table~\ref{tab:ipscoresacs} may appear rather small (albeit potentially significant in high-stakes decision making), we obtain comparatively drastic differences when evaluating forecasts in terms of IP calibration. We evaluate IP calibration under both the training as well as the test data model. Recall the definition of $\ell$-calibration without groups (Definition~\ref{def:actuarialfairnesswithoutgroups}):
\[
\Rup_\baseip\left(\Sl(\omega) - \Rup_{\mQ(\omega)}\left( \Sl \right) \right) = 0.
\]
Moreover, since we are interested in the trustworthiness of the action recommendations ($a=0$ or $a=1$), we evaluate via $(\ell,\actionpartition)$-calibration:
\[
\forall a \in \{0,1\}: \RomanNumeralCaps{1}_a \coloneqq \Rup_{\baseip}\left(\chi_{A=a}(\omega) \left(\ell(a,\omega) - \Rup_{\mQ(\omega)}(\ell_a)) \right)  \right) = 0.
\]

Note that if we look at the term $\RomanNumeralCaps{1}_a$ for some $a \in \mathcal{A}$, the magnitude of the term depends on how often $a$ is recommended. But we might also care about trustworthiness with respect to actions that are not recommended often. So we suggest instead plotting as a diagnostic, for each $a \in \{0,1\}$:
\begin{equation}
    \label{eq:deccaldiagnostic}
    \RomanNumeralCaps{2}_a \coloneqq \gbr_{\baseip}\left(\ell(a,\omega) - \Rup_{\mQ(\omega)}(\ell_a))  \mid a_{Q(\omega)}^* = a \right).
\end{equation}
Clearly, $\RomanNumeralCaps{1}_a = 0 \Leftrightarrow \RomanNumeralCaps{2}_a = 0$. A perfectly trustworthy forecast would satisfy the equality; otherwise, recall that in Section~\ref{sec:ipcalibration} we have argued for preferring sub-calibration (``$\leq 0$'') over the converse direction (``$\geq 0$''). This choice simply amounts to preferring pessimism over optimism, which in the context of high-stakes decision making may be warranted.

Since we do not have access to $\baseip$, we evaluate under the train data model $\hat{P}$ as well as the test data model $\hat{P}_{\text{test}}$.
Figure~\ref{fig:acs_deccal} depicts the diagnostic results under both the train and test data model on \textsc{acs pums} for both calibration without groups as well as the diagnostic in \eqref{eq:deccaldiagnostic} for the action-induced partition. For the other datasets, see Appendix~\ref{app:moreresults}.

We evaluate IP calibration under $\{\ell_{0.1},\ell_{0.3},\ell_{0.7},\ell_{0.9}\}$. Some general observations: The \textbf{GBR} is sub-calibrated in almost all cases, whereas the other forecasts can be wildly overoptimistic (see in particular the recommendation ``$a=1$'' on \textsc{celeba} in Appendix~\ref{app:moreresults}). A general, approximate tendency is that the \textbf{DRO(c)} models for small $c$ perform better (in terms of achieving sub-calibration) when evaluated under small $c'$ (referring to evaluation by $\ell_{c'}$) and worst for high $c'$, and conversely for \textbf{DRO(c)} models for large $c$. This observation highlights that rather than attempting to generalize the familiar concept of distribution calibration (see Appendix~\ref{app:relationtoclassicalcalibration}), which is agnostic of the decision problem, it was indeed meaningful to emphasize the primacy of the decision problem in the evaluation of uncertainty.
For the \textbf{ERM(log)} and \textbf{DRO(log)} forecasts, it depends on the concrete problem whether they perform better when evaluated under small or large $c'$.

\section{Related Work}
\label{sec:relatedworkscoringrules}
A highly relevant work to mention is that of \citet{zhao2021right}, who are interested in mechanisms to provide trustworthy individual predictions. Like us, they are motivated by a betting interpretation of probability, citing \citet{definetti1931sul}, and also take an insurantial perspective. 
Although they make use of specific imprecise probabilities in the binary case, parameterized as $[\mu-c,\mu+c] \subseteq [0,1]$, and corresponding marginally desirable gambles (see their Lemma~1) \citep{augustin2014introduction},\footnote{The gambles of the form $b(Y-\mu)-|b|c$ are exactly the marginally desirable gambles for a binary imprecise probability parameterized as $[\mu-c,\mu+c]$; this is effectively what Lemma~1 of \citet{zhao2021right} states.} they do not refer to literature on imprecise probabilities, or in particular the highly related recent developments on game-theoretic imprecise probabilities (see \eg \citep{de2017imprecise,de2021randomness,cooman2022randomness,persiau2022on}). In the latter stream, which has been a major motivation to us, calibration appears in the form of \textit{supermartingales}, built from marginally desirable gambles.
In the case of the cost-sensitive loss, we can establish a close relation between our IP calibration and the framework of \citet{zhao2021right}, see Appendix~\ref{app:zhaoermonlink}.
Another question of interest is: what are the semantics of the imprecision in \citet{zhao2021right}? The paper uses a non-stationary, but precise probabilistic setup, where the probability may vary over time; this appears to fit with the group DRO setting. However, the authors go on to suggest that the semantics of the imprecision refers to the lack of knowledge (from the perspective of the forecaster) about the true, underlying probability. To add to the complexity, their definition of \textit{asymptotic soundness} (essentially asymptotic calibration) fit with yet other frequentist semantics, studied by \citet{walley1982towards} and \citet{frohlich2024strictly}.\footnote{The conceptually puzzling situation can be cleared by consulting \citet{frohlich2024data} on two frequentist manifestations of imprecision; one is about hidden heterogeneity, and fits with the group distributional robustness perspective (where probability varies over time); the second is about instability of an aggregate, and fits with the definition of asymptotic soundness (the $\limsup$ is indicative of this), which the algorithm of \citet{zhao2021right} achieves. However, we note that neither semantics is, in our view, of epistemic character.} 
Our take-away is that we appeal to clarify which semantics one has in mind when working with imprecise probabilities.

Also in relation to this, \citet{zhao2021calibrating} have proposed the notion of decision calibration. We realized that our IP calibration can be seen as a generalization of decision calibration to the imprecise case. 
In our view (and our motivation stems from there), these concepts can be seen as grounded in the notion of marginal desirability and in game-theoretic probability \citep{shafervovk2001probabilityfinance,shafer2019game} (see also \citep[Chapter 6]{augustin2014introduction}). 
In Appendix~\ref{app:relationtoclassicalcalibration}, we discuss the relations between IP calibration, decision calibration of \citet{zhao2021calibrating} and the familiar notion of calibration. 
For the relation between scoring rules and calibration, see also the recent proposal of \textit{U-calibration} \citep{kleinberg2023u}.

In the literature there are various proposals for learning predictors which output imprecise probabilities,\footnote{The final output is not necessarily an imprecise probability. For example, the classifier proposed by \citet{corani2008learning} works with imprecise probabilities yet finally outputs a set-valued classification, \ie a set of labels.} \eg \citep{corani2008learning,caprio2023imprecise,nguyen2023learning,wang2024creinns,hofman2024quantifying}, with epistemic semantics for the imprecision in mind. For example, the imprecise forecast can be obtained by an ensemble (\eg \citep{hofman2024quantifying}), or, in the subjectivist approach, by applying the generalized Bayes rule to an imprecise prior (\eg \citep{corani2008learning}).
Following different routes and motivated by domain generalization issues, the recent papers by \citet{caprio2023imprecise} and \citet{singh2024domain} work with imprecise probabilities in a setting where data comes from a set of distributions.
Yet in these works, the model class is a set of functions from $\mX$ to $\mY$, so the imprecision is not located in the forecasts.\footnote{Of course a function $h : \mX \to \mY$ can be seen as a degenerate imprecise forecast in our sense, outputting a precise probability with all the mass on a single $Y \in \mY$, which in turn can be seen as an imprecise probability.} 
Also, these works employ evaluative criteria (specific generalization bounds and ``C-pareto improvements'', respectively) which are conceptually distinct from ours.
In light of these previous works, our hope is that the framework we have proposed here can aid the unification in the field.

\section{Discussion}
\label{sec:discussion}
In this work, we have proposed a framework for the evaluation of (im)precise forecasts under (im)precise data models based on a generalization of proper scoring rules and calibration.

A key insight, already appearing in the decomposition of  \citet{degroot1983comparison}, was the clear separation between two conceptually distinct qualities of good forecasts: first, recommending actions with favorable consequences (low loss values); second, accurately estimating the consequences (loss) to be expected when following the forecast. These two qualities, related to proper scoring rules and calibration, respectively, appear conflated in the precise case. We do not believe that their divergence in the imprecise case is pathological, but rather nicely reveals their distinct conceptual character.
Even further, we observed that it is possible that two distinct forecasts are both optimal in the sense of the first goal, but only one of them is calibrated, meaning only this forecast is appropriately aware of ``how well it is doing''.

In the opposite direction, we have investigated a phenomenon which can be disentangled under a precise data model, but not anymore under an imprecise data model. In the precise case, \textit{any} proper scoring rule can be used to learn the ``true'' probabilities in the limit of infinite data. These precise forecasts can then be used for optimal decision making under \textit{any} loss function (not necessarily proper) in the sense of minimizing expected loss. Whether data comes from a single, stable distribution in the real world so that the limit exists is another question \citep{frohlich2024strictly} which we put aside here, but at least in theory this shows that \textit{belief} and \textit{decision} (action) can be disentangled in the precise case. In contrast, we have seen in Proposition~\ref{prop:optimalforecast} that this breaks down in the imprecise case \textit{even in theory} (putting aside the problem of finite data, which can only make matters worse).
Again, we do not view this as pathological about the imprecise case, but rather take it as a fundamental lesson about the entanglement of belief and decision.
To give one example, this observation is crucial when group DRO is applied in a machine learning context with fairness desiderata. Here, it is not generally possible to choose a ``neutral'' proper scoring rule at training time which is later optimal for downstream decision making in the sense of minimizing worst-case (over the groups) expected loss. This contrasts with proposals in the literature that urge to cleanly separate belief from decision \citep{beigang2022advantages,kuppler2022fair}.

To sum it up: we have observed that two dualities (the qualities of forecasts and the belief-decision duality) behave surprisingly different in the imprecise case.
In both cases, although in opposite directions, we have argued that the more general imprecise case actually better reflects underlying conceptual subtleties.

We believe that many opportunities for future research arise from the framework presented here. Some of them being very practical, for example how to estimate the generalized Bayes rule appropriately, or to address training/test generalization gaps. Another interesting, potentially challenging avenue would be to go beyond the MinMax assumptions. We highlight that due to the MinMax assumption in~\eqref{eq:minmaxchoice}, imprecision in our framework refers more to pessimism (caution) than to abstaining or indecision, which is a common interpretation in the literature, and which is accompanied by other criteria for action selection \citep{troffaes2007decision}. 
Hence some advocates of imprecise probabilities might be dissatisfied with our proposal. We are tentatively skeptical whether a similar mathematical development could be conducted with other decision criteria; perhaps the Hurwicz criterion (\eg \citep{konek2019ip}) provides a starting point.

A fundamental difference to much of the literature is that here we have not focused on epistemic semantics of imprecision in combination with a precise data model, as in ``the data is \iid from $P$ but we are ignorant about $P$''. A proper scoring rule, according to our definition, would then suggest $P(\cdot|X)$ as the optimal forecast, a consequence of our viewpoint that propriety refers to the recovery of the parameterization of an underlying data model. We conjecture that this is essentially why previous attempts to develop proper scoring rules for imprecise probabilities with epistemic semantics have not reached their goal and instead yielded impossibility theorems. It would be premature to conclude that our framework offers nothing to those interested in epistemic semantics for imprecision, however, but the problem would be approached in a different way. For example, under a precise $P$ data model, the optimal forecast $P(\cdot|X)$ is precise and it is calibrated for any loss. If we do not have access to $P$ but to some estimate $\hat{P}$ (\eg the empirical distribution), we can construct an ambiguity set $\baseip(\hat{P})$ around it,\footnote{For example, by using a law invariant coherent risk measure. See \eg \citep{duchi2018learning} for such risk measures based on $f$-divergences.} as in the DRO and coherent risk measures literature,  and then consider two kinds of statements: first, it would be of interest to obtain bounds that relate, under $P$, the maximum entropy forecast with respect to $\baseip(\hat{P})$ or the generalized Bayes rule forecast $\baseip(\hat{P})_{|X}$ --- note that these are random quantities under $P$ --- to the optimal but unknown $P(\cdot|X)$; second, statements which control the size of the ambiguity set so that sub-calibration of the forecast $\baseip(\hat{P})_X$ holds with high probability \textit{under $P$}. Formally, with $\tilde{S}_\ell(\omega) = S_\ell(\baseip(\hat{P})_{|X(\omega)},\omega)$, 
\[
P\left(\E_P\left[\tilde{S}_\ell\right] - \E_P\left[\gbr_{\baseip(\hat{P})}(\tilde{S}_\ell|X)\right] \geq \alpha \right) \leq \beta,
\]
for some $\alpha,\beta>0$, and similar for the corresponding maximum entropy forecast.
In this way, we can obtain imprecise forecasts which satisfy desired ``safety'' guarantees under precise data models.

We conclude with an appeal to further investigate which benefits imprecise forecasts can bring to the practice of machine learning and statistics.

\section{Acknowledgements}
This work was funded by the Deutsche Forschungsgemeinschaft (DFG, German Research Foundation)
under Germany’s Excellence Strategy — EXC number 2064/1 — Project number 390727645. The authors
thank the International Max Planck Research School for Intelligent Systems (IMPRS-IS) for supporting
Christian Fröhlich. Many thanks to Rabanus Derr for the insightful and helpful discussions.

\bibliography{ipscoringrules.bib}
\bibliographystyle{tmlr}

\clearpage

\appendix
\section{Appendix}

\subsection{A Remark on Tie-Breaking}
\label{app:tiebreaking}
Throughout this paper, for picking an action we often face expressions of the form
\begin{equation} \label{eq:tiebreaking}
    a_{\baseip}^* \picka \argmin_{a \in \mathcal{A}} \max_{P \in \cobar(\baseip)} \E_P[\ell(a,\omega)].
\end{equation}
We resolve the potentially set-valued $\argmin$ to a single action $a_{\baseip}^* \in \mathcal{A}$ in a consistent way (for example by installing a total order on $\mathcal{A}$ and then using this order for tie-breaking); we denote this using the ``$\picka$'' operator. 
While the exact nature of the tie-breaking does not matter for our purposes, it must be so that we can view~\eqref{eq:tiebreaking} as a well-defined function from $\setofips$ to $\mathcal{A}$.

When the action $a_{\baseip}^*$ is fixed, we can extract a corresponding $P^*$ in the $\argmax$, which also may not be unique, so that the value of the MinMax objective is 
\[
\min_{a \in \mathcal{A}} \max_{P \in \cobar(\baseip)} \E_P[\ell(a,\omega)] = \E_{P^*}[\ell(a_{\baseip}^*,\omega)].
\]
The potential non-uniqueness of $P^*$ will not matter in this paper, although it could be resolved by installing a total order on $\Delta^k$ (\eg by restricting the lexicographic order on $\mR^k$).

\subsection{Proof of Proposition~\ref{prop:weakpropriety}}
\label{app:prop:weakpropriety}
\begin{proof}
    We want to show that
    \begin{align}
        \Rup_{\baseip}(S_\ell(\baseip,\omega)) &\leq \Rup_{\baseip}(S_\ell(\mQ,\omega))\\
        \Leftrightarrow \max_{P \in \cobar(\baseip)} \mathbb{E}_P[S_\ell(\baseip,\omega)] &\leq  \max_{P \in \cobar(\baseip)} \mathbb{E}_P[S_\ell(\mathcal{Q},\omega)] \quad \forall \mathcal{Q} \in \setofips.
    \end{align}
    Note that when taking the closed convex hull, we can use $\max$ instead of $\sup$.
    To this end, consider
    \[
    S_\ell(\omega,\baseip) = \ell(a_{\baseip}^*,\omega), \quad  a_{\baseip}^* \picka \argmin_{a \in \mathcal{A}} \max_{P \in \cobar(\baseip)} \E_P[\ell(a,\omega)],
    \]
    where we pick the $a_{\baseip}^*$ from the $\argmin$ in accordance with Remark~\ref{app:tiebreaking}.
    But then, by definition of $\argmin$ we have that
    \begin{equation}
    \label{eq:actionargminscore}
    \max_{P\in \cobar(P)} \E_P[\ell(a_{\mP^*}^*,\omega)] \leq \max_{P\in \cobar(P)} \E_P[\ell(a',\omega)] \;\; \forall a'\in \mathcal{A},
    \end{equation}
    since otherwise $a'$ and some $P \in \cobar(P)$ would achieve a strictly lower MinMax objective value. Hence it holds in particular for $a' = a_{\mQ}^*$, and thus $\Rup_{\baseip}(\ell(a_\baseip^*,\omega)) \leq \Rup_{\baseip}(\ell(a_\mQ^*,\omega))$. 
    
    We remark that \eqref{eq:actionargminscore} makes it clear that it does not matter which action $a^*$ is picked from the potentially set-valued $\argmin$ when the forecast is $\baseip$ itself, as the resulting $\max_{P\in \cobar(P)} \E_P[\ell(a^*,\omega)]$ is guaranteed to be the same. By contrast, consider $\Rup_{\baseip}(S_\ell(\mQ,\omega)) = \Rup_{\baseip}(S_\ell(a_{\mQ}^*,\omega))$, where
    \[
    a_{\mQ}^* \picka \argmin_{a \in \mathcal{A}} \max_{Q \in \cobar(\mQ)} \E_P[\ell(a,\omega)].
    \]
    Here, it is clear that we require a consistent way of picking the action from the $\argmin$, as there is no guarantee that the resulting $\Rup_{\baseip}(\ell(a_\mQ^*,\omega))$ is the same for all choices.
\end{proof}

\subsection{Proof of Proposition~\ref{prop:failureofstrict}}
\label{app:prop:failureofstrict}
\begin{proof}

We begin by showing~\ref{failurestrict2}: $\forall \ell \in \Lfuncs : \exists \baseip \in \setofips : \exists P \in \cobar(\baseip)$ so that $\Rup_{\baseip}(S_\ell(\baseip,\omega)) = \Rup_{\baseip}(S_\ell(P,\omega))$.
To this end, we make use of a Lemma.
\begin{lemma}
    Define $H : \Delta^k \to \mR$ by $H(P) \coloneqq \min_{a \in \mathcal{A}} \E_P[\ell(a,\omega)]$. Pick any $Q_0,Q_1$ in the interior of $\Delta^k$. Then $H$ is differentiable on all points of the segment $[Q_0,Q_1]$ except perhaps on a finite set of points.
\end{lemma}
\begin{proof}
    It is well-known that $H(P)$ is concave as a minimum over a set of linear functions, and any concave function is differentiable except perhaps on a countable set of points. We now exploit our assumption that the action space is finite, $|\mathcal{A}|<\infty$, to show that indeed the set must be finite. We restrict ourselves to looking at the segment $[Q_0,Q_1]$.

    For any $a_i \in \mathcal{A}$, 
    define $f_i : \mR^k \to \mR$ given by $f_{i}(P) \coloneqq \E_P[\ell(a_i,\omega)]$, so that $H(P) = \min_{a \in \mathcal{A}} f_i(P)$. Note that $f_i$ is linear, and it is differentiable. We acknowledge that when $P \in \mR^k$ but $P \notin \Delta^k$, we should prefer an inner product notation over $\mathbb{E}$, but for simplicity we stick to using $\mathbb{E}$. Thus $H$ is given as the minimum over a finite set of differentiable functions. 
The function $H$ can only fail to be differentiable at a point $P$ if $f_i(P)=f_j(P)$ for some pair $(i,j)$ (note that this is not \textit{sufficient} for a failure of differentiability, for example when $f_i(P)=f_j(P) \forall P \in \Delta^k$, meaning $a_i$ and $a_j$ always have the same consequences). Thus, to show that $H$ is differentiable except perhaps at a finite number of points, it suffices to bound the number of problematic points where $f_i(P)=f_j(P)$ for each pair $(i,j)$, noting that there are only finitely many such pairs.

    Consider the line $y(\alpha) \coloneqq \alpha Q_0 + (1-\alpha)Q_1$, $y : \mR \to \mR^k$. For $\alpha \in [0,1]$, we get the segment $[Q_0,Q_1]$.
Due to linearity of $f_i$, $f_i(y(\alpha)) = \alpha f_i(Q_0) + (1-\alpha) f_i(Q_1)$. The function $f_i \circ y : \mR \to \mR$ is a line. Therefore $f_i \circ y$ and $f_j \circ y$ either fully coincide, do not intersect at all, or intersect in a single point. But now since $f_i(P)=f_j(P) \Leftrightarrow f_i(y(\alpha^*))=f_j(y(\alpha^*))$, if $P=\alpha^* Q_0 + (1-\alpha^*)Q_1$ for some $\alpha^* \in \mR$, we either have $f_i(P)=f_j(P) \; \forall P \in [Q_0,Q_1]$, or $f_i(P)=f_j(P)$ for at most a single $P \in \mR^k$. If $f_i$ and $f_j$ fully coincide, then $P \mapsto \min\{f_i(P),f_j(P)\}$ is differentiable, and thus this pair $(i,j)$ poses no challenge to differentiability; also note that if $\ell(a_i,\omega)=\ell(a_j,\omega)$, we could have removed one of them from $\mathcal{A}$ in the setup of our scoring rule without any significant consequence.
\end{proof}

We now exploit Theorem 6.2 from \citet{grunwald2004game}. We slightly restate it using our notation and specialize to our needs.\footnote{We remark that the original statement of Theorem 6.2 by \citet{grunwald2004game} works with \textit{any} potentially randomized Bayes actions (``acts'') $\zeta^*$ (``Bayes act against $P^*$''). In our setup, we do not include randomized actions, and there always exists a non-randomized Bayes act $a_P^* \picka \argmin_{a \in \mathcal{A}} \E_{P^*}[\ell(a,\omega)]$. In our statement of~\ref{item:grunwald622}, we have inlined the definition of a ``saddle point'' in this setting; see Theorem 4.3 of \citet{grunwald2004game}.}
\begin{proposition}[{\citet[Theorem 6.2, adapted]{grunwald2004game}}]
    Let $\baseip \subseteq \Delta^k$ be convex and $P^*$ so that $H(P^*)=\max_{P \in \cobar(\baseip)} H(P)$. Suppose that $\forall P \in \baseip$ there exists $P_0 \in \Delta^k$ so that, with $Q_\lambda \coloneqq (1-\lambda) P_0 + \lambda P$ it holds:
    \begin{enumerate}[label=(\theproposition.\arabic*)]
        \item $P^* = Q_{\lambda^*}$ for some $\lambda^* \in (0,1)$,
        \item \label{item:grunwald622} and $\lambda \mapsto H(Q_\lambda)$ is differentiable at $\lambda=\lambda^*$.
        Then $a_{P^*}^* \picka \argmin_{a \in \mathcal{A}} \E_{P^*}[\ell(a,\omega)]$ satisfies
        \[
        \min_{a \in \mathcal{A}} \sup_{P \in \baseip} \E_P[\ell(a,\omega)] =  \sup_{P \in \baseip} \min_{a \in \mathcal{A}} \E_P[\ell(a,\omega)] = \E_{P^*}[\ell(a_{P^*}^*,\omega)] =  \sup_{P \in \baseip} \E_P[\ell(a_{P^*}^*,\omega)].
         \]
    \end{enumerate}
\end{proposition}
If the conditions of the theorem are satisfied, then $\Rup_{\baseip}(S_\ell(\baseip,\omega)) = \Rup_{\baseip}(S_\ell(P^*,\omega))$, because $\min_{a \in \mathcal{A}} \sup_{P \in \baseip} \E_P[\ell(a,\omega)]  = \Rup_{\baseip}(S_\ell(\baseip,\omega))$. This would mean that we obtain~\ref{failurestrict2}.

We now show that the conditions of the statement are satisfied.
We take any segment $[Q_0,Q_1]$ in the interior of $\Delta^k$, and then take some $\mathcal{P}$ so that $\mathcal{P} \subsetneq \operatorname{int}([Q_0,Q_1])$, and so that $H$ is differentiable on all of $\mathcal{P}$. In particular, $H$ is then continuous on $\mathcal{P}$ (a concave function is continuous on the interior of its domain), and thus attains its maximum on $\mathcal{P}$ at some $P^*$. The previous Lemma assures us that such $\mathcal{P}$ exists, since we can have non-differentiability at only finitely many points of $[Q_0,Q_1]$.  The choice that $\mathcal{P} \subsetneq \operatorname{int}([Q_0,Q_1])$ guarantees that for every $P \in \mathcal{P}$, there exists some $P_0 \in [Q_0,Q_1]$ so that with $Q_\lambda = (1-\lambda)P_0 + \lambda P$ we get that the desired $P^*=Q_{\lambda^*}$ for some $\lambda^* \in (0,1)$. 

It remains to show that then $\lambda \mapsto H(Q_\lambda)$ is differentiable at $\lambda=\lambda^*$. 
This mapping can be seen as a composition $H(g(\lambda))$, where $g : (0,1) \to \Delta^k$, $g(\lambda) = (1-\lambda)P_0 + \lambda P$, hence $H \circ g : (0,1) \to \mR$. Differentiability follows from the multivariate chain rule, considering that $H$ is differentiable on $[Q_0,Q_1]$ and $g$ is differentiable on $(0,1)$. Therefore Theorem 6.2 of \citet{grunwald2004game} applies.

Finally, we show by example that~\ref{failurestrict3}: $\exists \ell : \exists \baseip$ so that $\forall P \in \Delta^k : \Rup_{\baseip}(S_\ell(\baseip,\omega)) < \Rup_{\baseip}(S_\ell(P,\omega))$.

Let $\Omega=\{\omega_1,\omega_2\}$, $\mathcal{A}=\{a_1,a_2,a_3\}$ and $\baseip=\{P_1,P_2\}$. For the tie-breaking (see Appendix~\ref{app:tiebreaking}) we fix the ordering $a_1 > a_2 > a_3$. Let $P_1(\{\omega_1\})=1$, $P_2(\{\omega_2\})=1$. Define the loss function by $\ell(a_1,\omega_1)=2$, $\ell(a_1,\omega_2)=7$; $\ell(a_2,\omega_1)=6$, $\ell(a_2,\omega_2)=3$; and $\ell(a_3,\omega)=0.5\ell(a_1,\omega) + 0.5\ell(a_2,\omega)$, $\omega \in \Omega$. Hence $\E_{P_1}[\ell(a_3,\omega)]=4$ and $\E_{P_2}[\ell(a_3,\omega)]=5$. Observe that $a_3$ is the unique optimal action under $\baseip$:
\[
\{a_3\} = \argmin_{a \in \mathcal{A}} \max_{P \in \{P_1,p_2\}} \E_P[\ell(a,\omega)] < \min_{a \in \{a_1,a_2\}} \max_{P \in \{P_1,p_2\}} \E_P[\ell(a,\omega)].
\]
Now assume that there exists some $P^* \in \Delta^k$ which \textit{uniquely} recommends $a_3$:
\[
\{a_3\} = \argmin_{a \in \mathcal{A}}  \E_{P^*}[\ell(a,\omega)],
\]
where \textit{uniquely} means that the $\argmin$ is the singleton $\{a_3\}$. If the $\argmin$ was not a singleton, then the action recommendation would depend on the contingent tie-breaking procedure, which in this case would favor $a_1$ and $a_2$ over $a_3$, leading to a sub-optimal recommendation.

Assume $P^*(\{\omega_1\})=p$ and $P^*(\{\omega_2\})=1-p$. Then
\begin{align}
    \E_{P^*}[\ell(a_1,\omega)] &= p \cdot 2 + (1-p) \cdot 7,\\
    \E_{P^*}[\ell(a_2,\omega)] &= p \cdot 6 + (1-p) \cdot 3,\\
    \E_{P^*}[\ell(a_3,\omega)] &= p \cdot 4 + (1-p) \cdot 5.
\end{align}
If we want $P^*$ to uniquely recommend $a_3$, we need
\[
p \cdot 4 + (1-p) \cdot 5 < \min\left(p \cdot 2 + (1-p) \cdot 7, p \cdot 6 + (1-p) \cdot 3\right),
\]
but this has no solution. Observe that for $P^*=0.5 P_1 + 0.5 P_2$, which uniquely maximizes $H$ in this case, all actions get the same expected score, but the balance cannot be tilted towards $a_3$.
This situation relates to an insight of \citet{grunwald2004game}: if a $P^*$ maximizing $H$ on $\cobar(\baseip)$ exists but it has no unique Bayes action, then it is unclear which of these Bayes actions (where we would have to allow for randomized actions, as well) is ``robust Bayes'', which means solving the $\min_{a \in \mathcal{A}} \max_{P \in \cobar(\baseip)} \E_P[\ell(a,\omega)]$ objective. In the above example, $a_3$ can be essentially interpreted as a randomized action, where a coin flip decides between $a_1$ and $a_2$. Note that in this example, 
\begin{align}
H(P^*) = \E_{P^*}[S_\ell(P^*,\omega)] &= \max_{P \in \cobar{\baseip}} \min_{a \in \mathcal{A}} \E_P[\ell(a,\omega)]=4.5 \\
&< \Rup_{\baseip}(S_\ell(\baseip,\omega)) = \Rup_{\baseip}(\ell(a_3,\omega)) = 5\\
&< \Rup_{\baseip}(S_\ell(P^*,\omega)) =  \Rup_{\baseip}(\ell(a_1,\omega)) = 7.
\end{align}
meaning the maximum entropy is strictly smaller than the IP score of the optimal constant forecast, which is in turn smaller than the IP score of the forecast corresponding to the maximum entropy probability.

Finally, we note that if $|\Omega|>2$, the above proof can be straightforwardly adapted, where $P_1$ and $P_2$ only have the support on two elementary events. The logic regarding $P^*$ also works since a rescaling of $P^*$ does not affect the order of the relevant expected losses.

\end{proof}

\subsection{Proof of Proposition~\ref{prop:strongpropriety}}
\label{app:prop:strongpropriety}
To obtain the statement, we prove the following proposition.

\begin{proposition}
\label{prop:differentactionrecommendations}
    Let $\mathcal{P},\mathcal{Q} \in \setofips$  and $\mathcal{P} \notsamehull \mathcal{Q}$. Then there exist $\ell \in \Lfuncs$ such that 
    \[
    \left(a_{\mP}^* \picka \argmin_{a \in \mathcal{A}} \sup_{P \in \mathcal{P}} \E_P[\ell(a,\omega)] \right) \neq  \left(a_{\mathcal{Q}}^* \picka \argmin_{a \in \mathcal{A}} \sup_{Q \in \mathcal{Q}} \E_Q[\ell(a,\omega)] \right),
     \]
    and
    \[
    \exists \omega \in \Omega: S(\mathcal{P},\omega) \neq S(\mathcal{Q},\omega),
    \]
    and indeed that 
    \begin{align}
    \Rup_{\baseip}(S_\ell(\mP,\omega)) &< \Rup_{\baseip}(S_\ell(\mathcal{Q},\omega))\\
    \Leftrightarrow \Rup_{\mP}(\ell_{a_{\mP}^*}) &<  \Rup_{\mP}(\ell_{a_{\mathcal{Q}}^*}). 
    \end{align}
\end{proposition}
\begin{proof}
Since $\mathcal{P} \notsamehull \mathcal{Q}$, we can find some $X : \Omega \rightarrow \mR$ such that either
    $\Rup_{\mathcal{P}}(X) < b < \Rup_{\mathcal{Q}}(X)$ or $\Rup_{\mathcal{Q}}(X) < b < \Rup_{\mathcal{P}}(X)$\footnote{
    This holds since there is a one-to-one correspondence of closed convex sets of probabilities and upper expectations \citep[Section 3.6.1]{walley1991statistical}. We remark that \citet{walley1991statistical} does not work with probabilities in $\Delta^k$, but with so-called ``linear previsions''. For the equivalence of these viewpoints when $|\Omega|<\infty$, see \citep{frohlich2024strictly}.
    } --- but we cannot choose which way (consider $\mP=\Delta^k$, this never gives a smaller risk assessment than any other $\mathcal{Q}$).
    Then construct the loss function as $\ell(a_1) \coloneqq X(\omega)$, and $\ell(a_2)=b$, and $\ell(a)=c>\max(X(\omega),b)$ otherwise.
    
    First, assume that the direction is such that  $\Rup_{\mathcal{P}}(X) < b < \Rup_{\mathcal{Q}}(X)$. Clearly, the credal set $\mP$ will recommend $a_1$ whereas $\mathcal{Q}$ will recommend $a_2$. That is, they recommend different actions $a_{\mP}^* \neq a_{\mQ}^*$.
    Now we also show that:
   \[\Rup_{\baseip}(S_\ell(\mP,\omega)) < \Rup_{\baseip}(S_\ell(\mathcal{Q},\omega))
   \]
   We know that $\mP$ recommends $a_1$ hence the left term is $\Rup_{\baseip}(\ell(a_1,\omega))=\Rup_{\mP}(X)$, whereas $\mathcal{Q}$ recommends $a_2$, meaning the right term is $\Rup_{\baseip}(\ell(a_2,\omega))=b$. Since $\Rup_{\mP}(X)<b$ by assumption, this proves the statement for this direction. 
   
   Now assume conversely that $\Rup_{\mathcal{P}}(X) > b > \Rup_{\mathcal{Q}}(X)$ and use the same $\ell$. Then $\mP$ will recommend $a_2$, whereas $\mathcal{Q}$ will recommend $a_1$. But $\Rup_{\mP}(\ell(a_2,\omega))=b < \Rup_{\mP}(X)$.
\end{proof}

\subsection{Proof of Proposition~\ref{prop:optimalforecast}}
\label{app:prop:optimalforecast}
We first lift the action space to combine actions for the different $X=x$ in a single higher-order action; If $\mathcal{X}=\{x_1,..x_J\}$, we define $\tilde{\mathcal{A}} \coloneqq \{(a_1,..,a_J) : a_j \in \mathcal{A}, j=1..J\}$, and the lifted loss function is $\ell' : \tilde{A} \times \Omega \to \mR$, where
\[
\ell'(\ta,\omega) \coloneqq \ell(a_j,\omega) \text{ if } X(\omega)=x_j.
\]
We then consider
\[
\min_{\ta \in \tA} \max_{P \in \cobar(\baseip)} \E_P[\ell'(\ta,\omega)] \leq \min_{\mQ : \mathcal{X} \to \setofips} \max_{P \in \cobar{P}} \E_P[S_\ell(\mQ(X),\omega)] = \ipscore(\ell,\baseip).
\]
That the inequality holds is clear since for any $S_\ell(\mQ(X),\omega)$, we have the correspond lifted action $\ta_\mQ \coloneqq (a_{\mQ(x_1)}^*,..,a_{\mQ(x_J)}^*)$ on the left side, and $S_\ell(\mQ(X),\omega) = \ell'(\ta_\mQ,\omega)$.
Thus we aim to characterize the optimal $\ta^* = (a_1^*,..,a_J^*)$ of the MinMax objective on the left side first. If we can then further exhibit some (in this case precise) forecast $f^*: \mathcal{X} \to \Delta^k$, so that we have $a_j^* \picka \argmin_{a \in \mathcal{A}} \E_{f^*(x_j)}[\ell(a,\omega)] \; \forall x_j \in \mathcal{X}$, then $f^*$ is optimal on the right side.

We now focus on $\min_{\ta \in \tA} \max_{P \in \cobar(\baseip)} \E_P[\ell'(\ta,\omega)]$, and apply to it the maximum entropy principle of \citet{grunwald2004game}.
Adapted to our setup, Theorem 5.2 of \citet{grunwald2004game} provides the following, here stated informally.
\begin{proposition}[{adapted from Theorem 5.2 of \citet{grunwald2004game}}]
    Compute some maximum entropy probability
    \[
    P^* \in \argmax_{P \in \cobar(\baseip)} \min_{\ta \in \mathcal{A}} \E_{P}[\ell'(\ta,\omega)],
    \]
    noting that any $P^*$ in the $\argmax$ suffices. Then $P^*$ admits a potentially randomized Bayes action $\zeta^*$, which has the desired property of being a ``saddle point''.
\end{proposition}
From Proposition 3.1.iii) of \citet{grunwald2004game}, we know that if a probability $P$ has a randomized Bayes act, then it also has some non-randomized Bayes action. In combination, we therefore know that if $P^*$ has a \textit{unique} Bayes action, meaning $\argmin_{\ta \in \tA} \E_{P^*}[\ell'(\ta,\omega)]$ is a singleton, then this must be a non-randomized one, so it fits our setup. Thus we obtain, by inlining the definition of saddle point:
\begin{proposition}
    Compute some maximum entropy probability 
    \[
    P^* \in \argmax_{P \in \cobar(\baseip)} \min_{\ta \in \mathcal{A}} \E_{P}[\ell'(\ta,\omega)].
    \]
    If 
    \[
    \{\ta_{P^*}^*\} \coloneqq \argmin_{\ta \in \tA} \E_{P^*}[\ell'(\ta,\omega)],
    \]
    \ie the action recommendation on the lifted space is unique, then
    \[
    \min_{\ta \in \tA} \max_{P \in \cobar(\baseip)} \E_P[\ell'(\ta,\omega)]  =  \max_{P \in \cobar(\baseip)} \min_{\ta \in \tA} \E_P[\ell'(\ta,\omega)] = \E_{P^*}[\ell'(\ta_{P^*}^*,\omega)] = \max_{P \in \cobar(\baseip)} \E_P[\ell'(\ta_{P^*}^*,\omega)]
    \]
\end{proposition}
We want to show that indeed the forecast $P^*$, if this uniqueness condition holds, will recommend actions which when viewed as a single lifted action coincide exactly this lifted action $\ta_{P^*}^*$. %
\begin{lemma}
Assume $P^*(X=x_j)>0 \; \forall x_j \in \mathcal{X}$. 
    Let $\RomanNumeralCaps{1} \coloneqq \argmin_{\ta \in \tA} \E_{P^*}[\ell'(\ta,\omega)]$
and consider $\RomanNumeralCaps{2}_j \coloneqq \argmin_{a \in \mathcal{A}} \E_{P^*(\cdot|X=x_j)}[\ell(a,\omega)]$. 

Then $\RomanNumeralCaps{1}$ is a singleton (unique) if and only if $\RomanNumeralCaps{2}_j$ is a singleton (unique) for each $X=x_j$, $j=1..J$.
In the case of uniqueness, we get that $\RomanNumeralCaps{1} = (\RomanNumeralCaps{2}_1,..,\RomanNumeralCaps{2}_J)$.
\end{lemma}
\begin{proof}
    Consider
    \begin{align}
        &\min_{\ta \in \tA} \E_{P^*}[\ell'(a,\omega)]& \\ 
        = &\min_{\ta=(a_1,..,a_J) \in \tA} \sum_{j=1}^J \E_{P^*(\cdot|X=x_j)}[\ell(a_j,\omega)] P^*(X=x_j)
    \end{align}
    The minimization with respect to the $j$-th component of $\ta$ depends only on the $j$-th summand. Note that if we had $P^*(X=x_j)=0$, then the $j$-th component would remain fully undetermined. 
    Otherwise, it is clear that with $\ta^*=(\ta_1^*,..,\ta_J^*)$, the set of actions which we can obtain in the $j$-th component of an optimal solution $\ta^*$ coincides exactly with the potentially set-valued $\argmin_{a \in \mathcal{A}} \E_{P^*(\cdot|X=x_j)}[\ell(a,\omega)]$, since the scaling factor $P^*(X=x_j)$ is irrelevant for the solution, as long as it is strictly positive. Thus uniqueness is preserved in both directions.
\end{proof}
The desired statement now follows from the Lemma: the uniqueness of $a_{P^*}^*$ guarantes the uniqueness of the $\RomanNumeralCaps{2}_j \coloneqq \argmin_{a \in \mathcal{A}} \E_{P^*(\cdot|X=x_j)}[\ell(a,\omega)]$, and moreover that $a_{P^*}^*=(\RomanNumeralCaps{2}_1,..,\RomanNumeralCaps{2}_J)$. Therefore $\ipscore(\ell,\baseip) = \Rup_{\baseip}(S_\ell(f^*(X),\omega))$ upon defining $f^*(x_j) \coloneqq P^*(\cdot|X=x_j)$, $x_j \in \mathcal{X}$.

Finally, to obtain the form of Proposition~\ref{prop:optimalforecast}, it remains to show:
\begin{lemma}
We can rewrite the generalized conditional entropy as follows:
    \[
H_{\ell}(P|X) \coloneqq \E_{X,\omega \sim P}[S_\ell(P(\cdot|X),\omega)] = \min_{\ta \in \mathcal{A}} \E_{P}[\ell'(\ta,\omega)].
\]
\end{lemma}
\begin{proof}
Assume without loss of generality that $P(X=x_j)>0 \; \forall x_j \in \mX$ (convince yourself that this does not matter).
We begin by rewriting the left side, giving
\begin{align}
    \E_{X,\omega \sim P}[S_\ell(P(\cdot|X),\omega)] &= \sum_{j=1}^J \E_{\omega \sim P(\cdot|X=x_j)}\left[S_\ell(P(\cdot|X=x_j),\omega)\right]P(X=x_j) \\
    &= \sum_{j=1}^J \E_{\omega \sim P(\cdot|X=x_j)}\left[\ell(a_{P(\cdot|X=x_j)}^*,\omega)\right]P(X=x_j).
\end{align}
Now the right side gives
\begin{align}
    &\min_{\ta \in \mathcal{A}} \E_{P}[\ell'(\ta,\omega)]\\
    = &\min_{\ta=(a_1,..,a_J) \in \tA} \sum_{j=1}^J \E_{P(\cdot|X=x_j)}\left[\ell(a_j,\omega)\right] P(X=x_j).
\end{align}
Write $\ta^*=(\ta_1^*,..,a_j^*,..,\ta_J^*)$ for any of the solutions (without uniqueness guarantee). Observe that
\[
\E_{\omega \sim P(\cdot|X=x_j)}\left[\ell(a_j^*,\omega)\right] = \E_{\omega \sim P(\cdot|X=x_j)}\left[\ell(a_{P(\cdot|X=x_j)}^*,\omega)\right],
\]
although possibly $a_j^* \neq a_{P(\cdot|X=x_j)}^*$. We can thus conclude that $\E_{X,\omega \sim P}[S_\ell(P(\cdot|X),\omega)] = \min_{\ta \in \mathcal{A}} \E_{P}[\ell'(\ta,\omega)]$.
\end{proof}

\subsection{An Extended Example}
\label{app:exampleofmaxentfailure}
We first begin with an unconditional example without features, and then expand on it to include features.  Let $\Omega \coloneqq \{\text{rain=0},\text{rain=1}\}$ and $\mathcal{A} \coloneqq \{\text{umbrella=0},\text{umbrella=1}\}$.
Consider the cost-sensitive loss function $\ell$ with parameter $c=0.1$, given by the table:

    \begin{tabular}{ccc}
         &  rain=0 & rain=1 \\
         umbrella=0 & 0 & 0.9 \\
         umbrella=1 &  0.1 & 0\\
    \end{tabular}

To simply notation, we implicitly identify a probability measure $P$ for which $P(\text{rain=1})=p$ with the point $p \in [0,1]$. We abbreviate ``umbrella=0'' as ``u=0''. 
Consider $\mathcal{P} \coloneqq [0.85,0.95]$. Then the IP scores of the actions can be computed as:
\begin{align}
    &\Rup_{\baseip}(\ell(\text{u=0},\omega)) = \max_{P \in [0.85,0.95]} P \cdot 0.9 = 0.95 \cdot 0.9 = 0.855\\
    &\Rup_{\baseip}(\ell(\text{u=1},\omega)) = \max_{P \in [0.85,0.95]} (1-P) \cdot 0.1 = (1-0.85) \cdot 0.1 = 0.015.
\end{align}
So without question, the optimal action is to bring an umbrella. Note that a precise forecast $P$ uniquely recommends bringing an umbrella if $P>0.1$, and for $P=0.1$ both actions $u=0$ and $u=1$ have the same expected loss. Hence, in this case all $P \in \mathcal{P}$ agree on the action recommendation.

Without features, the conditional entropy of Definition~\ref{def:condentropy} (take a trivial, constant $X$) simply becomes the unconditional
\[
H_\ell(P) \coloneqq \min_{a \in \mathcal{A}} \E_P[\ell(a,\omega)].
\]
Then $P^*=0.1$, $H(P^*)=0.09$, is indeed the maximum entropy probability over $[0,1]$, as the following graphic shows:

\includegraphics[width=0.6\textwidth]{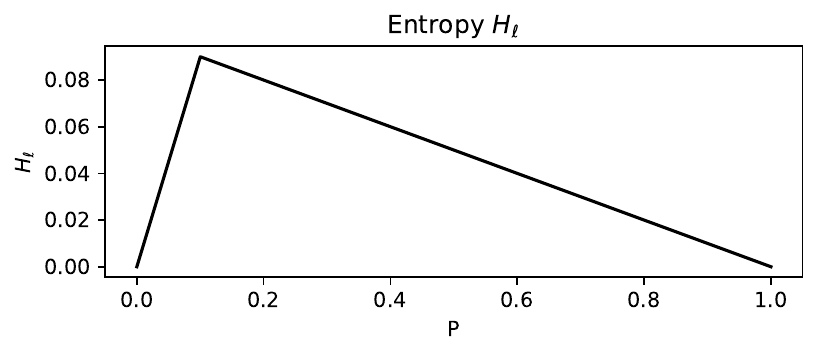}

And for this $P^*$, the condition of Proposition~\ref{prop:optimalforecast} is not satisfied, since the action recommendation is not unique.
To observe this failure, we can therefore consider an interval $\mathcal{P}' \coloneqq [0.05,0.15]$, which includes $P^*$. The IP scores of the actions are:
\begin{align}
    &\Rup_{\baseip'}(\ell(\text{u=0},\omega)) = \max_{P \in [0.05,0.15]} P \cdot 0.9 = 0.15 \cdot 0.9 = 0.135\\
    &\Rup_{\baseip'}(\ell(\text{u=1},\omega)) = \max_{P \in [0.05,0.15]} (1-P) \cdot 0.1 = (1-0.05) \cdot 0.1 = 0.095.
\end{align}
Thus, it is again uniquely optimal to bring an umbrella under $\baseip'$, but now there is disagreement among the precise probabilities in $\baseip'$, and the maximum entropy principle cannot be applied since $P^*$ does not give a unique recommendation. Clearly, any precise forecast $P>0.1$ is optimal in this example, as well as any imprecise forecast $[P,Q]$ for which $P>0.1$ (we always assume for an interval $[P,Q]$ that $P \leq Q$).

We now lift this example to include features $\mathcal{X} \coloneqq \{\text{cloudy},\text{sunny}\}$. Then $\Omega = \mathcal{X} \times\mathcal{Y}$, $\mathcal{Y} \coloneqq \{\text{rain=0},\text{rain=1}\}$.
Assume that for some country A, the weather can be described by
\begin{align}
    &P^1(X=\text{cloudy})=0.4,\\
    &P^1(Y=\text{rain=1}|X=\text{cloudy})=0.95,\\
    &P^1(Y=\text{rain=1}|X=\text{sunny})=0.05.
\end{align}
In contrast, for country B:
\begin{align}
    &P^2(X=\text{cloudy})=0.9,\\
    &P^2(Y=\text{rain=1}|X=\text{cloudy})=0.85,\\
    &P^2(Y=\text{rain=1}|X=\text{sunny})=0.15.
\end{align}
Let us consider the lifted action space 
\[
\tA \coloneqq \{(u=0,u=0), (u=0,u=1), (u=1,u=0), (u=1,u=1)\},
\]
as in the proof of Proposition~\ref{prop:optimalforecast}. Correspondingly, the lifted loss function is 
\[
\ell'(\ta=(a_1,a_2),(x,y)) \coloneqq \ell(a_1,y) \text{ if } x=\text{cloudy}; \; \ell(a_2,y) \text{ otherwise}.
\]

The GBR forecast (see Section~\ref{sec:gbrforecast}) is $[0.85,0.95]$ for $X=\text{cloudy}$, and $[0.05,0.15]$ for $X=\text{sunny}$. Therefore, in light of what we have seen before, it recommends the lifted action $(u=1,u=1)$.

We now compute the IP scores of the lifted actions under $\tilde{\baseip} = \{P^1,P^2\}$ (to compute upper expectations it suffices to consider the extreme points):
\begin{align}
    \Rup_{\tilde{\baseip}}(\ell'((u=0,u=0),\omega)) = &\max_{P \in \{P^1,P^2\}} P(X=\text{cloudy}) P(Y=\text{rain=1}|X=\text{cloudy}) \cdot 0.9 \\ &+ (1-P(X=\text{cloudy})) P(Y=\text{rain=1}|X=\text{sunny}) \cdot 0.9 \\
    &= 0.702,
\end{align}
\begin{align}
    \Rup_{\tilde{\baseip}}(\ell'((u=0,u=1),\omega)) = &\max_{P \in \{P^1,P^2\}} P(X=\text{cloudy}) P(Y=\text{rain=1}|X=\text{cloudy}) \cdot 0.9 \\ &+ (1-P(X=\text{cloudy})) P(Y=\text{rain=0}|X=\text{sunny}) \cdot 0.1 \\
    &\approx 0.697,
\end{align}
\begin{align}
    \Rup_{\tilde{\baseip}}(\ell'((u=1,u=0),\omega)) = &\max_{P \in \{P^1,P^2\}} P(X=\text{cloudy}) P(Y=\text{rain=0}|X=\text{cloudy}) \cdot 0.1 \\ &+ (1-P(X=\text{cloudy})) P(Y=\text{rain=1}|X=\text{sunny}) \cdot 0.9 \\
    &= \mathbf{0.029},
\end{align}
\begin{align}
    \Rup_{\tilde{\baseip}}(\ell'((u=1,u=1),\omega)) = &\max_{P \in \{P^1,P^2\}} P(X=\text{cloudy}) P(Y=\text{rain=0}|X=\text{cloudy}) \cdot 0.1 \\ &+ (1-P(X=\text{cloudy})) P(Y=\text{rain=0}|X=\text{sunny}) \cdot 0.1 \\
    &= 0.059.
\end{align}
Thus, we see that the lifted action $(u=1,u=0)$ is uniquely optimal, that is, bringing an umbrella on cloudy days but not on sunny days. Recall that country A recommends an umbrella only on cloudy days, hence the optimal forecast with respect to $\tilde{\baseip}$ coincides with the optimal forecast for $P^1$, whereas $P^2$ recommends an umbrella for both cloudy and sunny days. It can be checked that indeed $P^1$ is the unique maximum entropy probability in this example, where we now have to consider the conditional entropy of Definition~\ref{def:condentropy}, \ie the full Bayes risk. Here, the marginals on $X$ are crucial, since they affect the full Bayes risk. Since $P^1$ uniquely recommends $(u=1,u=0)$, the maximum entropy principle of Proposition~\ref{prop:optimalforecast} applies (at least approximately).

To observe that $P^1$ is the unique maximum entropy probability, first note that 
\[
H_\ell(P) = \min((1-0.1)P,0.1(1-P)),
\]
and then compute
\[
\dagger \coloneqq \E_{P^\lambda}\left[H_\ell(P^\lambda(\cdot|X))\right], \quad P^\lambda \coloneqq \lambda P^1 + (1-\lambda) P^2,
\]
which gives
\begin{align}
\dagger &= (0.6-0.2\cdot \lambda)\cdot\min\left(0.9\cdot y, 0.1\cdot (1-y)\right) + (1-(0.6-0.2\cdot \lambda))\cdot \min\left(0.9\cdot z,0.1\cdot (1-z)\right), \\
 y &\coloneqq (\lambda\cdot 0.95\cdot 0.4+(1-\lambda)\cdot 0.85\cdot 0.9)/(\lambda\cdot 0.4+(1-\lambda)\cdot 0.6), \\
 z &\coloneqq (\lambda\cdot 0.05\cdot 0.6 + (1-\lambda)\cdot 0.15\cdot 0.1)/(1 - (\lambda\cdot 0.4+(1-\lambda)\cdot 0.6)).
\end{align}
This term has a unique maximum at $\lambda=1$ (a computer algebra system comes to the rescue).

\textit{Failure of sub-calibration}.
First, we illustrate that it is possibly that two distinct ($\notsamehull$) forecasts are optimal, but only of them is sub-calibrated.
Consider again $\mathcal{P}'=[0.05,0.15]$. An optimal forecast is $Q=0.15$ with action recommendation $u=1$, but it is not sub-calibrated without groups (in this case, the action-induced partition is the trivial one, since the action recommendation is constant).
Sub-calibration would mean that:
\begin{align}
    &\Rup_{\baseip'}\left(\ell(u=1,\omega) - \E_{Q=0.15}[\ell(u=1,\omega)]\right) \leq 0 \\
    \Leftrightarrow \forall P' \in \baseip': &\E_{P'}[\ell(u=1,\omega)] \leq \E_{Q=0.15}[\ell(u=1,\omega)]),
\end{align}
but for $P'=0.05$, we have $\E_{P'=0.05}[\ell(u=1,\omega)]=0.095 > \E_{Q=0.15}[\ell(u=1,\omega)])=0.085$, and therefore sub-calibration does not hold.
In contrast, the forecast $\baseip'$ is also optimal (it also recommends $u=1$) and is trivially (sub-)calibrated; but the crucial difference is in the \textit{risk assessment} of the consequences.

We can also show that in the extended example above, which includes features ``cloudy'' and ``sunny'', where the condition of Proposition~\ref{prop:optimalforecast} applies, the optimal forecast $f^*$ of Proposition~\ref{prop:optimalforecast} is neither sub-calibrated without groups nor $\mathcal{B}_{f^*;\mathcal{A}}$-sub-calibrated. We know that $P^1$ which recommends $(u=1,u=0)$ is the unique maximizer of the generalized conditional entropy and thus corresponds to the optimal forecast in Proposition~\ref{prop:optimalforecast}.
To check sub-calibration without groups, we need to see if
\begin{align}
    T \coloneqq \Rup_{\baseip}\left(\ell'((u=1,u=0),\omega) - \E_{P^1(\cdot|X=\cdot)}[\omega' \mapsto \ell'((u=1,u=0),\omega')]\right) \leq 0.
\end{align}
Write $T = \max(T_1,T_2)$, noting that we only need to look at the extreme points of $\mathcal{P}$, with
\begin{align}
    T_i \coloneqq \underbrace{\E_{P^i}\Bigl[\ell'((u=1,u=0),\omega)\Bigr]}_{\text{actual expected loss of $P^1$ forecast under $P^i$}} - \underbrace{\E_{X \sim P^i}}_{\text{sampling the forecast}}\Bigl[
    \underbrace{\E_{\omega \sim P^1(\cdot|X)}[\ell'((u=1,u=0),\omega)]}_{\text{entropy of the $P^1$ forecast for fixed $X$}}
    \Bigr]
\end{align}
which is equal to
\begin{align}
    T_i \coloneqq \Bigl(&P^i(X=\text{cloudy}) P^i(Y=\text{rain=0}|X=\text{cloudy}) \cdot 0.1 \\
    + &P^i(X=\text{sunny}) P^i(Y=\text{rain=1}|X=\text{sunny}) \cdot 0.9)\Bigr) \\
    - \Bigl(&P^i(X=\text{cloudy}) P^1(Y=\text{rain=0}|X=\text{cloudy}) \cdot 0.1 \\
    + &P^i(X=\text{sunny}) P^1(Y=\text{rain=1}|X=\text{sunny}) \cdot 0.9
    \Bigr)
\end{align}
By plugging in the respective numbers we obtain that $T=\max(0,0.011)>0$, however.
Hence the $P^1$ forecast is not sub-calibrated without groups, and therefore also cannot be (sub-)calibrated on any finer partition, in particular the action-induced partition.

\subsection{Proof of Proposition~\ref{prop:partitionsubcal}}
\label{app:prop:partitionsubcal}
Recall that a partition $\mathcal{B}_1$ is coarser than $\mathcal{B}_2$ if every set in $\mathcal{B}_2$ is a subset of some set in $\mathcal{B}_1$. This can usefully be reformulated as follows.
\begin{lemma}[known]
    If $\mB_1$ is coarser than $\mB_2$, then every set in $\mB_1$ can be written as a disjoint union of some sets from $\mB_2$.
\end{lemma}
\begin{proof}
    Fix $B_1 \in \mB_1$ which we want to write as a disjoint union of sets from $\mB_2$. Consider all those $B_{2i} \in \mB_2$ for which $B_{2i} \subseteq B_1$. We claim that the union of those disjoint sets is exactly $B_1$.
    First, it is clear that the union cannot be larger than $B_1$. To show that the union is exactly $B_1$ we must show that every $b \in B_1$ is in some $B_{2i}$. Since $\mB_2$ is a partition, $b \in \tilde{B}_2$ for some $\tilde{B}_2 \in \mB_2$. From the refinement property, we know that there exists some $\tilde{B}_1 \in \mB_1$ for which $\tilde{B}_2 \subseteq \tilde{B}_1$, so $b \in \tilde{B}_1$. But $\mB_1$ is a partition, so either $\tilde{B}_1 = B_1$ (then we are done, since $\tilde{B}_2$ is one of the $B_{2i}$), or $\tilde{B}_1 \cap B_1 = \emptyset$, but this contradicts the assumption that $b \in B_1$.
\end{proof}
We can now prove the Proposition.

\begin{proof}
    Consider some $B_1 \in \mB_1$. Write $B_1 = \bigcup_{i=1}^J B_{2_i}$ for the disjoint union of sets from $\mB_2$. Then $\chi_{B_1} = \sum_{i=1}^J\chi_{B_{2_i}}$. Since $\Rup_{\baseip}$ is subadditive, we get
    \begin{align}
    &\Rup_{\baseip}\left(\chi_{B_1}(\omega) \left(\tilde{S}(\omega) - \Rup_{\mQ(\omega)}(\tilde{S})\right) \right) \\
    = &\Rup_{\baseip}\left(\sum_{i=1}^J \chi_{B_{2i}}(\omega) \left(\tilde{S}(\omega) - \Rup_{\mQ(\omega)}(\tilde{S})\right) \right) \\
    &\leq \sum_{i=1}^J \Rup_{\baseip}\left(\chi_{B_{2i}}(\omega) \left(\tilde{S}(\omega) - \Rup_{\mQ(\omega)}(\tilde{S})\right)\right) \leq 0,
    \end{align}
    by the assumption of $(\ell,\mB_2)$-sub-calibration, and therefore we have shown $(\ell,\mB_1)$-sub-calibration.
\end{proof}

\subsection{Proof of Proposition~\ref{prop:gbrcalandentropybound}}
\label{app:prop:gbrcalandentropybound}
\begin{proof}
First we clarify notation. When writing expressions of the form
\[
\Rup_{\baseip}\left(f(\omega) - \Rup_{\mQ(\omega)}(\omega' \mapsto f(\omega'))\right) ,
\]
it can be necessary to distinguish in the notation the $\omega$ corresponding to the gamble on which we here evaluate the outer $\Rup_{\baseip}$, and an additional inner variable. For each $\omega \in \Omega$, the term $\Rup_{\mQ(\omega)}(\omega' \mapsto f(\omega'))$ here is simply a constant. When the distinction is not necessary, we fall back to using $\omega$ as the variable.

    To understand that the GBR is $\ellB$-calibrated, we rewrite the definition of calibration using Remark~\ref{remark:rewritingcalgbr} and plug in the definition of the GBR forecast:
    \[
    \forall B \in \mathcal{B}: \gbr_{\baseip}\Bigl(S_\ell(\mP_{|B},\omega) - \gbr_\baseip\left(\omega' \mapsto S_\ell(\mP_{|B},\omega') | B\Bigr) |B\right) = 0.
    \]
    We highlight again to the reader the subtle distinction of $\omega$ \versus $\omega'$ here. Note that for each $B \in \mB$, the quantity $\gbr_\baseip\left(\omega' \mapsto S_\ell(\mP_{|B},\omega') | B\right)$ is a constant, and can therefore be pulled outside\footnote{We have $\Rup(X-c)=0 \Leftrightarrow \Rup(X)=c$ when $c$ is a constant.} 
    to obtain
    \[
    \forall B \in \mathcal{B}: \gbr_{\baseip}(S_\ell(\mP_{|B},\omega) | B) = \gbr_\baseip\left(S_\ell(\mP_{|B},\omega) | B\right),
    \]
    which is true. For the second statement, let $\mQ_\mB$ be another $\mB$-measurable and $\ellB$-calibrated forecast, so
\[
    \forall B \in \mathcal{B}: \gbr_{\baseip}\Bigl(S_\ell(\mQ_B,\omega) - \Rup_{\mQ_B}\left(\omega' \mapsto S_\ell(\mQ_B,\omega') \right) |B \Bigr) = 0.
    \]
    using Remark~\ref{remark:rewritingcalgbr}. Since the forecast is $\mB$-measurable, we can pull the constant outside to obtain
    \[
    \forall B \in \mathcal{B}: \gbr_{\baseip}(S_\ell(\mQ_B,\omega) |B) = \Rup_{\mQ_B}\left(S_\ell(\mQ_B,\omega) \right).
    \]
    We know from weak propriety (Proposition~\ref{prop:weakpropriety}) that
    \[
     \gbr_{\baseip}(S_\ell(\mP_{|B},\omega) | B) \leq  \gbr_{\baseip}(S_\ell(\mQ_{B},\omega) | B),
    \]
    and therefore $\gbr_{\baseip}(S_\ell(\mP_{|B},\omega) | B) \leq \Rup_{\mQ_B}\left(S_\ell(\mQ_B,\omega) \right)$. Equivalently, $\overline{H}_\ell(\baseip_{|B}) \leq \overline{H}_\ell(\mQ_B)$. 
    
    Finally, we show by example that it is possible to find $(\ell,\baseip,\mB,\mQ_\mB)$ satisfying the respective assumptions so that $\overline{H}_\ell(\baseip_{|B}) < \overline{H}_\ell(\mQ_B)$. Take the trivial partition $\mB=\{\Omega\}$ and $\baseip=\{p\}$. Consider the constant forecast $p$, which equals the GBR in this case, and a constant (\ie $\mB$-measurable) competitor forecast $q$ which we assume to be also $\ellB$-calibrated. Now construct $p,q$ and $\ell$ in such a way that $\E_p[\ell(a_p^*,\omega)] < \E_p[\ell(a_q^*,\omega)] = \E_q[\ell(a_q^*,\omega)] < \E_q[\ell(a_p^*,\omega)]$, which is clearly possible when $|\mathcal{A}|\geq 2$. That is, $q$ will recommend a sub-optimal action $a_q^*$ (under the $p$ data model), but will agree with $p$ on the risk assessment for this action. Then both $p$ and $q$ will be $\ellB$-calibrated under a $p$ data model; for $p$ this is obvious; since $\E_p[\ell(a_q^*,\omega)] = \E_q[\ell(a_q^*,\omega)]$, the $q$ forecast is also $\ellB$-calibrated. To sum up, in this case both $p$ and $q$ are $\mB$-calibrated, but $\overline{H}_\ell(p) < \overline{H}_\ell(q)$.
    \end{proof}

\subsection{Proof of Proposition~\ref{prop:scoreentropybound}}
\label{app:prop:scoreentropybound}
\begin{proof}
    The total IP score of $\mQ_\mB$ can be decomposed as
    \begin{align}
        \Rup_{\baseip}(S_\ell(\mQ_\mB(X),\omega)) = &\Rup_{\baseip}\left(\sum_{B} \chi_B(\omega) S_\ell(\mQ_B,\omega) \right) \\
        \leq & \sum_{B}\Rup_{\baseip}\left(\chi_B(\omega) S_\ell(\mQ_B,\omega)\right)\\
        = &\sum_{B} \sup_{P \in \baseip} \E_{P(\cdot|B)}[S_\ell(\mQ_B,\omega)]P(B) \\
        \leq &\sum_{B} \gbr_{\baseip}(S_\ell(\mQ_B,\omega)|B) \cdot \Rup_{\baseip}(\chi_B)\\
        = &\sum_{B} \overline{H}_\ell(\mQ_B) \cdot \Rup_{\baseip}(\chi_B).
    \end{align}
\end{proof}

\subsection{Proof of Proposition~\ref{prop:gbruniqueness}}
\label{app:prop:gbruniqueness}
\textit{Proof of~\ref{item:gbruniqueness1}}. We want to show: If $\mQ_B \subseteq \baseip_{|B} \forall B$ and $\mQ_B \subsetneq \baseip_{|B}$ for at least one $B$, then $\mQ_\mB$ cannot be $\mB$-sub-calibrated for all loss functions, therefore in particular not $\mB$-calibrated for all loss functions.
Recall that $\ellB$-sub-calibration requires that
\[
\forall B \in \mB: \gbr_\baseip(S_\ell(\mQ_B,\omega)) \leq \Rup_{\mQ_B}(S_\ell(\mQ_B,\omega))
\]
(see the proof of Proposition~\ref{prop:gbrcalandentropybound}).
To show failure of $\ellB$-sub-calibration, we therefore exhibit some $B \in \mB$ and $\ell \in \Lfuncs$ so that
\[
\gbr_\baseip(S_\ell(\mQ_B,\omega)) > \Rup_{\mQ_B}(S_\ell(\mQ_B,\omega)).
\]
If we can find some $Z \in \linfty$ so that $\gbr_{\baseip}(Z|B) > \Rup_{\mQ_B}(Z)$, then when defining the trivial $\ell(a,\omega) \coloneqq Z(\omega)$, ignoring the action entirely, we are done (then $S_\ell(\mQ_B,\omega)=Z(\omega)$); the optimal action is determined only by tie-breaking (Appendix~\ref{app:tiebreaking}). 

In general, if $\mQ,\mQ' \in \setofips$ are closed convex and $\mQ \subsetneq \mQ'$, then we have the monotonicity
\[
\forall Z \in \linfty: \Rup_{\mQ}(Z) \leq \Rup_{\mQ'}(Z),
\]
and the inequality has to be strict for at least one $Z^* \in \linfty$, otherwise $\Rup_{\mQ}$ and $\Rup_{\mQ'}$ would always coincide, but there is a one-to-one relation between closed convex sets of probabilities and upper expectations (see the footnote in Appendix~\ref{app:prop:strongpropriety}).
But for this $Z^*$, we obtain in the above that $\gbr_\baseip(S_\ell(\mQ_B,\omega)) > \Rup_{\mQ_B}(S_\ell(\mQ_B,\omega))$, and therefore $\mQ_B$ is does not satisfy the $\ellB$-sub-calibration condition on $B \in \mB$.

The proofs of~\ref{item:gbruniqueness2} and~\ref{item:gbruniqueness3} are analogous.

\textit{Proof of~\ref{item:gbruniqueness2}}. We want to show: If $\baseip_{|B} \subseteq \mQ_B \forall B$, then $\mQ_\mB$ is $\ellB$-sub-calibrated.
Upper expectations are monotone in the sense that if $\baseip_{|B} \subseteq \mQ_B$, then $\Rup_{\baseip_{|B}}(Z) \leq \Rup_{\mQ_B}(Z) \; \forall Z \in \linfty$; recall $\gbr(Z|B) = \Rup_{\baseip_{|B}}(Z)$. Now take $Z(\omega) \coloneqq S(\mQ_B,\omega)$. For $\mQ_\mB$ to be $\ellB$-sub-calibrated, we need
\begin{align}
    &\forall B \in \mathcal{B}: \gbr_{\baseip}\Bigl(S_\ell(\mQ_{B},\omega) - \Rup_{\mQ_B}(\omega' \mapsto S_\ell(\mQ_B,\omega')) | B \Bigr) \leq 0\\
    \Leftrightarrow
    &\forall B \in \mathcal{B}: \gbr_{\baseip}(S_\ell(\mQ_{B},\omega) | B)
    \leq \Rup_{\mQ_B}(S_\ell(\mQ_B,\omega)).
\end{align}
which holds due to the monotonicity.

\textit{Proof of~\ref{item:gbruniqueness3}}. We want to show: if $\exists B\in \mB: \baseip_{|B} \neq \mQ_B$, then
        $\mQ$ cannot be $\mB$-calibrated for all loss functions.
 Since $\baseip_{|B} \notsamehull \mQ_B$ (they are closed convex by assumption, then ``$\neq$'' implies ``$\notsamehull$''), we can find some $Z \in \linfty$ so that
\[
\Rup_{\baseip_{|B}}(Z) < \Rup_{\mQ_B}(Z)  \text{ or } \Rup_{\baseip_{|B}}(Z) > \Rup_{\mQ_B}(Z),
\]
(see the proof of Proposition~\ref{prop:strongpropriety} in Appendix~\ref{app:prop:strongpropriety}).
Now set $\ell(a,\omega)=Z(\omega)$. 
Then \[
\Rup_{\baseip_{|B}}(S_\ell(\mQ_B,\omega)) \neq \Rup_{\mQ_B}(S_\ell(\mQ_B,\omega))
\]
and therefore calibration fails.

\subsection{Relation to Classical Calibration}
\label{app:relationtoclassicalcalibration}
While we were differently motivated by the centrality of the decision-theoretic entropy concept, the notion of \textit{marginally desirable gambles} in the IP literature \citep{walley1991statistical,augustin2014introduction} as well as an insurantial perspective, it turns out that our notion of IP calibration (approximately) specializes, in the precise case, to \textit{decision calibration} of \cite{zhao2021calibrating}.
To illustrate the connection, we specialize both the data model as well as the forecast to be precise --- note that it may also be of interest to consider IP sub-calibration of imprecise forecasts under a precise data model to model epistemic uncertainty.

\begin{remark}[compare Definition 2 of \citet{zhao2021calibrating}]
Let $\ell \in \Lfuncs$. When $\baseip=\{P\}$ and the forecast $Q : \mathcal{X} \to \Delta^k$ is precise in Definition~\ref{def:actuarialfairnesswithoutgroups} (calibration without groups), we obtain, with $\tilde{S}(\omega) = S_\ell(Q(\omega),\omega)$:
    \begin{equation}
        \label{eq:decisioncal}
        \E_{P}\left[\tilde{S}(\omega) - \E_{Q}[\tilde{S}]\right] = 0.
    \end{equation}
\end{remark}
A slight difference in the setup is that for \citet{zhao2021calibrating}, a forecast returns a probability on the set of labels, whereas in our setup a forecast returns a probability on the whole $\Omega$, allowing the forecast $f(X=x)$ even to disbelief the $X=x$ information. Also, in their setup a loss function has the type signature $\ell: \mathcal{Y} \times \mathcal{A} \to \mR$, and can therefore not directly depend on the elementary event.

Another difference is that \citet{zhao2021calibrating} allow that some loss function $\ell$ is used to pick the optimal action (in this case, the Bayes optimal action under $\ell$), but that \textit{another} loss function is used to evaluate the consequences of the action.\footnote{For example, the proof of part 3 of Theorem 1 (available in Appendix C.1, part 3,  in the arxiv version \citep{zhao2021calibratingarxivversion}), uses the log-loss to choose the optimal action (with $\mathcal{A}$ as the probability simplex, the optimal action $a_P^*$ is then the probability $P^*$ itself due to strict propriety of the log-loss), but another loss (in this case any bounded loss) to evaluate the consequences.} 
We are not aware of a reason to choose different loss functions for selecting the action and evaluating its consequences.

Interestingly, \citet{zhao2021calibrating} prove the following (here informally stated) proposition:\footnote{We are unable to comprehend some steps of the proof.} decision calibration holds for all bounded loss functions if and only if distribution calibration holds. When the loss functiond depends on $\omega$ only through $Y$, distribution calibration demands that $\forall q: \E_P[Y|Q(X)=q]=q$; this is the familiar notion of calibration.

\subsection{Detailed Information About Datasets}
\label{app:data}
In this appendix, we provide detailed information about how we obtained the datasets.

\subsubsection{\textsc{acs pums}}
\label{app:data:acs}
Inspired by \citet{ding2021retiring}, we use census data from the ACS PUMS survey (year: 2018). From the raw data (see the github repository for scripts which download and preprocess the data, we keep the following subset of columns:
\begin{quote}
    AGEP, SCHL, MAR, SEX, ESP, MIG, GCL, HICOV, SCIENGP, DIS, DEAR, DEYE, DREM, CIT, NATIVITY, ANC, RAC1P, LANX
\end{quote}
For the meaning of these features, we refer to the official ACS PUMS documentation, available at \url{https://www2.census.gov/programs-surveys/acs/tech_docs/pums/data_dict/PUMS_Data_Dictionary_2018.pdf} (Accessed: 2024-10-07). A row in the dataset corresponds not exactly to an individual person, but rather to a \textit{statistical individual}, carrying a statistical weight (see the official ACS PUMS documentation for how the weights are computed; we neglect the weights in our experiments, and treat a row essentially as a person). For example, ``AGEP'' refers to age, ``SCHL'' represents level of education. We add two features to the dataset: a binary feature ``isHisp'', indicating Hispanic origin in a binary way (based on the raw, more fine-grained ``HISP'' feature); and the binary ``isWhiteOnly'' feature, indicating whether a person is ``white and not Hispanic''. 

As the binary target label, we construct the target based on the raw ``ESR'' feature. We consider a person as ``having a job'' ($Y=1$), if they are either ``Civilian employed, at work'', ``Civilian employed, with a job but not at work'', ``Armed forces, at work'' or ``Armed forces, with a job but not at work''; contrariwise, we consider a person as ``not having a job'' ($Y=0$) if they are either ``Unemployed'' (unemployed proper) or ``Not in labor force''. In this way, we follow \citet{ding2021retiring}, who introduced an employment prediction task with the ACS PUMS data. We remark that, in the economic sense, a person who is ``Not in labor force'' is not considered to be unemployed.

As \citet{ding2021retiring}, we filter for $16 < \text{age} < 90$. We form $50$ groups based on dividing the data into the $50$ U.S. states (\eg we do not include data for Puerto Rico and D.C.). The sizes of the group-wise datasets are highly uneven; we do not perform any balancing, since in practice, we may wish to use all data which is available, without throwing data away to achieve balancing.

In addition, we then used \texttt{PolynomialFeatures(degree=2, interaction\_only=True, include\_bias=False)} from the \texttt{sklearn.preprocessing} library to obtain polynomial features, which means introducing interaction terms in the model.

To the reader who wonders about why we have chosen certain features and the task; in the present paper we use all datasets rather in a toy-like fashion, to have realistic data but not in an intentionally meaningful real-world prediction/decision context, but we have previously used the \textsc{acs pums} dataset described here for fairness-related investigations in a different context.

As a final step in our preprocessing pipeline, for each model which we train, we apply a standard scaler which we fit on the training data.

\subsubsection{\textsc{framingham}}
We downloaded the dataset from \url{https://www.kaggle.com/datasets/aasheesh200/framingham-heart-study-dataset} (Accessed: 2024-08-16). The dataset is of tabular character, with $16$ columns (such as gender, age, education, being a smoker, having diabetes \etc), and the target label is the binary ``10-year risk of future coronary heart disease'' ($Y=1$ means ``at risk'' and $Y=0$ means ``not at risk'').

For preprocessing, we imputed missing values following \url{https://www.kaggle.com/code/kapiluniyara/framingham-dataset}. In contrast to them, we did not remove outliers (an exploratory analysis suggested a very limited effect of this).
As in the \textsc{acs pums} dataset, we used \texttt{PolynomialFeatures(degree=2, interaction\_only=True, include\_bias=False)} to obtain polynomial features.

We formed two groups, ``young'' and ``old'', by thresholding the ``age'' feature at $60$ ($\leq 60$ \versus $>60$). For each group, we then use \texttt{RandomOverSampler} of the \texttt{imblearn.over\_sampling} library to balance the frequency of the labels (so approximately $50\%$ have a positive label in each group); otherwise, there are relatively few examples of $Y=0$ in the dataset. By doing this we intend to illustrate that balancing is not sufficient to remove the influence of asymmetric loss functions (in DRO and in ERM). Note that this way of balancing is different to balancing the group sizes (see Appendix~\ref{app:data:acs}).

We use a random subset of relative size $0.2$ as the test set, and the rest as the training data. As a result, the ``young'' group has $5{,}091$ training examples and $2{,}567$ positive training labels, and the old group has $6{,}62$ training examples and $330$ positive training labels.
As to the test data, the ``young'' group has $1{,}237$ examples and $615$ positive labels, and the old group has $166$ examples and $84$ positive labels.

As a final step in our preprocessing pipeline, for each model which we train, we apply a standard scaler which we fit on the training data.

\subsubsection{\textsc{celeba}}
We downloaded the dataset from \url{https://www.kaggle.com/datasets/jessicali9530/celeba-dataset} (Accessed: 2024-09-03), where the train and test split is also defined (we do not use the validation data). 
We use the ``Wearing\_Earrings'' attribute as our binary target label, where $Y=1$ means ``wearing earrings'' and $Y=0$ means ``not wearing earrings''. We divide the data based on the ``male'' attribute in two groups, ``male'' and ``non-male'', where the latter included non-binary identities.

We use a pretrained Resnet50 from the \texttt{pytorch} library (with the default weights) to obtain embeddings of train and test data, whereby ``embedding'' we mean computing a forward pass of the model and saving the output of the average pooling layer (just before the final linear layer), which is simply achieved by replacing the final linear layer with the identity mapping. The embeddings are then of dimension $2{,}048$. We then perform a PCA, which we fit on the combined training data, with $100$ components to reduce the dimensionality.

From the train data, we used only the first $501$ batches of size $64$ for each group, giving $32{,}064$ training examples for each group. We use all the available test data ($7{,}715$ examples for the ``male'' group and $12{,}247$ for the ``non-male'' group).
 The ``male'' group then has $480$ positive train examples and $182$ positive test exampels; the ``non-male'' group has $12{,}247$ positive train examples and $3{,}943$ positive test examples. In this case, we do not balance the label frequency to have this as a contrasting example to \texttt{framingham}.

As a final step in our preprocessing pipeline, for each model which we train, we apply a standard scaler which we fit on the training data.

\subsection{Sigmoid Approximation of the Asymmetric Scoring Rule}
\label{app:sigmoidapprox}
\begin{figure}
    \centering
    \includegraphics[width=0.32\textwidth]{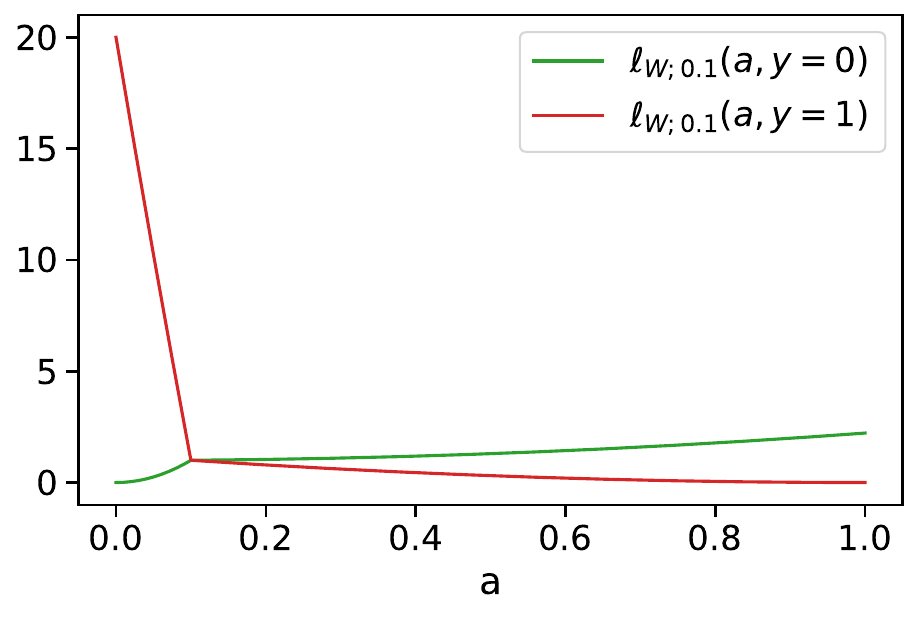}
    \includegraphics[width=0.32\textwidth]{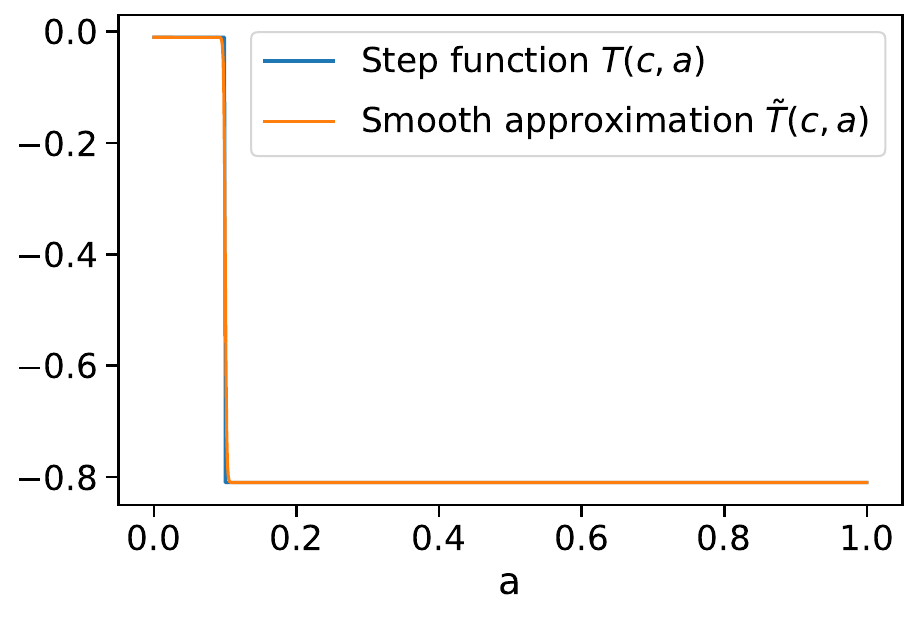}
    \includegraphics[width=0.32\textwidth]{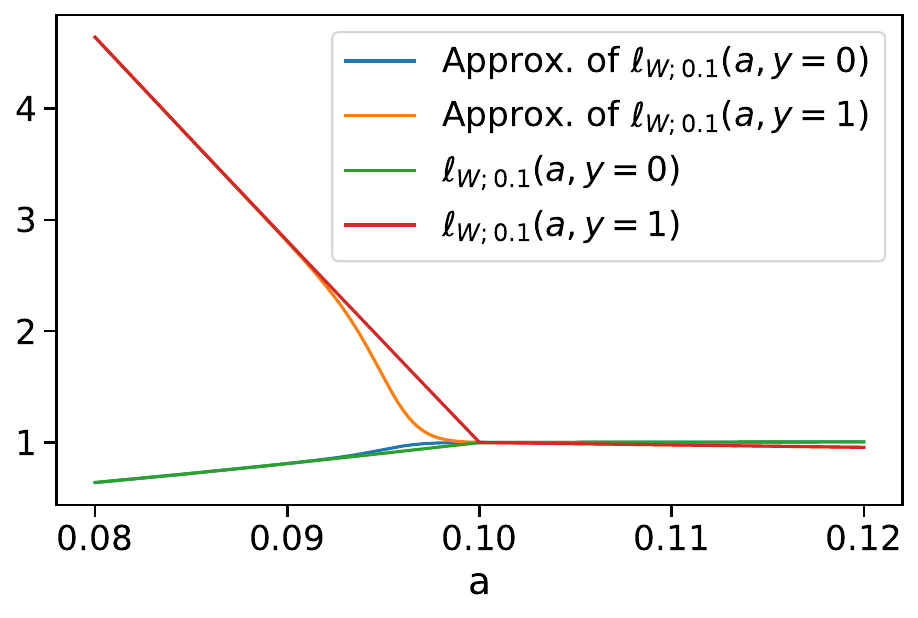}
    \caption{Left: a visualization of the asymmetric loss function (the two curves are the partial loss functions, meaning each curve has a fixed outcome $y \in \{0,1\}$). Middle: the discrete step function and the sigmoid approximation. The difference is hardly discernible. Right: by zooming in near $c=0.1$, the difference between the original loss function and its approximation can be detected. }
    \label{fig:winklerloss}
\end{figure}

When specializing to the Brier (squared) loss, the asymmetric scoring rule proposed by \citet{winkler1994evaluating} is given by:
 \[
    \ellwc(a,Y(\omega)=y) = 1 - \frac{\ell^2(a,y) - \ell^2(c,y)}{T(c,a)}; \quad  T(c,a) \coloneqq \begin{cases}
        -\ell^2(c,1) & a \geq c \\
        -\ell^2(c,0) & a < c
    \end{cases}; \quad \ell^2(a,y) \coloneqq (a-y)^2.
    \]
This loss function is depicted in Figure~\ref{fig:winklerloss} (left) for $c=0.1$.
At the point $c$, this loss is not differentiability due to the discrete step function $a \mapsto T(c,a)$. We therefore use the following variant in practice:
\[
\tilde{T}(c,a) \coloneqq -\alpha \ell^2(c,1) - (1-\alpha) \ell^2(c,0), \quad \alpha \coloneqq \sigma(f \cdot (a-c)),
\]
where $\sigma(z)=1/(1+\exp(-z))$ is the sigmoid function, and we choose $f=1000$ for a close approximation. With such a high value of $f$, the difference between the step function and its approximation is hardly discernible visually (see Figure~\ref{fig:winklerloss} (middle)). By zooming in around $c$, the small difference in the loss function can be discerned (Figure~\ref{fig:winklerloss} (right)). Since the difference between the original loss function and its smooth approximation is tiny and appears negligible in practice, we do not introduce new notation for the approximation (which we use in the experiments).

\subsection{Entropy Comparison}
\begin{figure}[h]
    \centering
    \includegraphics[width=0.4\linewidth]{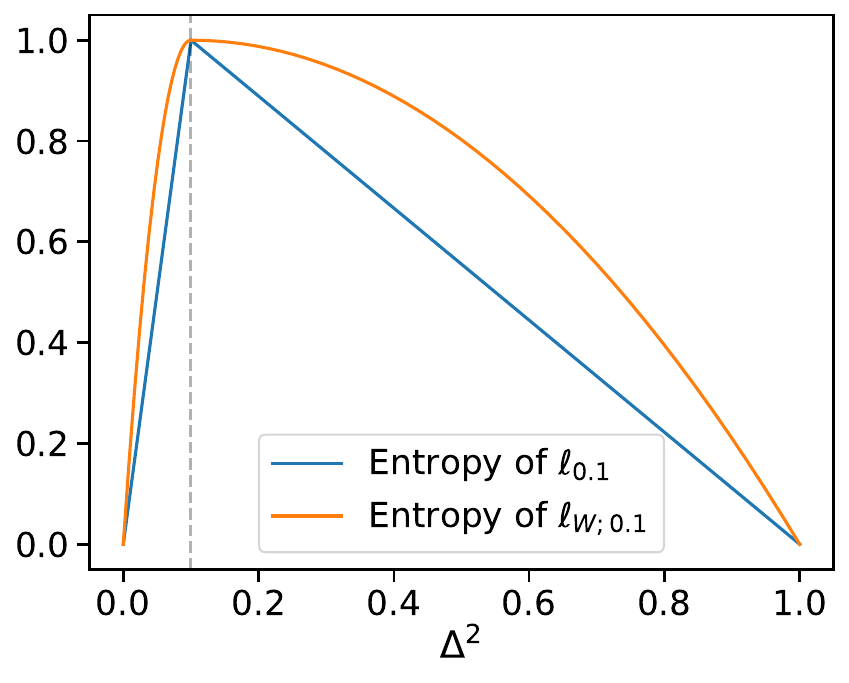}
    \caption{Comparison of unconditional entropies $\hat{H}_{\ell_c}$ and $\hat{H}_{\ellwc}$ (rescaled to obtain maximum value of $1$) of $\ell_c$ and $\ellwc$ with $c=0.1$. The x-axis corresponds to the set of probabilities $\Delta^2$ on the binary $Y$.}
    \label{fig:entropycomparison}
\end{figure}

Figure~\ref{fig:entropycomparison} compares the unconditional entropy for the loss functions $\ell_{0.1}$ and $\ell_{W;0.1}$ (recall Section~\ref{sec:dataanddecisions}).

\subsection{Detailed Information About Training Procedures}
\label{app:training}
Our models have the following structure: a linear layer followed by a sigmoid, which when the input data is $X \in \mR^{N\times D}$ outputs a precise probabilistic forecast $Q \in \mR^{N \times 2}$ (where the forecast for a single datum is in $\Delta^2$, since $Y \in \{0,1\}$). This forecast is then fed into a loss function of signature $\ell : \Delta^2 \times \mathcal\{0,1\} \to \mR$.

\textbf{ERB(log)}:
For the ERM model, the training objective is:
\[
 \min_{Q \in \operatorname{Lin}} \E_{\hat{P}}\left[S_\ell(Q(X),Y)\right],
\]
where $\hat{P}$ is the empirical distribution of the stacked data, \ie the combined data of all groups.
We train using the Adam optimizer (with the \texttt{pytorch} default options) for $5{,}000$ iterations and a learning rate of $0.001$. If we use batches (see below), then we draw a random batch at each iteration.

\textbf{GBR}: To obtain the GBR, we perform a similar procedure, where for each group we train an individual model, meaning
\[
 \min_{\mQ_i : \in \operatorname{Lin}} \E_{\hat{P}_i}\left[S_\ell(Q_i(X),Y)\right],
\] 
where $\hat{P}_i$ is the empirical train distribution (the training data) of group $i$.
We then use 
\[\mQ(X) \coloneqq \{Q_i(X) : i=1..G\}
\]
as an estimate of the GBR. We also train with the Adam optimizer (with the \texttt{pytorch} default options) for $5{,}000$ iterations and a learning rate of $0.001$. As above, if we use batches (see below), then we draw a random batch at each iteration.

\textbf{DRO(log)} and \textbf{DRO(c)}:
We apply the maximum entropy principle, so the minimization of the IP score can be expressed as maximizing entropy:
\[
 \max_{\lambda \in \Delta^G} \min_{Q \in \operatorname{Lin}} \E_{\hat{P}_\lambda}\left[S_\ell(Q(X),Y)\right], \quad \hat{P}_\lambda \coloneqq \sum_{i=1}^G \lambda_i \hat{P}_i,
\]
where $\lambda=(\lambda_1,..,\lambda_G)^\intercal \in \Delta^G$. The outer (maximizing) routine is described in Section~\ref{sec:modelsandtraining}; more concretely, we perform $2{,}000$ such outer iterations with a learning rate of $\eta = 0.1$. In our examples, we have observed that this choice worked well as a compromise between stability and updating $\lambda$ sufficiently. 

For each outer (maximizing) iteration, we approximate the inner minimization for fixed $\lambda$. We do this using the Adam optimizer (with the \texttt{pytorch} default options) for $500$ iterations with a learning rate of $0.001$. At each iteration, we draw a random batch (if we use batches) for each group, or use the full data for each group (if we don't use batches), and then compute the mean loss on the training distribution for each group. Weighting by the currently fixed $\lambda$ then gives an estimate of $\E_{\hat{P}_\lambda}\left[S_\ell(Q(X),Y)\right]$. 

At the end of the inner (minimizing) routine, we need to estimate the gradient $\nabla \lambda$, which simply corresponds to the current mean group losses. When using batches, these can be highly variable, so we instead compute the group losses for multiple batches simultaneously to reduce the variance in our estimate for $\nabla \lambda$. In our experiments, we fixed the number of these batches to $10$. 

Batch Training: For \textsc{acs pums} and \textsc{celeba}, we use a relatively large batch size of $512$ to improve stability.
For \textsc{framingham}, we use the full data instead of batches due to its manageable size.

\subsection{Additional Results}
\label{app:moreresults}
In this appendix, we provide additional results for the datasets, which are described in Appendix~\ref{app:data}.

\begin{table}[]
\scriptsize
    \centering
\begin{tabular}{lllll}
\toprule
 & $\ell_{0.1}$ & $\ell_{0.3}$ & $\ell_{0.7}$ & $\ell_{0.9}$ \\
Predictor &  &  &  &  \\
\midrule
GBR & 0.0502 & 0.1477 & 0.1412 & 0.0497 \\
DRO($0.1$) & 0.0662 & 0.1522 & 0.1438 & 0.0666 \\
DRO($0.3$) & \textbf{0.0487} & \textbf{0.1333} & 0.1370 & 0.0586 \\
DRO($0.7$) & 0.0509 & 0.1465 & \textbf{0.1316} & \textbf{0.0491} \\
DRO($0.9$) & 0.0497 & 0.1461 & 0.1666 & 0.0530 \\
ERM(log) & 0.0500 & 0.1459 & 0.1576 & 0.0492 \\
DRO(log) & 0.0492 & 0.1429 & 0.1347 & 0.0507 \\
\bottomrule
\end{tabular}
\begin{tabular}{lllll}
\toprule
 & $\ell_{0.1}$ & $\ell_{0.3}$ & $\ell_{0.7}$ & $\ell_{0.9}$ \\
Predictor &  &  &  &  \\
\midrule
GBR & 0.0516 & 0.1535 & \textbf{0.1376} & \textbf{0.0482} \\
DRO($0.1$) & 0.0737 & 0.1599 & 0.1831 & 0.0916 \\
DRO($0.3$) & \textbf{0.0500} & \textbf{0.1430} & 0.1542 & 0.0771 \\
DRO($0.7$) & 0.0545 & 0.1500 & 0.1524 & 0.0536 \\
DRO($0.9$) & 0.0518 & 0.1488 & 0.1922 & 0.0886 \\
ERM(log) & 0.0507 & 0.1452 & 0.1759 & 0.0705 \\
DRO(log) & 0.0508 & 0.1434 & 0.1548 & 0.0633 \\
\bottomrule
\end{tabular}

    \caption{Train (left) and test (right) IP Scores of different forecasts (rows), evaluated under different loss functions (columns) on \textsc{framingham} data.}
    \label{tab:ipscoresframingham}
\end{table}

\begin{figure}
\centering
 \includegraphics[width=1.0\linewidth]{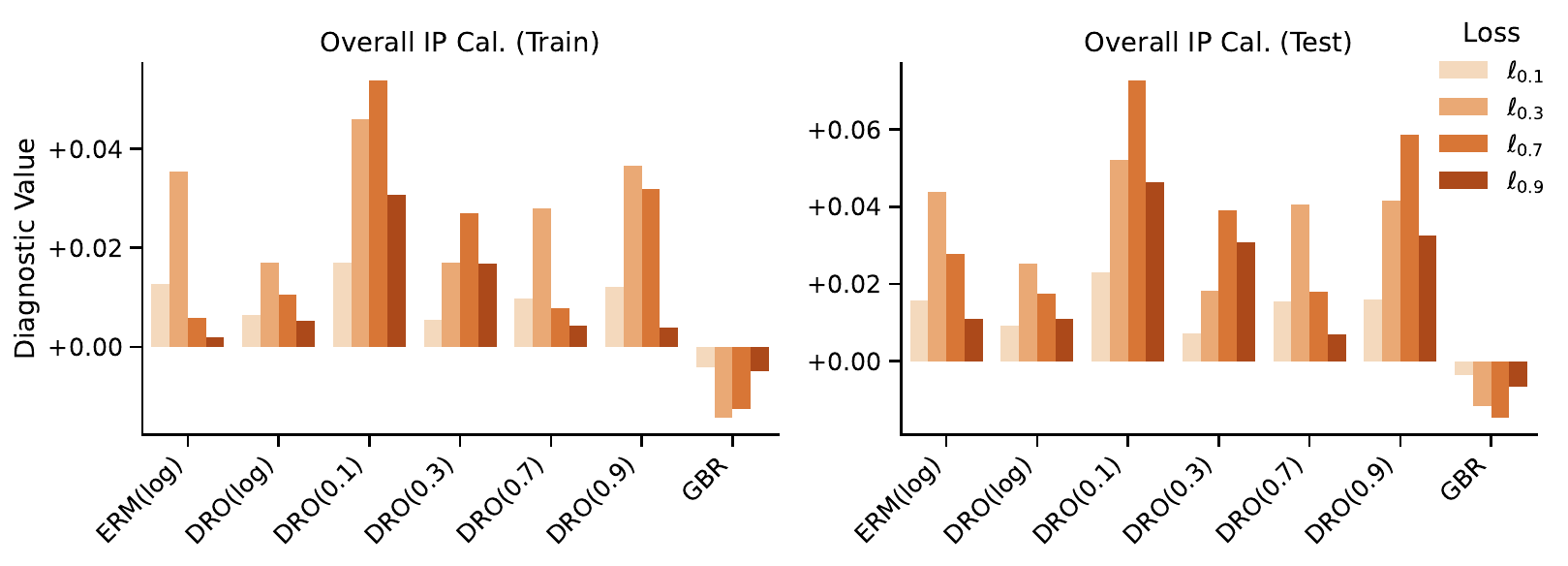}
  \includegraphics[width=1.0\linewidth]{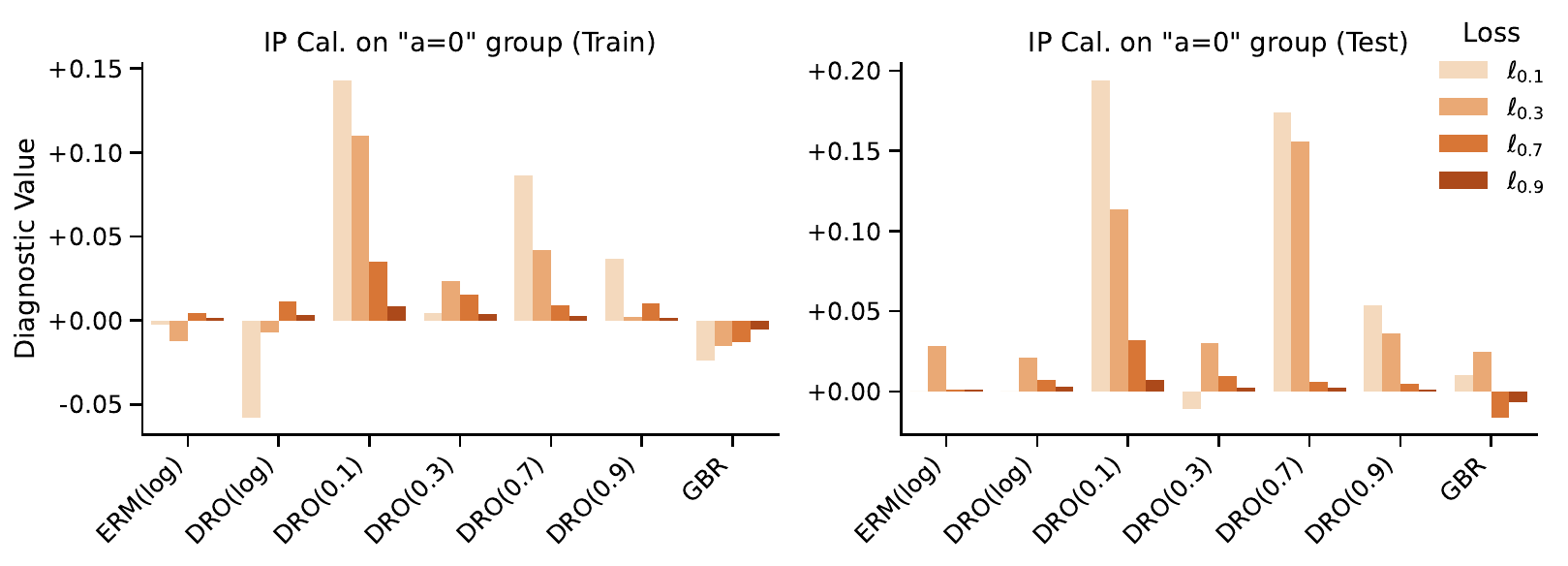}
   \includegraphics[width=1.0\linewidth]{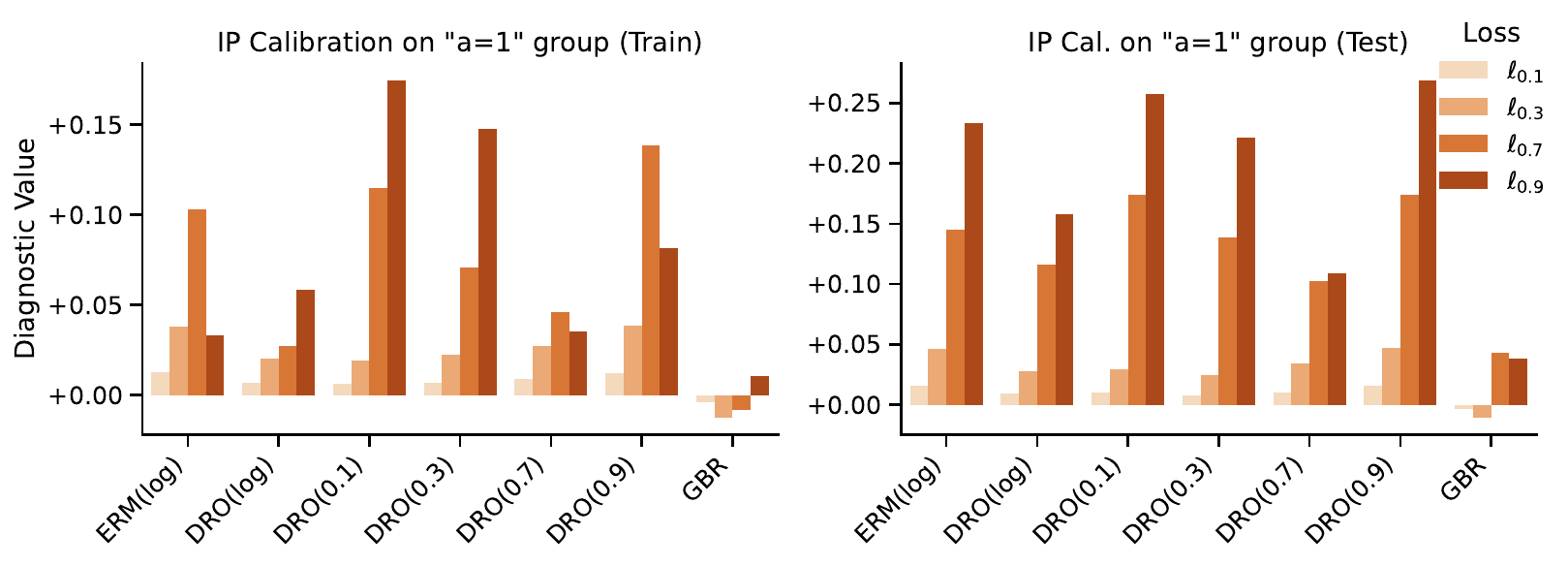}
    \caption{Evaluation of IP calibration under the train (left column) and test (right column) data models on \textsc{framingham}. In each subplot, the forecasts $\mQ$ vary along the $x$-axis and the loss function used for evaluation corresponds to a hue. The top row shows diagnostic values for IP calibration without groups, that is, the $y$-axis shows $\Rup_{\hat{\baseip}}\left(\Sl(\omega) - \Rup_{\mQ(\omega)}\left( \Sl \right) \right)$ (top left) and $\Rup_{\hat{\baseip}_\text{test}}\left(\Sl(\omega) - \Rup_{\mQ(\omega)}\left( \Sl \right) \right)$ (top right).
    In the other subplots, the $y$-axis shows the diagnostics for IP calibration with respect to the action-induced partition (of Equation~\ref{eq:deccaldiagnostic}) under $\hat{P}$ (left) and $\hat{P}_{\text{test}}$ (right), respectively.
    }
    \label{fig:framingham_deccal_train}
\end{figure}

\begin{table}[]
\scriptsize
    \centering
\begin{tabular}{lllll}
\toprule
 & $\ell_{0.1}$ & $\ell_{0.3}$ & $\ell_{0.7}$ & $\ell_{0.9}$ \\
Predictor &  &  &  &  \\
\midrule
GBR & 0.0878 & 0.1323 & 0.0919 & 0.0306 \\
DRO($0.1$) & \textbf{0.0660} & 0.1362 & 0.0896 & 0.0310 \\
DRO($0.3$) & 0.0861 & \textbf{0.1282} & 0.0874 & 0.0306 \\
DRO($0.7$) & 0.0874 & 0.1446 & \textbf{0.0851} & 0.0306 \\
DRO($0.9$) & 0.0869 & 0.1435 & 0.0855 & \textbf{0.0302} \\
ERM(log) & 0.0769 & 0.1487 & 0.0899 & 0.0308 \\
DRO(log) & 0.0882 & 0.1450 & 0.0855 & 0.0306 \\
\bottomrule
\end{tabular}
\begin{tabular}{lllll}
\toprule
 & $\ell_{0.1}$ & $\ell_{0.3}$ & $\ell_{0.7}$ & $\ell_{0.9}$ \\
Predictor &  &  &  &  \\
\midrule
GBR & 0.0880 & 0.1375 & 0.0966 & \textbf{0.0322} \\
DRO($0.1$) & \textbf{0.0672} & 0.1396 & 0.0957 & 0.0326 \\
DRO($0.3$) & 0.0863 & \textbf{0.1280} & 0.0913 & 0.0324 \\
DRO($0.7$) & 0.0880 & 0.1449 & \textbf{0.0893} & 0.0323 \\
DRO($0.9$) & 0.0869 & 0.1442 & 0.0902 & 0.0324 \\
ERM(log) & 0.0774 & 0.1559 & 0.0945 & 0.0326 \\
DRO(log) & 0.0886 & 0.1452 & 0.0909 & 0.0324 \\
\bottomrule
\end{tabular}
    \caption{Train (left) and test (right) IP Scores of different forecasts (rows), evaluated under different loss functions (columns) on \textsc{celeba} data.}
    \label{tab:ipscoresceleba}
\end{table}

\begin{figure}
\centering
 \includegraphics[width=1.0\linewidth]{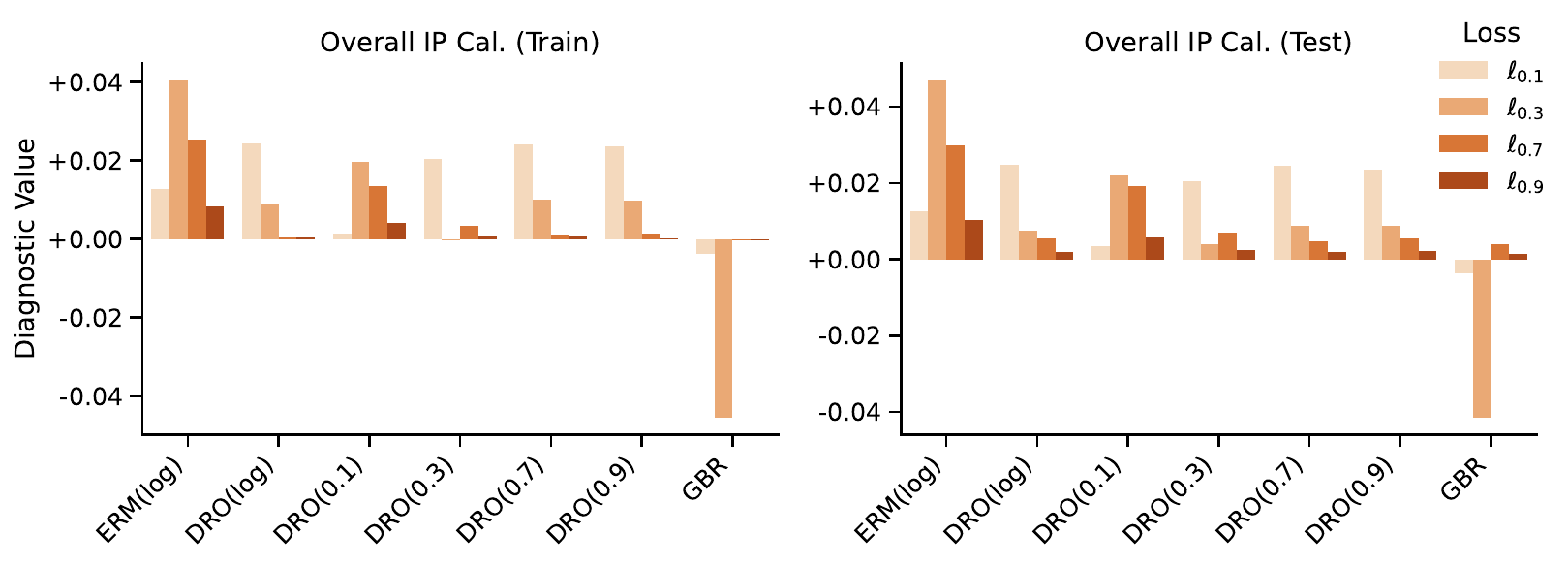}
  \includegraphics[width=1.0\linewidth]{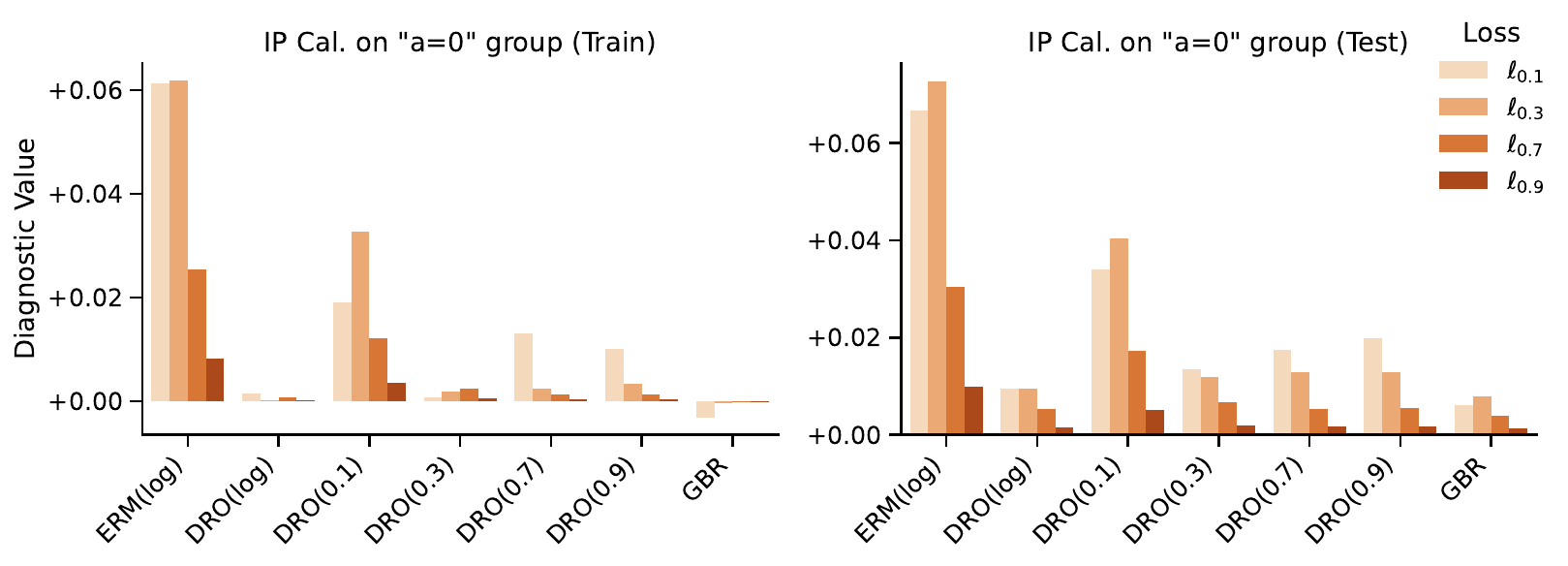}
   \includegraphics[width=1.0\linewidth]{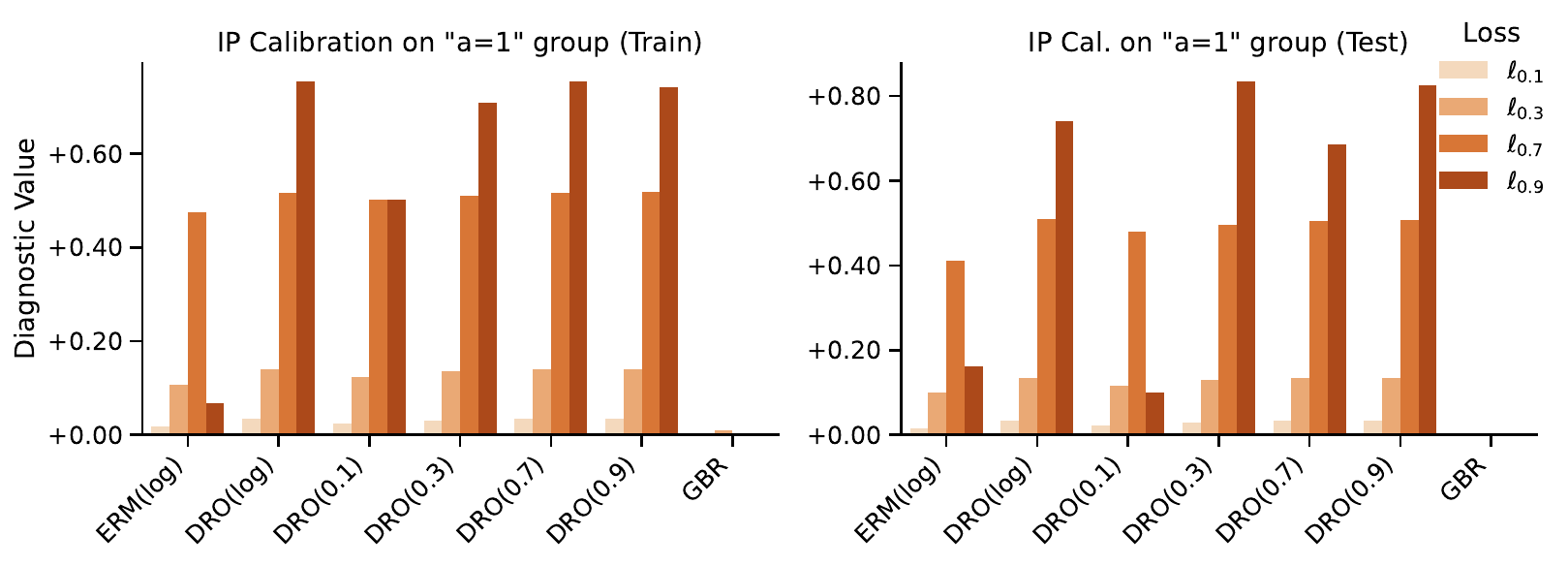}
     \caption{Evaluation of IP calibration under the train (left column) and test (right column) data models on \textsc{celeba}. In each subplot, the forecasts $\mQ$ vary along the $x$-axis and the loss function used for evaluation corresponds to a hue. The top row shows diagnostic values for IP calibration without groups, that is, the $y$-axis shows $\Rup_{\hat{\baseip}}\left(\Sl(\omega) - \Rup_{\mQ(\omega)}\left( \Sl \right) \right)$ (top left) and $\Rup_{\hat{\baseip}_\text{test}}\left(\Sl(\omega) - \Rup_{\mQ(\omega)}\left( \Sl \right) \right)$ (top right).
    In the other subplots, the $y$-axis shows the diagnostics for IP calibration with respect to the action-induced partition (of Equation~\ref{eq:deccaldiagnostic}) under $\hat{P}$ (left) and $\hat{P}_{\text{test}}$ (right), respectively.
    }
    \label{fig:celeba_deccal_train}
\end{figure}

\clearpage
\subsection{The link to \citet{zhao2021right}}
\label{app:zhaoermonlink}
To formally relate our IP calibration to the framework of \citet{zhao2021right}, consider the case of a binary event $\omega$ and an imprecise forecast $\mQ=[\underline{q},\overline{q}] \subseteq [0,1]$ and the cost-sensitive loss $\ell_c$. Then, $\Rup_{\mQ}(\ell(a_\mQ^*,\omega)) = (1-c)\overline{q}$ if $a_{\mQ}^*=0$, and $\Rup_{\mQ}(\ell(a_\mQ^*,\omega)) = c(1-\underline{q})$ if $a_{\mQ}^*=1$. \citet{zhao2021right} suggest using the stakes $b = \ell(a,1)-\ell(a,0)$ (their Proposition~2) in the marginally desirable gamble $\operatorname{MG} \coloneqq b(Y-\mu)-|b|c$; the gambles of this form are marginally desirable if $[\underline{q},\overline{q}]=[\mu-c,\mu+c]$ which we now assume. 
Now consider setting the stakes $b$ based on the action recommendation $a_{\mQ}^*$.
The agent will then shoulder the loss:
\[
L \coloneqq \ell(a_{\mQ}^*,\omega) - \operatorname{MG}.
\]
First, note that $\operatorname{MG}$ can be rewritten as:
\[
\begin{cases}
    +b(\omega-\overline{q}) & b \geq 0,\\
   -b(\underline{q}-\omega) & b<0
\end{cases}
\]
We now compute $L$ (plugging in the respective stakes $b$) for all combinations of $a_{\mQ}^*$ and $\omega$.

If $a_{\mQ}^*=1$ and $\omega=0$:  
\[
L = \ell(0,0) - (1-c)(0-\overline{q})=(1-c)\overline{q} = \Rup_{\mQ}(\ell_{a_{\mQ}^*}).
\]
If $a_{\mQ}^*=1$ and $\omega=1$:
\[
L = \ell(0,1) - (1-c)(1-\overline{q}) = (1-c)\overline{q} = \Rup_{\mQ}(\ell_{a_{\mQ}^*}).
\]

If $a_{\mQ}^*=1$ and $\omega=0$:  
\[
L = \ell(1,0) - c(\underline{q}-0) = c(1-\underline{q}) = \Rup_{\mQ}(\ell_{a_{\mQ}^*}).
\]

If $a_{\mQ}^*=1$ and $\omega=1$:  
\[
L = \ell(1,1) - c(\underline{q}-1) = c(1-\underline{q}) = \Rup_{\mQ}(\ell_{a_{\mQ}^*}).
\]
Hence the agent will, in any case, shoulder the constant loss $\Rup_{\mQ}(\ell_{a_{\mQ}^*})$.

\end{document}